# NEUROGRAF v1

## AN END-TO-END ROMANIAN-FIRST LANGUAGE MODEL FAMILY

Ionuț Adrian ROTARU [1]

*Romania reports the lowest rate of enterprise artificial-intelligence adoption in the European Union. Romanian users are served predominantly by multilingual models whose training data and tokenization are centered on higher-resource languages, increasing the cost and complexity of small, locally deployed systems.*

*We present Neurograf v1, an experimental end-to-end Romanian-first language-model pipeline developed to study how a Romanian-centered system behaves when the corpus, tokenizer, pre-training, instruction tuning and embedding stages are designed together from scratch. The prototype comprises RoSfera, a 635 GB Romanian pre-training corpus containing 175.7 million documents and 92.4 billion words; an 80k byte-level BPE tokenizer selected through a 21-point vocabulary-size sweep; five decoder-only base models from 160M to 6B parameters trained on the same 131-billion-token stream; three supervised fine-tuned assistants trained on 178,502 synthetic Romanian instruction entries spanning approximately 400k conversational turns; and two contrastively post-trained embedding models. The full pipeline was trained and evaluated using computing infrastructure at ELI-NP/IFIN-HH.*

*The controlled model family exhibits regular scaling of validation loss and substantial growth in Romanian factual and cultural knowledge with model size. On the 1,000-question WWTBM Romanian benchmark, Neurograf-6B reaches 69.89% accuracy on Romanian-context questions, compared with 53.05% for RoLlama2-7B, while Neurograf-3.2B reaches 67.38%, compared with 55.20% for LLMic_v2-3B. These external comparisons are contextual rather than controlled, but indicate that models trained from scratch within the Romanian-first pipeline can reach competitive capability at modest scale. Compute-optimal analysis additionally yields a parameter-scaling exponent of approximately 0.423 over the measured regime.*

*Neurograf v1 is intended as an experimental demonstrator rather than a production system. The experiments identify practical constraints in Romanian-first model construction, including diacritic degradation, heterogeneous web quality, limited leverage from oversampling small curated sources, and language-dependent tokenization effects, while also showing that the pretrained bases can be converted into instruction-following assistants and semantic embedding models. The lessons learned here are intended to guide a future production-oriented Neurograf system for efficient local Romanian-language deployment, including document retrieval, administrative assistance and other public-sector digitalization tasks.*

[1] Extreme Light Infrastructure Nuclear Physics, Horia Hulubei National Institute for R & D in Physics and Nuclear Engineering, 077125 Bucharest-Magurele, Romania

**Table of contents**

## 1. Introduction

Romania has the lowest rate of enterprise artificial-intelligence adoption in the European Union. In 2025, 5.2% of Romanian enterprises with at least ten employees reported using AI technologies, compared with an EU average of 20.0% [1]. Denmark recorded the highest share at 42.0%, followed by Finland at 37.8% and Sweden at 35.0%. Romania therefore remained last among EU member states, behind Poland and Bulgaria, with an adoption rate approximately eight times lower than that of the leading country.

Adoption gaps of this size have many causes, but language is one factor that compounds it: when models handle the working language inefficiently or unreliably, deployment becomes more expensive, less capable, or dependent on external services. Romanian users are served predominantly by multilingual models whose training and tokenization are centered on higher-resource languages, and that arrangement carries costs at every layer of the stack.

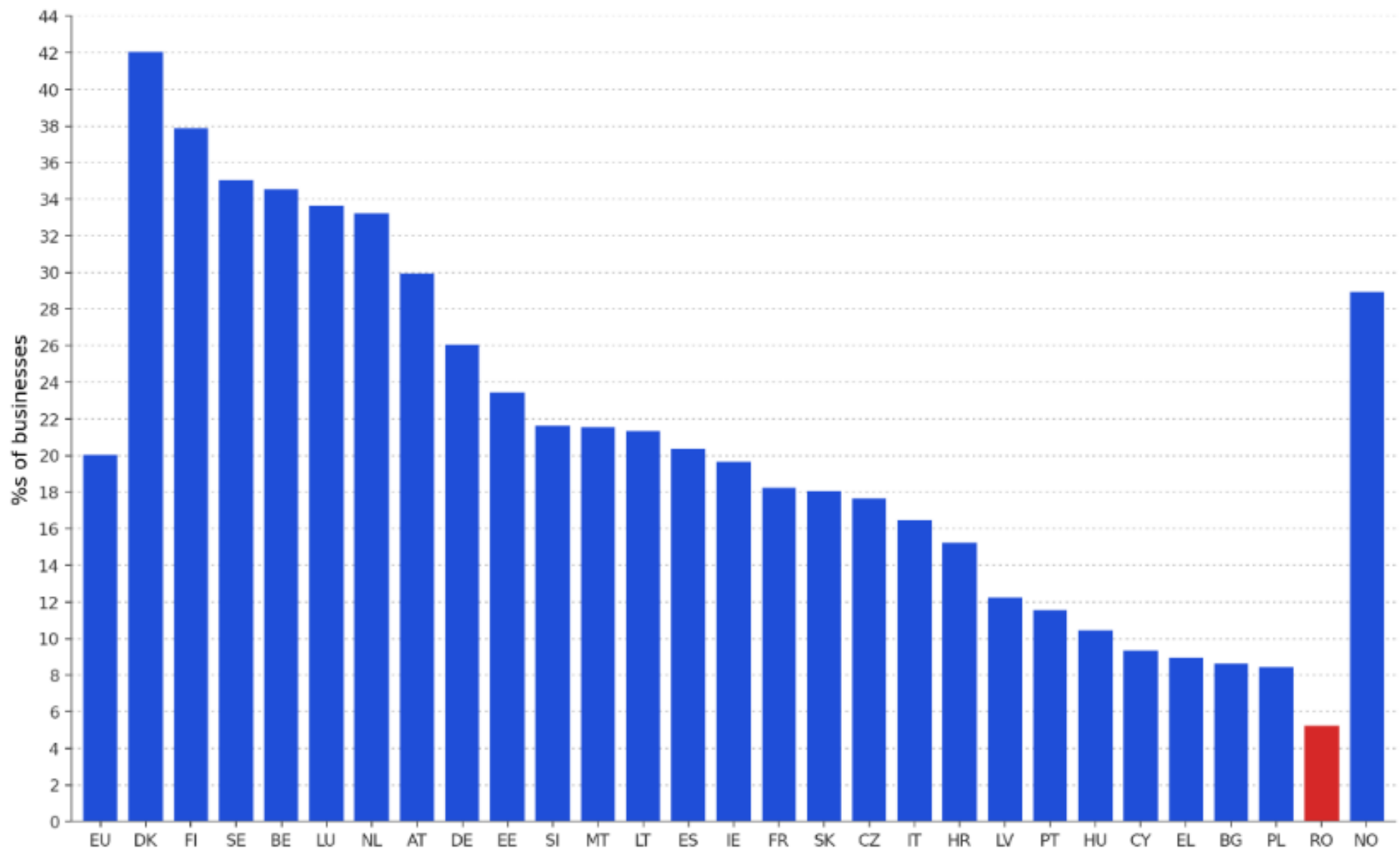

*Figure 1: Enterprises using artificial-intelligence technologies in 2025, expressed as a percentage of enterprises with at least ten employees. Romania is highlighted in red. Source: Eurostat [1]*

One of the most direct and measurable costs appears at the tokenizer level. Subword vocabularies fitted predominantly to English fragment Romanian text into more tokens per unit of content, and Romanian diacritics (ă, â, î, ş, ţ) may also receive inefficient subword representations, and in byte-fallback tokenizers can ultimately decompose into byte-level units.

To quantify this, we tokenized 10,000-character samples of Romanian web text drawn from FuLG [2] at each of the three filtering levels described in Appendix A, using four widely deployed English-centric tokenizers and the production Neurograf 80k tokenizer (Table 1). At the lightest filtering level, the English-centric vocabularies require between 1.22× and 1.57× as many tokens as Neurograf for identical content. The more informative result is what happens as the filtering tightens. The heavy curated tier admits only lines containing all five Romanian diacritics and is therefore a proxy for diacritic density. Across that sweep, every English-centric tokenizer inflates — by 12.1% for Gemma and 27.2% for Phi-2 — while Neurograf remains flat to within 2%. The gap at the heavy tier reaches 2.03× for Phi-2: an English-centric vocabulary requires twice as many tokens to encode the same well-formed Romanian sentences.

| Tokenizer | lite curated | medium curated | heavy curated | drift |
|---|---|---|---|---|
| Phi-2 (50k) [3] | 2.77 c/t | 2.5 c/t | 2.18 c/t | -27.2% |
| Llama-3 [4] (128k) | 3.22 c/t | 3.09 c/t | 2.8 c/t | -14.8% |
| DeepSeek R1[5] (129k) | 3.19 c/t | 3.07 c/t | 2.83 c/t | -12.9% |
| Gemma [6] (256k) | 3.56 c/t | 3.51 c/t | 3.17 c/t | -12.1% |
| Neurograf (80k) | 4.33 c/t | 4.46 c/t | 4.42 c/t | +1.9% |

*Table 1: Python characters per token (c/t – a higher value is better) when encoding 10,000-character samples of Romanian text at three curation levels, with relative drift from lite to heavy curation.*

The consequence is mechanical. For a fixed token budget, the Neurograf tokenizer can represent approximately 1.4× to 2.0× more Romanian text than the English-centric tokenizers tested here. In deployment, this can translate into longer effective context and fewer generated tokens for equivalent Romanian content. The penalty is not uniform across text either: it is largest precisely on well-formed, fully diacritized prose like the register of legal, administrative and institutional documents, which is the register the unserved deployment cases depend on. In other words, the tokenization penalty grows as the text becomes more characteristically Romanian.

This matters most precisely where the adoption gap is widest. Frontier models accessed through commercial APIs will continue to serve users who can afford them. Local models are particularly relevant in deployment settings such as public administration handling citizen data that cannot leave the country, small organizations without cloud budgets, and interactive services where latency and per-token cost decide feasibility. These need small, efficient, locally deployable models that speak Romanian natively, not frontier capability, but reliable competence at form completion, document retrieval, classification, translation, and guided user interaction.

Work on Romanian language models has followed several complementary approaches. The OpenLLM-Ro project has produced a series of Romanian models — RoLlama, RoMistral, RoGemma [7,8] — by adapting existing open models through continued pre-training and instruction tuning, with much of the instruction data obtained through translation. Separately, LLMic [9] provides a 3B bilingual

Romanian–English model trained from scratch with a dedicated tokenizer, alongside FuLG, a large Romanian corpus extracted from the CommonCrawl. Earlier Romanian-specific encoder models, including Romanian BERT [10] and RoBERT [11], established the effectiveness of monolingual pre-training for Romanian language understanding.

These are substantial contributions and have directly inspired the present work, while leaving a different design space unexplored. Continued-pretraining approaches generally retain the tokenizer and architecture of the parent model, including any Romanian fragmentation cost already embedded in that vocabulary. A bilingual foundation model instead allocates its training capacity across Romanian and English.

In the present work, we present Neurograf, an end-to-end Romanian language model family trained from scratch. Every stage is Romanian-first: the corpus, the tokenizer, the base models, the instruction data and the embedding models.

**The pipeline comprises:**

- **RoSfera**, a Romanian pre-training corpus of 175.7 million documents, 92.4 billion words and 606.2 billion characters, approximately 635 GB as UTF-8, assembled from nine sources across web, institutional and literary registers.
- **An 80k byte-level BPE tokenizer** trained on RoSfera, selected on the basis of a 21-point vocabulary sweep spanning 8k to 2M tokens in Romanian and English.
- **Five base models:** 160M, 560M, 1.5B, 3.2B and 6B parameters — each trained on the identical 131-billion-token stream.
- **A synthetic instruction pipeline** in which approximately 100k topic seeds mined from Romanian FineWeb-2 URL metadata are expanded and converted into Romanian instruction data by a locally hosted GPT-OSS-120B generator, producing a final dataset of 178,502 entries spanning approximately 400k conversational turns.

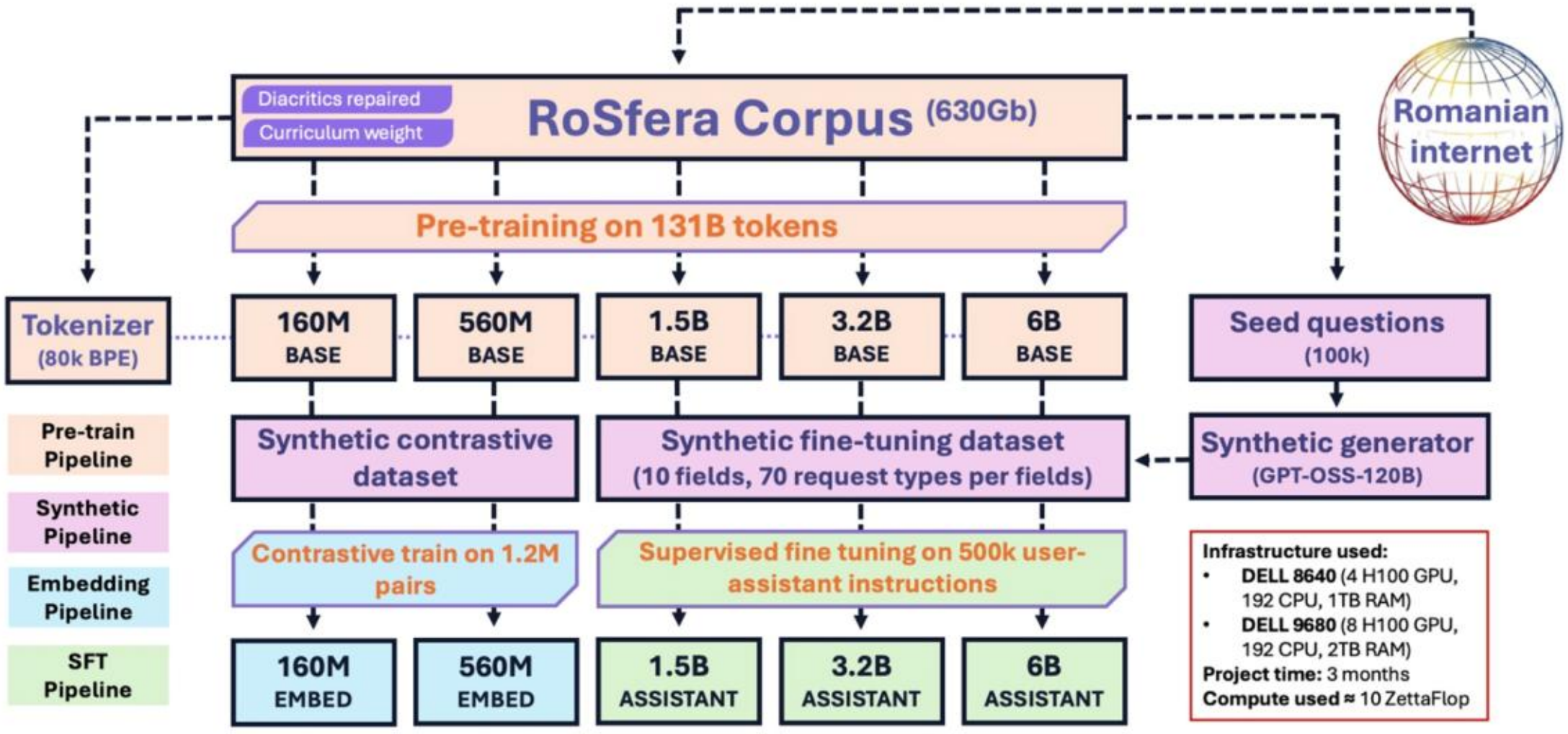

*Figure 2: System overview diagram of the entire Neurograf pipeline*

- **Three assistant models** (1.5B, 3.2B, 6B) supervised fine-tuned on that data.
- **Two embedding models** (160M, 560M) trained contrastively on 1.2 million pairs derived from the same synthetic material.

Because the instruction data is generated natively in Romanian rather than translated from English, and because the generator is Apache-2.0 licensed, this avoids dependency on the terms of service of proprietary model APIs used for synthetic-data generation.

Training was performed on two local systems equipped with four and eight NVIDIA H100 GPUs respectively, over approximately three months. Pre-training consumed 9.06 ZettaFLOP across 5,682 H100-hours. The training code is a modified version of nanoChat [15]; we describe our modifications in the Pre-training Section.

This is an experimental project. It was undertaken to establish whether a Romanian-native model family could be built end to end with modest resources, and to learn how. The objective is therefore to document and characterize the construction, training, and behavior of the system, rather than to claim state-of-the-art downstream performance.

## 2. The RoSfera Corpus

RoSfera is a Romanian pre-training corpus of 175,673,214 documents, 92.42 billion words and 606.21 billion characters, occupying approximately 635 GB as UTF-8. It is assembled from nine sources spanning three registers: web-derived text (88.9% of words), institutional and administrative text (9.1%), and literary or encyclopedic text (2.0%).

**2.1 On measuring corpus size.** We report corpus size in documents, words, characters and bytes, and avoid per-source token counts.

Token counts are a property of the tokenizer, not of the text. The same Romanian document yields substantially different counts under an English-centric subword vocabulary than under one fitted to Romanian, and the ratio varies further by register. Published token totals for Romanian corpora are therefore not comparable across papers unless the tokenizer is held fixed. Token counts therefore characterize the interaction between corpus and tokenizer rather than corpus size alone. Section 4 quantifies this directly.

In total, pre-training consumed 131 billion tokens under the tokenizer of Section 3.9, corresponding to 4.63 characters or 4.85 bytes per token across the corpus.

**2.2 Sources:**

- **Web-derived (88.9%).** Three large-scale web extractions: the Romanian portion of the latest HPLT [12], the Romanian portion of FineWeb [13], and a curated subset of FuLG [2]. FuLG is not used as distributed, the subset retained here is the product of a deliberately strict filtering for high-quality language, reducing it to 21,408,360 documents and 7.54 billion words, about 8.2% of the corpus, well below its nominal share of available Romanian web text (see Appendix A). Among the web-derived sources, the heavy curated FuLG subset scores best on all corpus-quality indicators examined in this work. We read this as evidence that aggressive curation of Romanian web text works, and that it is expensive in volume. HPLT provides broad web coverage; FineWeb-Ro provides a more strongly filtered view of overlapping web material; the heavily curated FuLG subset provides a high-quality tier. Because these sources overlap in origin, material surviving increasingly stringent filtering may effectively receive increased exposure during training.

- **Institutional (9.1%).** Administrative PDF documents, court decisions (Justice), published jurisprudence, and parliamentary transcripts. These registers are less naturally represented by general web-derived corpora and are particularly relevant to the public-administration applications motivating this work. They are also, as shown below, the least lexically productive and the most internally heterogeneous material in the corpus.
- **Literary and encyclopedic (2.0%).** Books and Romanian Wikipedia. Small in absolute terms — Wikipedia is 0.03% of corpus words — but the cleanest text available, and the reason oversampling was attempted at all.

Diacritic handling in the pipeline is encoding normalization, not restoration: Turkish-derived cedilla forms (ş, ţ, namely U+015F, U+0163) are mapped to the Romanian comma-below forms (ș, ț, namely U+0219, U+021B). No attempt is made to restore diacritics absent from the source text, and Section 2.5 shows that a substantial fraction of the corpus carries none.

**2.3 Measurement protocol.** Each source is partitioned into approximately 1,000 shards. The basic unit of data processing is the text block, typically corresponding to a single source item (for example, a web page, PDF document, or Wikipedia article), although some sources are segmented into fixed-size blocks as described below. For every text block we extract ten statistics: block identifier, total characters, letters, diacritics, digits, whitespace, punctuation, words, unique words, and mean-word length. Diacritics are counted as a subset of letters. We use these metrics to compute the following:

**2.3.1. Lexical productivity.** Vocabulary grows sublinearly with text length: the longer a text, the more distinct word forms it contains, but each additional word is less likely to be one not already seen. Heaps' law [16,17] states this as follows:

$$V = KN^{\alpha} \quad (1)$$

for vocabulary size V against word count N, with α between 0 and 1. The exponent measures how readily a source keeps producing new forms — high for material that continually introduces new subject matter, low for formulaic registers that reuse a fixed inventory. Because it is a growth rate rather than a count, it is comparable across sources of very different sizes. We fit the law in logarithmic form:

$$\log(V) = \log(K) + \alpha \log(N) \quad (2)$$

over blocks of at least 100 words, mass-weighted by N, with 95% confidence intervals from a shard-level bootstrap [18].

Two choices matter. We fit log V directly rather than regressing log(V/N) on log N. The ratio is the more familiar quantity, but N then appears on both sides of the equation, and long blocks yield small ratios by construction; the regression returns a negative slope that is an artefact of the variables rather than a property of the text. We also bootstrap over shards rather than blocks, because blocks within a shard share sites, document batches and extraction runs, and block-level resampling therefore treats correlated observations as independent, yielding intervals far too narrow.

The same dependence motivates the block-length restrictions used elsewhere in this section. Since vocabulary is a function of length, any comparison between two populations must hold length constant or the difference in length will present as a difference in vocabulary. Accordingly, where we compare populations within a source in Sections 2.5.1 and 2.7, we restrict the analysis to a fixed length window.

**2.3.2. Romanian Fidelity Index.** We define the RFI of a text block as the ratio of diacritic characters to letters. Correctly written Romanian carries diacritics at a characteristic rate — ă, â, î, ş and ţ occur frequently enough in ordinary prose that their near-absence is conspicuous. Informally typed Romanian usually omits them: keyboard layouts, messaging habits and legacy encodings all favor the unmarked forms, and much of the Romanian web is written this way. The RFI detects this directly, at block granularity, without a language model or a lexicon, and at a cost low enough to run over the entire corpus.

The index measures rate, not correctness. A block with an RFI near zero has lost its diacritics, but the converse does not hold: a diacritized block may still be poor Romanian, and the reference rate itself varies by register, since some vocabulary carries more marked characters than others. We use the RFI throughout this section as a positional measure within a distribution rather than as a quality threshold, and Section 2.5.3 sets out why it cannot serve as the latter.

### 2.4 Curriculum construction.

| Source | Register | Passes | Raw share | Curriculum share |
|---|---|---|---|---|
| HPLT | Web | 1 | 50.72% | 50.16% |
| FineWeb-Ro | Web | 1 | 30.02% | 29.69% |
| FuLG (heavy curated) | Web | 1 | 8.16% | 8.07% |
| Administrative PDF | Institutional | 1 | 8.70% | 8.61% |
| Justice | Institutional | 3 | 0.179% | 0.531% |
| Jurisprudence | Institutional | 3 | 0.121% | 0.360% |
| Parliamentary | Institutional | 3 | 0.116% | 0.344% |
| Books | Literary | 1 | 1.96% | 1.94% |
| Wikipedia | Encyclopedic | 10 | 0.031% | 0.302% |
| **Total** | | **24 streams** | **92.42 Bw** | **93.44 Bw (+1.11%)** |

*Table 2: RoSfera sources, oversampling factors and mixture shares. Each source contributes one independently shuffled stream per pass, giving 24 streams in total. Raw share is the source's proportion of the 92.42-billion-word corpus; curriculum share its proportion of the 93.44-billion-word training mixture. Shares are word-mass weighted and differ from document- or byte-weighted shares: Justice is 0.76% of documents but 0.18% of words, its blocks averaging 123 words. Ten passes over Wikipedia move it from 0.031% to 0.302% of the mixture, and the web-derived share falls only from 88.90% to 87.93%*

The training curriculum is built by expanding each source into one stream per pass (24 streams in total) independently shuffling each with a fixed seed and emitting 1,000 master shards by taking position $i$ from every stream in turn. A second stage shuffles blocks within each master shard. Because composition is fixed entirely by the first stage, every master shard carries near-identical source proportions.

This is a deliberate choice. Holding source proportions constant at the granularity of the batch update, rather than only in expectation over an epoch, eliminates source-composition drift across training updates and reduces same-source clustering within batches.

Oversampling factors are 10 passes for Wikipedia; 3 for jurisprudence, justice and parliamentary text; and a single pass for all other sources.

**2.5 Diacritic degradation.** The dominant quality defect in Romanian web text is missing diacritics. Across the corpus, 20.8% of text mass carries essentially none (RFI below 0.005), and 26.0% falls below 0.02.

The degradation concentrates in the two largest sources: HPLT at 27.3% and FineWeb-Ro at 22.1%, which together constitute 81% of the corpus. The contrast with the highly-curated FuLG subset, at 0.000%, and Wikipedia, at 0.23%, spans three orders of magnitude within the same language drawn from the same web. The prevalence of diacritic loss is therefore strongly dependent on source selection, extraction and curation rather than being an unavoidable property of Romanian web text.

Jurisprudence exhibits a third profile: only 4.2% of its mass is fully stripped, but 17.2% is degraded. This is partial rather than total loss, a distinct failure mode in which some diacritics survive and others do not, probably because of the PDF extraction tools employed.

**2.5.1 Diacritic loss destroys word types.** Romanian marks meaningful distinctions with its diacritics. Remove them and distinct words collapse onto the same spelling: *fată* (a girl) and *fata* (the girl) both become *fata*; *și* (and) and *si* (a musical note) both become *si*; *în* (in) and *in* (flax) both become *in*; *țară* (country) and *tara* (the flaw) both become *tara*; *păr* (hair) and *par* (a stake) both become *par*.

A text that has lost its diacritics therefore contains fewer distinct word forms than the same text with them intact — not because it says less, but because words that were different now look identical. This is measurable, and it is what we measure here.

The measurement needs one control. Longer texts naturally contain more distinct forms, so a raw count would confound diacritics with length. We restrict the comparison to blocks of 1,000–4,000 words and score each block against how many distinct forms its own source's Heaps fit predicts for a text of that length. This removes the dominant block-length effect, allowing the remaining vocabulary difference to be associated more directly with diacritic status.

| Source | Fewer distinct forms | 95% CI |
| --- | --- | --- |
| HPLT | -5% | −5.3%, −4.7% |
| Jurisprudence | -3.3% | −4.1%, −2.4% |
| FineWeb-RO | -2.4% | −2.5%, −2.3% |

*Table 3: Vocabulary lost to diacritic stripping. Distinct word forms in fully stripped blocks, relative to fully diacritized blocks of the same length from the same source. Confidence intervals from a shard-level bootstrap. Only these three sources admit the comparison: FuLG and Wikipedia contain almost no stripped text, Books and parliamentary transcripts were extracted at fixed block size so their Heaps fits are not identified (Section 2.6), and administrative PDF and justice carry large collapsed-diversity populations that would dominate the contrast.*

Jurisprudence gives the cleanest evidence. Its stripped and diacritized documents share the same genre and the same document templates, so almost nothing but the diacritics differs between them. In the two web sources the two populations are different documents from different sites, and part of the gap may reflect that rather than the diacritics.

Across sources covering 81% of the corpus, stripping costs between 2.4% and 5.0% of the distinct word forms, with intervals excluding zero. This measurement captures only one component of the damage: the disappearance of distinct surface forms. It does not quantify the ambiguity introduced into surviving forms. It counts only forms that merged with another form, not the ambiguity created inside forms that survive, every occurrence of *fata* in stripped text is now ambiguous, and the count registers that as one form either way.

The effect is real but small: 0.18 to 0.30 of the variation blocks already show within a single source. It measures a defect in the corpus as a whole and cannot identify which individual documents are damaged, which is why no RFI threshold separates stripped text from clean (Section 2.5.3).

**2.5.2 The dual burden this imposes on the model.** The type-loss measurement understates the difficulty, because inconsistent diacritization imposes two opposite problems simultaneously:

- **Merging.** Where diacritics are absent, distinct words share a surface form. *fată* (a girl) and *fata* (the girl) both appear as *fata*; *în* (in) and *in* (flax) both as *in*; The model must recover the intended sense from context alone, with no orthographic signal.

- **Splitting.** Conversely, because different sources in the corpus are diacritized to different degrees, a single word appears under multiple surface forms. *Mamă* occurs as both *mamă* and *mama* depending on which source a document came from. The model must learn that the alternations *ă/a, â/a, î/i, ș/s* and *ț/t* are usually orthographic noise carrying no meaning — while simultaneously learning the cases shown above, where the same alternations carry the entire meaning.

These pull in opposite directions and cannot be distinguished from surface form. The model must learn a context-dependent disambiguation that a consistently diacritized corpus would never require, requiring the model to represent an additional source of orthographic variation unrelated to the intended semantics. We regard this as one of the largest obvious avoidable defects in the Romanian web data examined here, and we cannot separate its contribution from other factors in our results.

**2.5.3 What the RFI does not measure.** We caution against reading the RFI as a pure fidelity signal. Three observations argue against it.

First, the curated FuLG subset contains no diacritic-free text at all, yet its mean RFI (0.062) sits well below that of Books (≈0.09). The gap must be compositional rather than a matter of missing marks.

Second, Books and Wikipedia, both clean, edited Romanian, occupy different RFI positions, approximately 0.09 and 0.06 respectively (Figure 3). There is therefore no single value that characterizes correctly written Romanian; the reference is genre-dependent, and literary prose is diacritic-denser than encyclopedic prose for reasons of vocabulary rather than fidelity.

Third, the partially diacritized band (RFI 0.01–0.03) shows higher lexical surplus than fully diacritized text in both large web sources (FineWeb-Ro +0.039, [+0.0387, +0.0401]; HPLT +0.034, [+0.030, +0.038]) — the opposite of what type merging predicts, and consistent with that band containing non-Romanian material such as English loanwords, brand names and proper nouns, which is lexically rich.

Consequently, there is no RFI threshold that means "this text needs repair." The index locates a block in a distribution; it does not classify it.

### 2.6 Lexical productivity and heterogeneity.

| Source | α | $R^2$ |
|---|---|---|
| Wikipedia | 0.811 | 0.976 |
| FuLG (curated) | 0.802 | 0.981 |
| FineWeb-Ro | 0.768 | 0.982 |
| HPLT | 0.755 | 0.973 |
| Administrative PDF | 0.722 | 0.914 |
| Justice | 0.687 | 0.913 |
| Jurisprudence | 0.638 | 0.943 |

*Table 4: Per-source Heaps exponents. Vocabulary growth exponent α from the fit of Section 2.3, estimated over blocks of at least 100 words, mass-weighted by N, with 95% confidence intervals within ±0.012 from a shard-level bootstrap. Higher α indicates a source that continues to introduce new word forms as it grows. Books and parliamentary transcripts are omitted: both were extracted at fixed block size, so log N has insufficient variance to identify a slope.*

Applying the fit of Section 2.3 to each source separately gives a per-source lexical productivity exponent. Because α is a growth rate rather than a count, the values are directly comparable across sources despite their very different sizes, from 0.031% of corpus words to 50.7%. The exponent bounds are interpretable: α near 1 would mean nearly every additional word introduces a form not previously seen, and α near 0 that a source reuses a fixed inventory almost exclusively. The observed range, 0.638 to 0.811, sits well inside those limits, and its ordering is the substantive result.

All fits give $R^2$ above 0.91 — the power law accounts for over 91% of the variance in log V across blocks — with confidence intervals within ±0.012. The ordering is the one the registers would predict: encyclopedic and curated prose are the most lexically productive, and legal text the least, possibly because of its formulaic templates and fixed procedural vocabulary reusing a small inventory of forms.

Two sources are omitted. Books and parliamentary transcripts were extracted at fixed block size, approximately 18,000 and 15,000 characters, so log N is nearly constant across their blocks. A slope cannot be estimated without variation in the predictor, and the fits confirm it: they return α = 0.552 ($R^2$ = 0.30) and α = 2.156 ($R^2$ = 0.25), the latter impossible, since vocabulary cannot exceed word count and $\alpha$ must lie below 1. We report both as unavailable rather than as measurements. The restriction is worth stating as a methodological point: Heaps' law cannot be estimated from fixed-size chunks, and any pipeline that segments text at a constant length forecloses this measurement.

For a related reason we report α per source rather than pooled. Sources differ in intercept K as well as slope, and pooling them fits a single line through what are several parallel relationships at different heights, flattening the result. The pooled value of 0.733 lies below four of its own constituents, which is diagnostic of the problem rather than a summary of the family.

We make no cross-linguistic claim from these values. The pooled exponent falls within the range reported for English, but the comparison is not meaningful without a control using an identical extractor and N window across both languages, which we have not performed.

**2.7 Source geometry in the fidelity–productivity plane.** Section 2.6 characterizes each source by a single number, its Heaps exponent. That number describes the central tendency of a source but says nothing about how tightly its blocks cluster around it. Dispersion around a source's own fit — how much individual blocks over- or under-produce vocabulary relative to what their length predicts — carries separate information, and because the fit removes everything the length–vocabulary relationship explains, this residual is largely independent of the RFI. Plotting the two against each other gives a two-dimensional fingerprint for each source (Figure 3). The vertical axis measures internal heterogeneity, the horizontal axis diacritic rate, and the shape of the resulting cloud is diagnostic in ways that neither axis alone is.

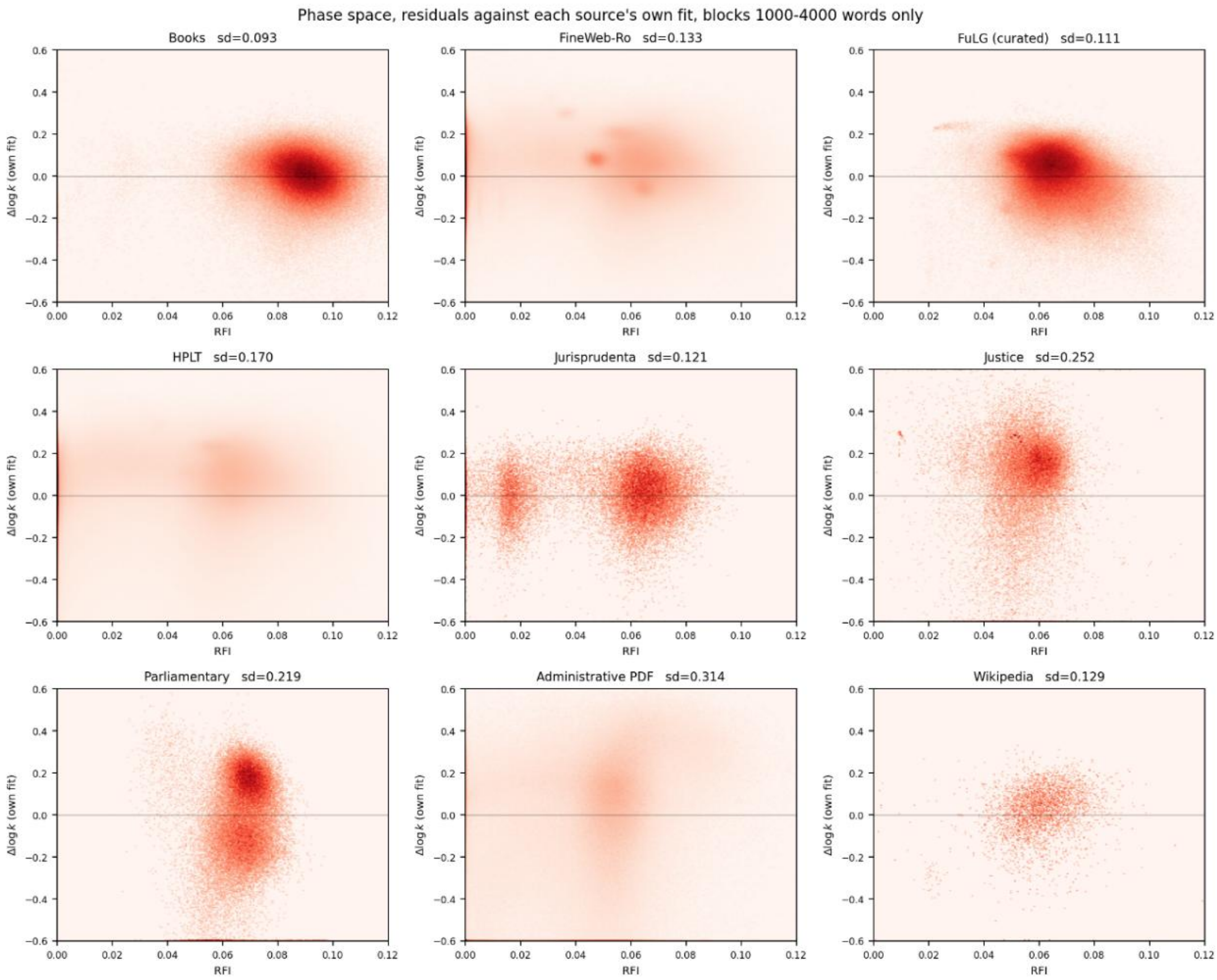

*Figure 3: Source geometry in the fidelity–productivity plane. For each source, residual Δlog k against that source's own Heaps fit (vertical) is plotted against the Romanian Fidelity Index (horizontal), restricted to blocks of 1,000–4,000 words so that block length is controlled. Because the vertical axis removes everything the length–vocabulary relationship explains, the two axes are largely independent, and the sources separate. Note the vertical stripe at RFI = 0 in HPLT and FineWeb-Ro (fully stripped text), its complete absence in curated FuLG, the bimodality of jurisprudence, and the accumulation at the lower plot boundary in administrative PDF and justice, which corresponds to blocks of collapsed lexical diversity. Residual standard deviations are given per panel.*

Read this way, the panels separate into several distinct profiles:

- **Uncurated web.** HPLT and FineWeb-Ro both carry a pronounced vertical stripe at RFI = 0, the fully stripped population, alongside a broad diacritized body. The dispersion is wide, HPLT sd = 0.170, the largest of any web source, which is the expected consequence of aggregating text from a very large number of unrelated sites, and is not in itself a defect: breadth of register and subject matter is what makes general web text useful for pre-training. FineWeb-Ro additionally shows several discrete concentrations within its diacritized body rather than a single mode. These concentrations may correspond to large individual domains or recurring site templates, but we did not trace their provenance.

- **Extraction dependent diacritics.** Jurisprudence is visibly bimodal in RFI, with two well-separated lobes near 0.02 and 0.065 rather than a continuum of partial loss. Bimodality of this kind indicates two document populations rather than degrees of damage within one. The likely mechanism is upstream of the corpus: legal material is distributed as PDF, and diacritic survival depends on how each document was produced and which extraction path recovered it, so the split plausibly reflects two document-production regimes rather than two writing conventions. This is consistent with Section 2.5.1, where jurisprudence provides the cleanest type-loss estimate precisely because genre is held constant across the two lobes.
- **Structurally heterogeneous sources.** Administrative PDF is the most dispersed source in the corpus (sd = 0.314), more than three times the dispersion of Books (0.093). This is unsurprising for a category that mixes blank forms, tabular material and running prose within a single register, and the dispersion is therefore a property of the category rather than a quality signal. Parliamentary transcripts show two separate concentrations on the productivity axis at similar RFI. We note this without a confident interpretation; procedural material and substantive debate differ considerably in lexical variety, and the two lobes may correspond to that distinction, but we have not verified it.
- **Degenerate blocks.** Administrative PDF and justice both accumulate mass at the lower plot boundary, corresponding to blocks whose vocabulary has collapsed. Quantifying this directly, the share of text mass with a unique-to-total word ratio below 0.15 at $N \geq 500$ is 31.9% for administrative PDF and 12.3% for justice, against 2.3% for HPLT, 1.3% for FineWeb-Ro, 0.02% for FuLG and 0.005% for Books. Administrative PDF further contributes 0.96% punctuation-dominated mass, largely tables of contents rendered with dot leaders; one sampled block consisted of 32,385 punctuation characters out of 47,663. These are the failure modes that aggregate statistics conceal, a source-level exponent absorbs them into a slightly lower $\alpha$ and gives no indication that a third of the material is near-degenerate.

We offer this representation as a descriptive tool for corpus assembly rather than as a diversity metric. It makes visible, at a glance, whether a candidate source occupies a region of the plane already covered by existing material, and it surfaces defects, stripped populations, degenerate blocks, bimodality, that aggregated statistics do not. We have not validated it against any downstream outcome, and we make no claim that separation in this plane predicts training value.

**2.8 Limits of oversampling in a web-dominated low-resource corpus.** The curriculum used for pre-training was designed to increase exposure to high-quality curated text. Measured against the raw corpus, it does almost nothing.

Applying the pass counts yields a curriculum of 93.44 billion words against 92.42 billion raw, meaning an increase of 1.1%. Wikipedia moves from 0.031% to 0.302% of the mixture. The web-derived share moves from 88.89% to 87.92%. Exposure to diacritic-degraded text moves from 20.77% to 20.56%, an improvement of two tenths of a percentage point. Table 2 places raw and curriculum shares side by side.

The arithmetic is unforgiving. Bringing Wikipedia to 3% of the mixture would require roughly 100 passes, far beyond the repetition budget at which language modelling degrades (repetition is approximately benign up to four epochs and clearly damaging past sixteen [19]).

We state the resulting conclusion as the principal finding of this section. For a corpus like RoSfera with a large degree of web dominance, curated sources are too small for realistic oversampling factors to substantially change the training mixture before repetition becomes a concern. This makes improvement of the dominant web component a considerably stronger lever than oversampling alone. The curated FuLG subset, at 0.000% diacritic-free text, demonstrates that this is achievable; its 8.2% share demonstrates what it costs.

## 3. Tokenizer

Romanian is commonly tokenized by multilingual vocabularies whose training data is dominated by English and other high-resource languages. To establish the resulting cost, and to choose a vocabulary size for Neurograf on evidence rather than convention, we trained 21 byte-level BPE [20-22] tokenizers at vocabulary sizes from 8,000 to 2,000,000 on matched Romanian and English corpora, and measured how their behavior diverges.

**3.1 Setup.** We use rustbpe [15] with GPT-4-style pretokenization [23]. The tokenizer begins with 256 byte-level base symbols and reserves 18 special tokens. Each BPE merge operation introduces one additional learned vocabulary token. Thus, for a nominal vocabulary size N, the vocabulary contains 256 byte tokens, 18 special tokens, and N−274 learned BPE tokens. Training corpora are the Romanian and English portions of FineWeb [13], each truncated to exactly $10^9$ Python Unicode characters.

Matching on characters rather than bytes is deliberate. Romanian text carries a UTF-8 encoding overhead that English largely does not, so matching the corpora on bytes would expose the Romanian tokenizer to systematically less linguistic content. This would confound encoding overhead with the linguistic comparison. Matching on characters isolates the latter; Section 3.2 measures the former separately.

Rather than re-reading the corpora, we recover their exact byte counts from the merge trace. For each single-character symbol represented in the trace, the excess UTF-8 cost is $(b-1)*f$, where $b$ is its byte length and $f$ its frequency. Summing this excess over all such symbols and adding the character count recovers the total corpus byte count. The method validates on English, where the recovered premium is 1.0037, which is close to the 1.00 expected for near-ASCII text.

We additionally reconstructed the encoder from the rank-ordered merge list and verified it through round-trip encoding, vocabulary-membership checks, idempotence, and monotonicity across nested vocabularies. We also reconstructed the merge tree and found zero orphan tokens in either language.

The study tokenizers are not the production tokenizer. The sweep is trained on FineWeb, whereas the shipped 80k tokenizer is trained on RoSfera after normalization. Absolute values from the sweep therefore characterize the Romanian–English comparison rather than the production artefact itself. Section 3.9 reports measurements of the production tokenizer.

**3.2 The encoding tax.** Romanian requires approximately 1.035 bytes per character in the matched FineWeb sample, compared with 1.004 for English. Relative to a one-byte ASCII baseline, this corresponds to a 3.5% encoding overhead for Romanian. Of the Romanian corpus, 33,316,662 characters are encoded using two bytes (3.33%) and 830,318 using three bytes (0.083%). In English, the proportions are much smaller and are dominated by three-byte typographic characters such as curly quotes and dashes. The Romanian two-byte class is dominated by comma-below and circumflex diacritics, although it also contains accented characters from loanwords and foreign proper nouns; 3.33% is therefore an upper bound on the Romanian-diacritic rate itself.

This encoding overhead exists before BPE learning and is therefore independent of tokenizer vocabulary quality. It directly affects storage and any processing scheme whose cost scales with the UTF-8 byte stream. We report it separately for this reason and exclude it from the character-based comparisons that follow except where noted.

| | **Romanian** | **English** |
|---|---|---|
| **Corpus characters** | 1 Billion | 1 Billion |
| **Recovered bytes** | 1,034,977,298 | 1,003,651,037 |
| **Byte premium** | 1.0350 | 1.0037 |
| **Multibyte character occurrences** | 34,146,980 (3.41%) | 1,927,784 (0.19%) |

*Table 5: UTF-8 encoding overhead in matched Romanian and English FineWeb samples. Both corpora contain one billion Unicode characters. Byte premium denotes bytes per character. Romanian requires 1.035 bytes per character, compared with 1.004 for English, with most of the additional byte cost arising from two-byte characters containing Romanian diacritics.*

A note on the observed rate. The 3.33% two-byte-character rate is substantially below the diacritic rates observed in clean Romanian text: Chapter 2 finds values of approximately 6% in encyclopedic prose and 9% in literary prose. The deficit is consistent with the corpus-level diacritic degradation documented there: FineWeb Romanian, like much Romanian web text, contains a substantial fraction of material written without diacritics.

Separately, we find no Turkish-cedilla forms (U+015F, U+0163) in the analyzed FineWeb Romanian trace. The source is consistently comma-below encoded, so the legacy cedilla contamination encountered in some Romanian text collections is not apparent here.

**3.3 Compression, and where the languages cross.** At small vocabularies, the English tokenizer compresses its matched English corpus more efficiently than the Romanian tokenizer compresses its Romanian corpus. At 8,007 tokens, English covers 13.8% more characters per token. The gap then closes as vocabulary grows: 9.7% at 31k, 4.2% at 101k, 1.9% at 203k, and 0.5% at 406k. The ordering subsequently reverses. By 812k, Romanian covers marginally more characters per token than English, and this advantage persists to 1M.

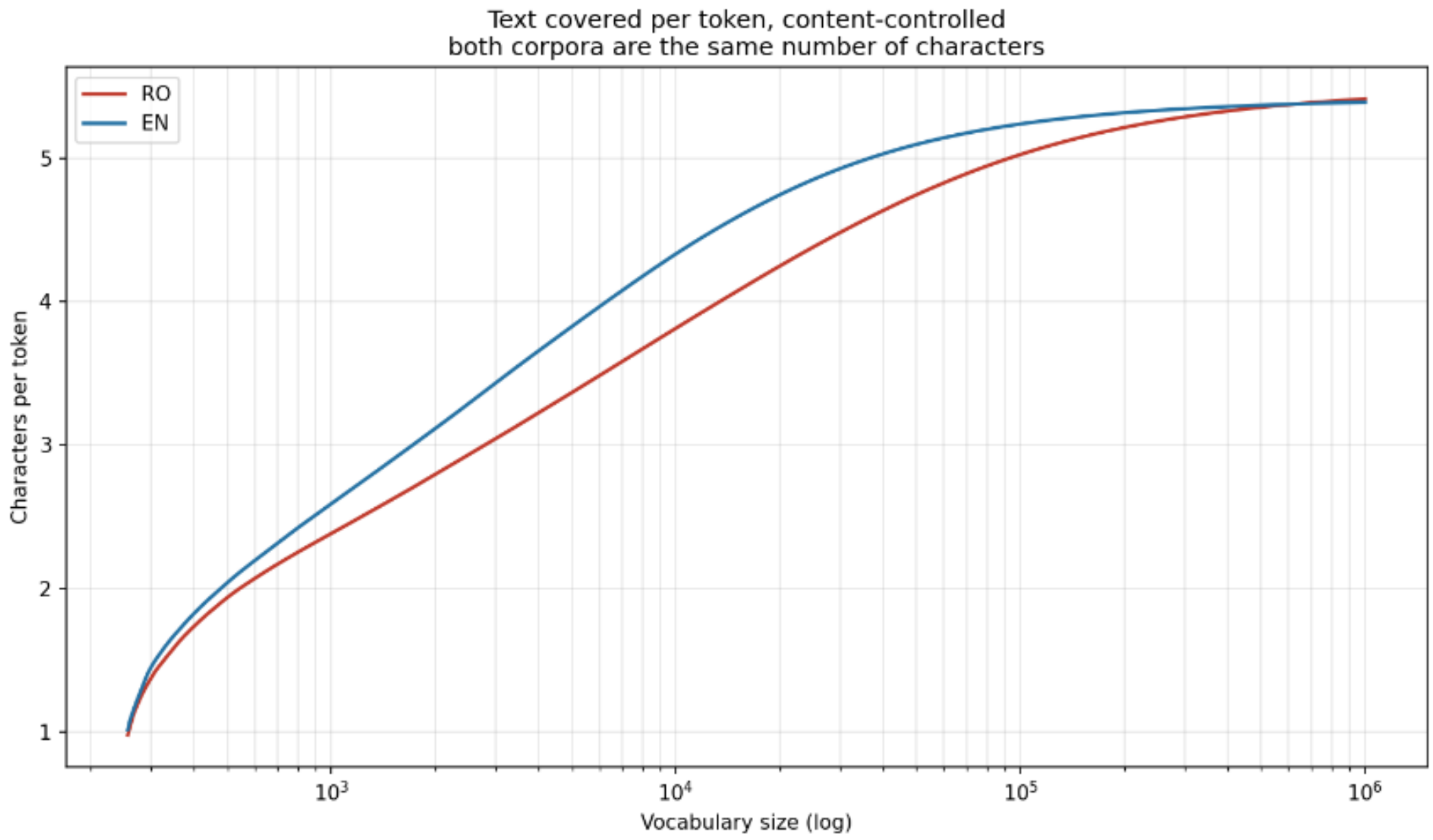

*Figure 4: Character-level compression as a function of vocabulary size. Characters per token are measured on content-controlled FineWeb samples containing exactly one billion characters in each language. English is more efficient across most of the sweep, but the gap narrows as vocabulary grows and eventually reverses.*

Both curves have the expected shape: compression rises steeply at small vocabulary sizes, where each additional merge captures a frequent sequence, and progressively flattens as newly added merges become rarer. The important quantity here is the gap between languages. English remains above Romanian across most of the sweep, but the separation closes at large vocabulary sizes and eventually reverses.

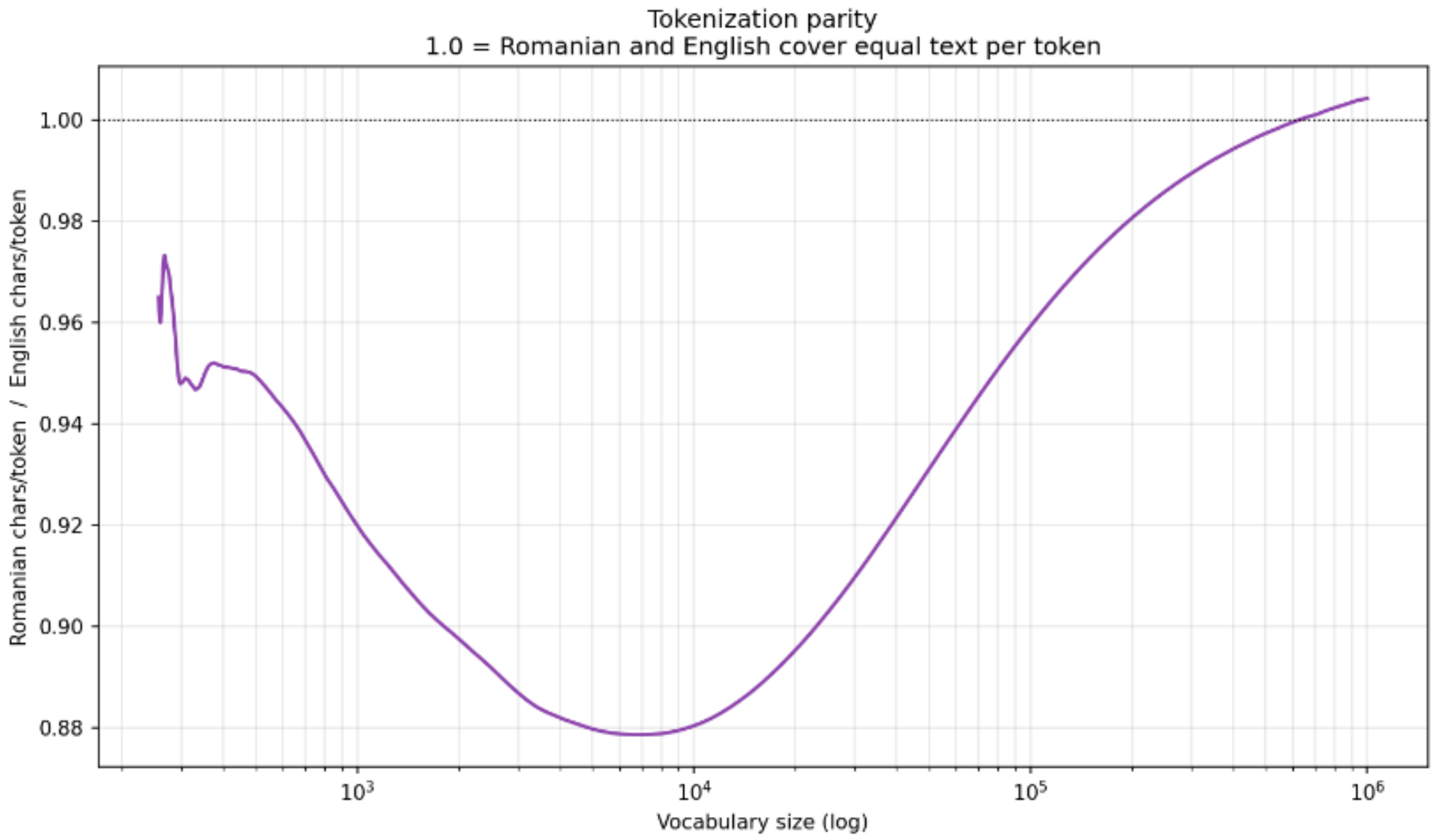

*Figure 5: The parity ratio, Romanian characters-per-token divided by English, where 1.0 denotes equal coverage. Measured in characters per token; see text for the bytes-per-token comparison, which places the crossover considerably earlier.*

The ratio isolates that behavior and makes the crossover directly visible. Romanian coverage reaches a minimum of approximately 87.9% of English in the low-vocabulary regime, then rises steadily across the remainder of the sweep and crosses parity near 570k. Thus the two curves do not merely approach one another within the observed range; their ordering reverses. A sweep terminating at conventional vocabulary sizes around 128k–256k would observe only a narrowing gap and could therefore suggest convergence. Extending the sweep beyond that range reveals the crossover.

The location of the crossover depends on the metric. In characters per token it falls near 570k. In bytes per token, which folds in the encoding tax of Section 3.2, Romanian overtakes English between 101k and 203k.

| Vocabulary | chars/token | bytes/token |
|---|---|---|
| 8,007 | EN +13.8% | EN +10.3% |
| 31,328 | EN +9.7% | EN +6.4% |
| 101,397 | EN +4.2% | EN +1.0% |
| 202,757 | EN +1.9% | RO +1.2% |
| 405,698 | EN +0.5% | RO +2.5% |
| 812,021 | RO +0.3% | RO +3.3% |

*Table 6: The crossover under both metrics. Text covered per token for Romanian relative to English at six vocabulary sizes, measured in characters per token and in bytes per token. Positive entries indicate which language covers more text per token and by how much. The two metrics differ by the UTF-8 encoding tax of Section 3.2, which places the bytes-per-token crossover between 101k and 203k, against 570k in characters per token.*

Both metrics are useful but answer different questions. Characters per token measures how much written content is represented by each token while controlling for UTF-8 encoding overhead. Bytes per token measures compression of the encoded byte stream and therefore includes that overhead. Notably, the vocabulary-composition crossover identified independently in Section 3.5 also falls near 200k, close to the bytes-per-token crossover and far below the character-based crossover.

The token counts underlying both metrics are recovered from the merge trace rather than by re-encoding the corpus. Under byte-level BPE, encoding begins with one token per byte, and each application of a merge rule replaces two adjacent tokens with one, reducing the count by exactly one. The number of tokens a vocabulary of size V produces on the training corpus is therefore the corpus byte count, minus the total number of merge applications, which is the summed frequency of the first V − 274 merges:

$$T(V) = B - \sum_{i=1}^{V-274} f_i \tag{3}$$

where T(V) is the number of tokens produced by a vocabulary of nominal size V, B is the corpus byte count, and $f_i$ is the training frequency of the *i*-th learned merge. The relation follows directly from byte-level BPE: encoding begins with one token per byte, and each application of a merge rule replaces two adjacent tokens with one, decreasing the sequence length by exactly one. Because the merge trace records the number of applications of each learned merge, the token count on the training corpus can be recovered exactly without re-encoding the corpus for every vocabulary size.

This is exact on the training corpus and requires only the rank-ordered merge list with frequencies, which is why the full 21-point sweep can be characterized without re-encoding a billion characters twenty-one times. The V − 274 offset is the 256 base byte tokens and 18 special tokens of Section 3.1, which are not merges.

**3.4 How much vocabulary each language needs.** BPE learns merges in decreasing order of frequency, so vocabulary size also determines the frequency of the rarest learned sequence. Reading the sweep in the opposite direction (asking how large a vocabulary each language requires before all sequences occurring at least $f$ times have been absorbed) separates the cost of high-frequency structure from that of the long tail.

| **Frequency floor** | **Romanian** | **English** | **RO / EN** |
|---|---|---|---|
| 100,000 | 1,017 | 1,066 | 0.95 |
| 10,000 | 5,454 | 5,120 | 1.07 |
| 1,000 | 28,343 | 21,460 | 1.32 |
| 100 | 118,194 | 73,391 | 1.61 |
| 10 | 437,474 | 250,255 | 1.75 |
| 5 | 665,954 | 376,878 | 1.77 |
| 2 | not reached | 710,965 | — |

*Table 7: Vocabulary size required to reach a given merge-frequency floor. For each frequency threshold, the table reports the nominal vocabulary size at which all merges occurring at least that often in the training corpus have been learned. Both tokenizers are trained on one billion characters, making merge frequencies directly comparable. The Romanian trace does not reach frequency 2 within the 2M-vocabulary sweep.*

The ratio increases systematically as the frequency floor is lowered and approaches approximately 1.8 over the lowest thresholds reached by both languages. At the highest frequency floor the requirements are nearly identical, with Romanian marginally smaller: covering all merges occurring at least 100,000 times requires a nominal vocabulary of 1,017 in Romanian and 1,066 in English. The divergence emerges progressively in the tail. At a floor of 1,000, Romanian requires approximately one third more vocabulary; at 100, approximately 60% more; and at frequencies 10 and 5, approximately 75–80% more.

The additional vocabulary requirement therefore lies primarily in the low-frequency tail rather than in the highest-frequency material. Romanian morphology provides a natural explanation: nouns, adjectives and verbs occur in multiple inflected surface forms, distributing frequency across a larger inventory of individually rarer forms. The composition analysis of Section 3.5 provides independent evidence for this interpretation. Quantitatively, reaching comparable merge-frequency depth requires roughly 1.8 times as much vocabulary in Romanian at the lowest shared thresholds measured here.

The asymmetry continues at the bottom of the table. English reaches merges of frequency 2 at a nominal vocabulary of approximately 711k, whereas the Romanian trace does not reach that floor anywhere within the 2M sweep. Romanian therefore retains a substantially larger tail of repeated low-frequency material. Section 3.6 examines the saturation behavior associated with this difference.

**3.5 What the vocabularies contain.** Compression is a summary statistic: it describes how much text a token covers on average, but not what linguistic material the vocabulary has learned. To inspect the mechanism behind the compression behavior of Section 3.3, we classify learned material as whole words, stems, prefixes, suffixes, word-internal fragments, or alphanumeric strings.

Classification is performed against a fixed 800k reference vocabulary rather than independently against each tokenizer's own vocabulary. This keeps the linguistic reference set constant across the sweep, so changes in category share reflect which material has been learned by a given vocabulary size rather than merely the number of available token slots.

At small vocabularies, Romanian contains substantially fewer whole-word tokens than English, a deficit of 8.3 percentage points at 8k, and correspondingly more stem and fragment material. As vocabulary grows, the whole-word deficit narrows, reaches parity near 200k, and then reverses. At the 800k reference point, whole words account for 54.8% of Romanian linguistic material against 46.9% for English, while Romanian contains less word-internal fragment material.

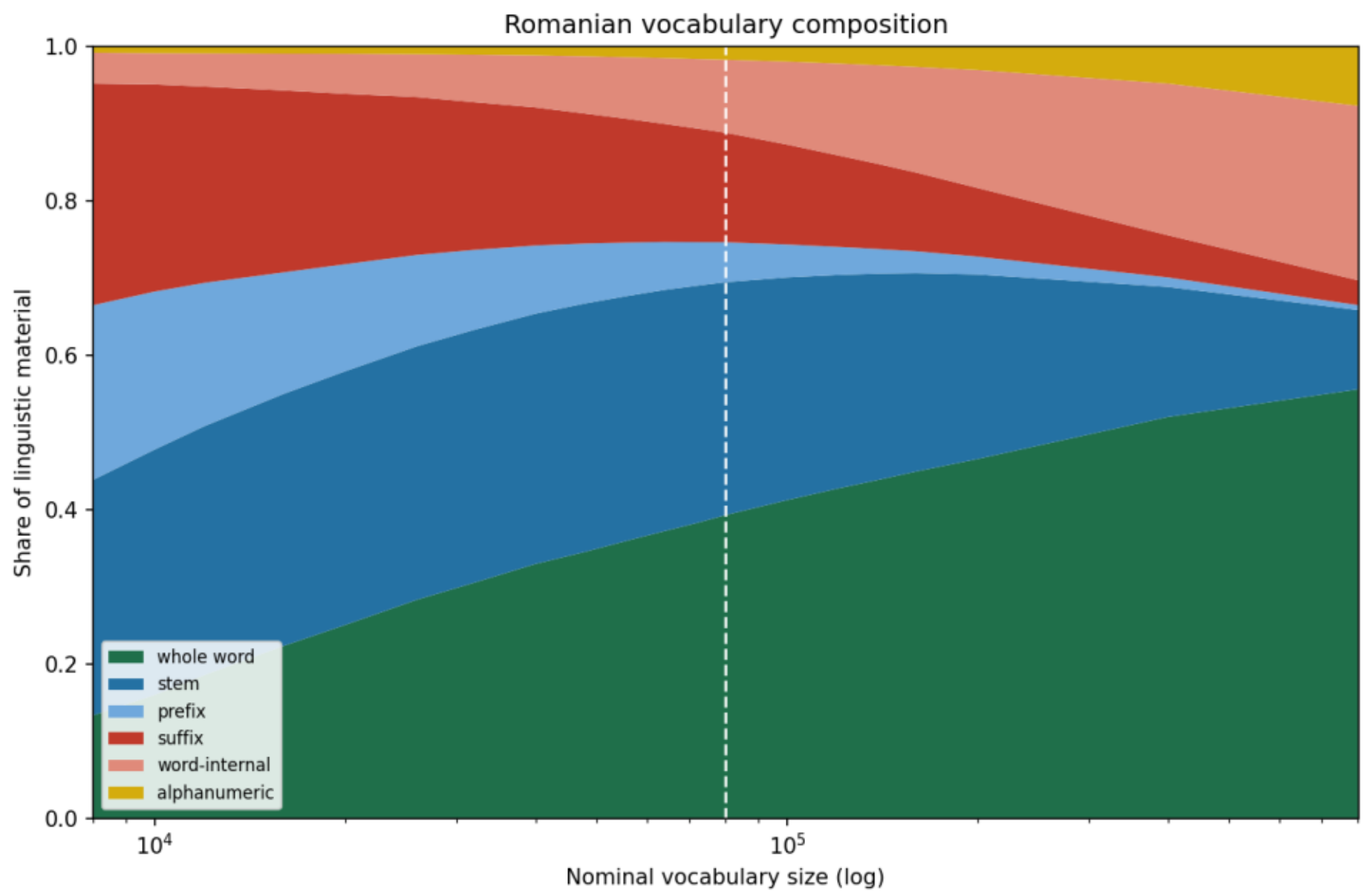

*Figure 6: Composition of the Romanian learned vocabulary as a function of nominal vocabulary size. Tokens are classified against a fixed 800k reference vocabulary as whole words, stems, prefixes, suffixes, word-internal fragments, or alphanumeric material. Using a fixed reference prevents category counts from increasing trivially with vocabulary size. The white dashed line in the middle represents our 80k Neurograf production tokenizer.*

Figure 6 shows the Romanian vocabulary composition in absolute terms. Whole-word material rises steadily from roughly 40% at the smallest vocabulary size to 54.8% at the 800k reference point, while stems also gain importance over the lower and intermediate part of the sweep. At the same time, prefixes, suffixes and especially word-internal fragments decline as a share of the learned vocabulary. The overall trend is therefore a gradual redistribution from subword fragments toward increasingly complete lexical units as vocabulary capacity grows.

This shift is not uniform across the sweep. At small vocabulary sizes, the tokenizer must reuse a relatively limited set of fragments to represent many different words, so stems, affixes and word-internal pieces occupy a larger fraction of the available vocabulary. As the vocabulary expands, additional merges increasingly permit frequent Romanian words to be represented as complete units rather than compositions of smaller fragments. The growth of the whole-word category therefore provides a complementary view of the compression results discussed above: larger vocabularies do not merely reduce token count, but also change the linguistic granularity of the representations being learned. Figure 7 makes the corresponding Romanian–English difference explicit.

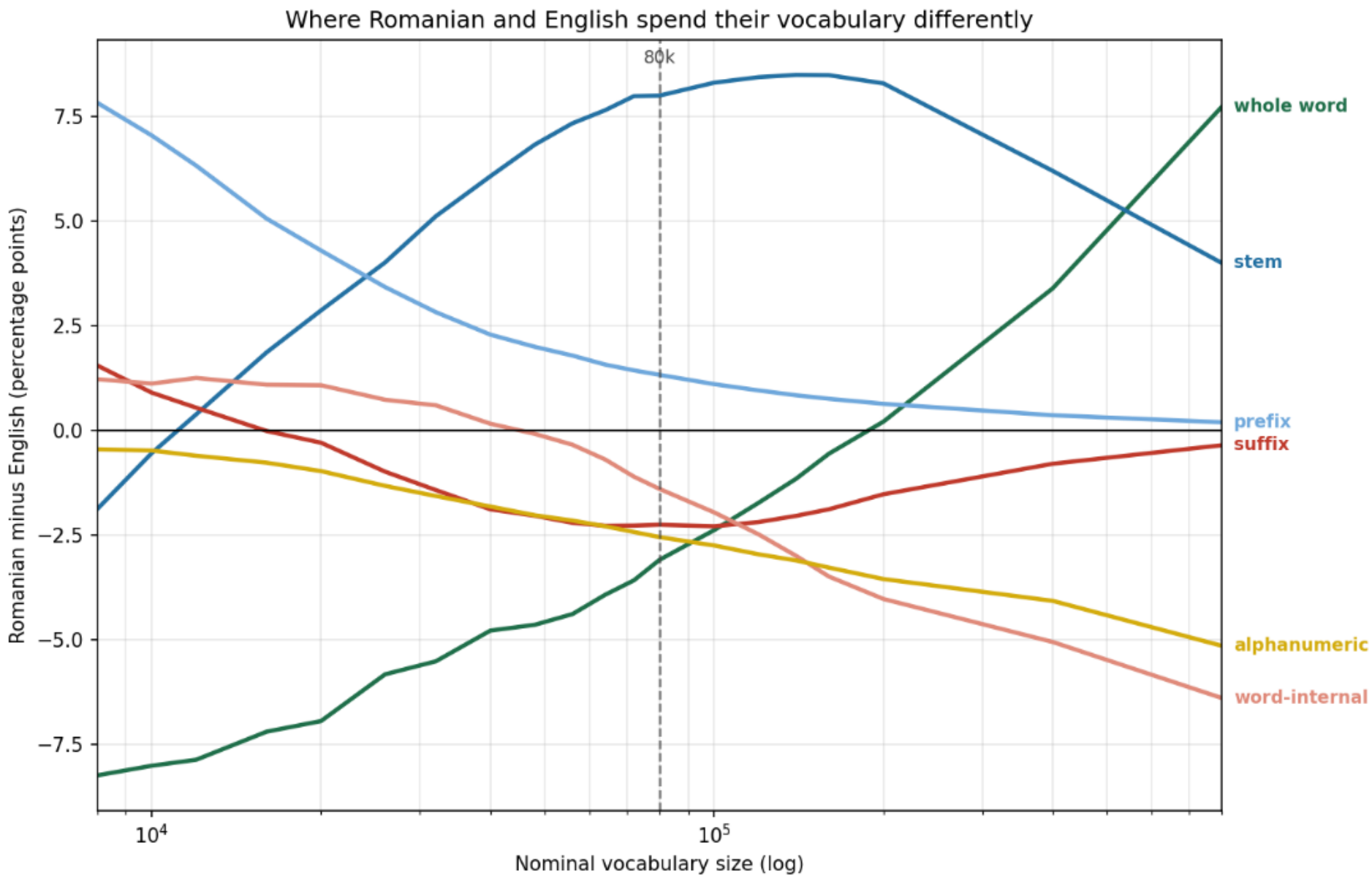

*Figure 7: Difference in vocabulary composition between Romanian and English, expressed in percentage points, with the 80k operating point marked. Positive values indicate a larger Romanian share. The whole-word difference changes sign near 200k, while the excess Romanian stem share decreases over the same range.*

The whole-word curve provides the clearest correspondence with the compression results. Romanian begins 8.3 percentage points below English, rises steadily with vocabulary size, and crosses zero near 200k, close to the bytes-per-token crossover identified independently in Section 3.3. The stem difference moves in the opposite direction: Romanian contains substantially more stem material at small vocabularies, but the excess declines as more complete forms enter the vocabulary.

Vocabulary composition therefore changes in parallel with compression. In the low-vocabulary regime, where Romanian compresses less efficiently, it also contains fewer whole words and more fragmented linguistic material. At larger vocabularies both relationships reverse. The proximity of the independently measured crossovers suggests that the compression gap is closely associated with how completely Romanian surface forms are represented in the vocabulary.

Two negative results are also informative. First, Romanian does not exhibit a larger suffix inventory in this classification: English contains more suffix tokens in absolute terms (27,974 versus 25,333). The additional Romanian vocabulary requirement therefore does not appear simply as a larger set of learned endings. Instead, the stronger difference emerges in the transition from stems and fragments toward complete word forms. Second, the approximately five-point alphanumeric difference primarily reflects alphanumeric noise in the English web sample and should not be interpreted as a linguistic property of either language.

Regarding affix discovery order, all fifteen Romanian nominal-inflection and enclitic-article affixes tracked in this analysis (*ul, ului, ii, ilor, ele, elor, lui, lor, ile, ea, ei, ale, uri, urile, urilor*) are already present at 8k. The tracked derivational affixes appear later and much less uniformly: *esc* at 32k, *ește* at 40k, *ase* at 56k, *ist* at 100k, *ător* at 120k, *ime* at 400k, and *izare* only at 2M, while *ibil*, *ificare* and *asem* do not appear within the sweep. Within this sample, inflectional morphology is therefore captured substantially earlier than the lower-frequency derivational material.

**3.6 Saturation.** A BPE vocabulary cannot continue gaining useful new material indefinitely on a fixed training corpus. As vocabulary size increases, newly learned merges become progressively rarer, and eventually additional merge operations may produce byte strings already representable by existing vocabulary entries. At that point, increasing nominal vocabulary size adds increasing redundancy rather than genuinely new string coverage. We examine this regime using two diagnostics, merge ambiguity and merge depth, which behave differently between Romanian and English.

The first diagnostic is merge ambiguity: the fraction of learned tokens whose byte string admits more than one valid decomposition into other vocabulary members. As later tokens become increasingly compositional with respect to material already present in the vocabulary, this fraction rises. We therefore use increasing ambiguity as an indicator of growing representational redundancy. The second is merge depth: the number of successive merge operations required to construct a token from base symbols. Greater depth indicates that later vocabulary entries are increasingly built through long chains of previously learned material.

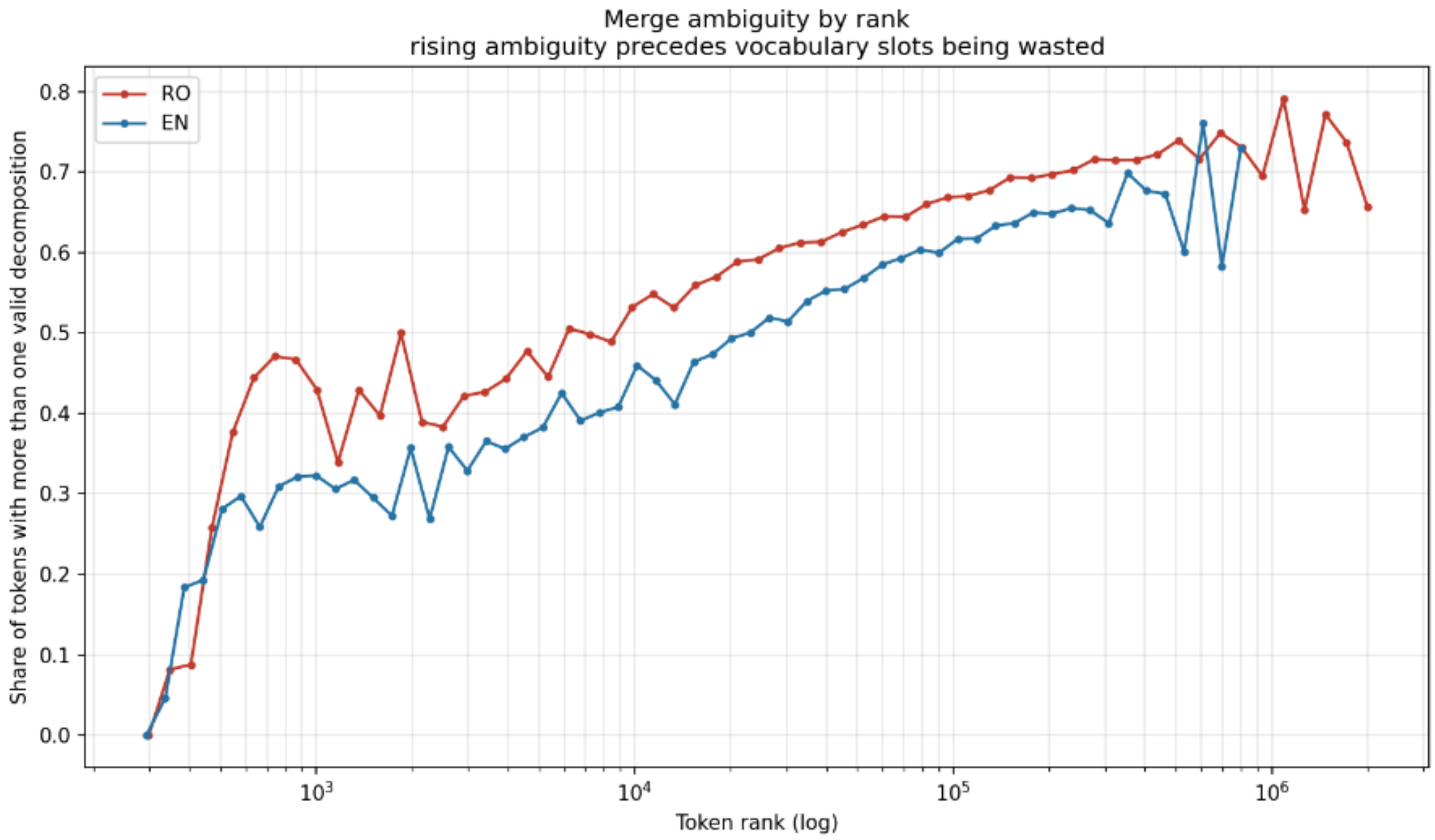

*Figure 8: Merge ambiguity as a function of token rank. The vertical axis gives the share of tokens admitting more than one valid decomposition into other vocabulary members. Increasing ambiguity indicates growing overlap with material already representable by the vocabulary. Romanian lies above English across much of the sweep. The English series terminates near 800k, whereas Romanian continues to 2M; estimates become noisy at the largest ranks because the final bins contain fewer tokens.*

Ambiguity rises with token rank in both languages, as later and longer entries admit more possible decompositions. Across most of the middle and upper sweep, Romanian exhibits a higher ambiguity fraction than English. The English trace also terminates substantially earlier because its learned merge sequence ceases to produce new distinct vocabulary strings before the Romanian sequence does.

Merge depth shows a related divergence. The two languages track one another closely through approximately 10k tokens. Beyond that point, mean Romanian merge depth rises above English and continues increasing across the remainder of the available sweep, while the English curve flattens before terminating.

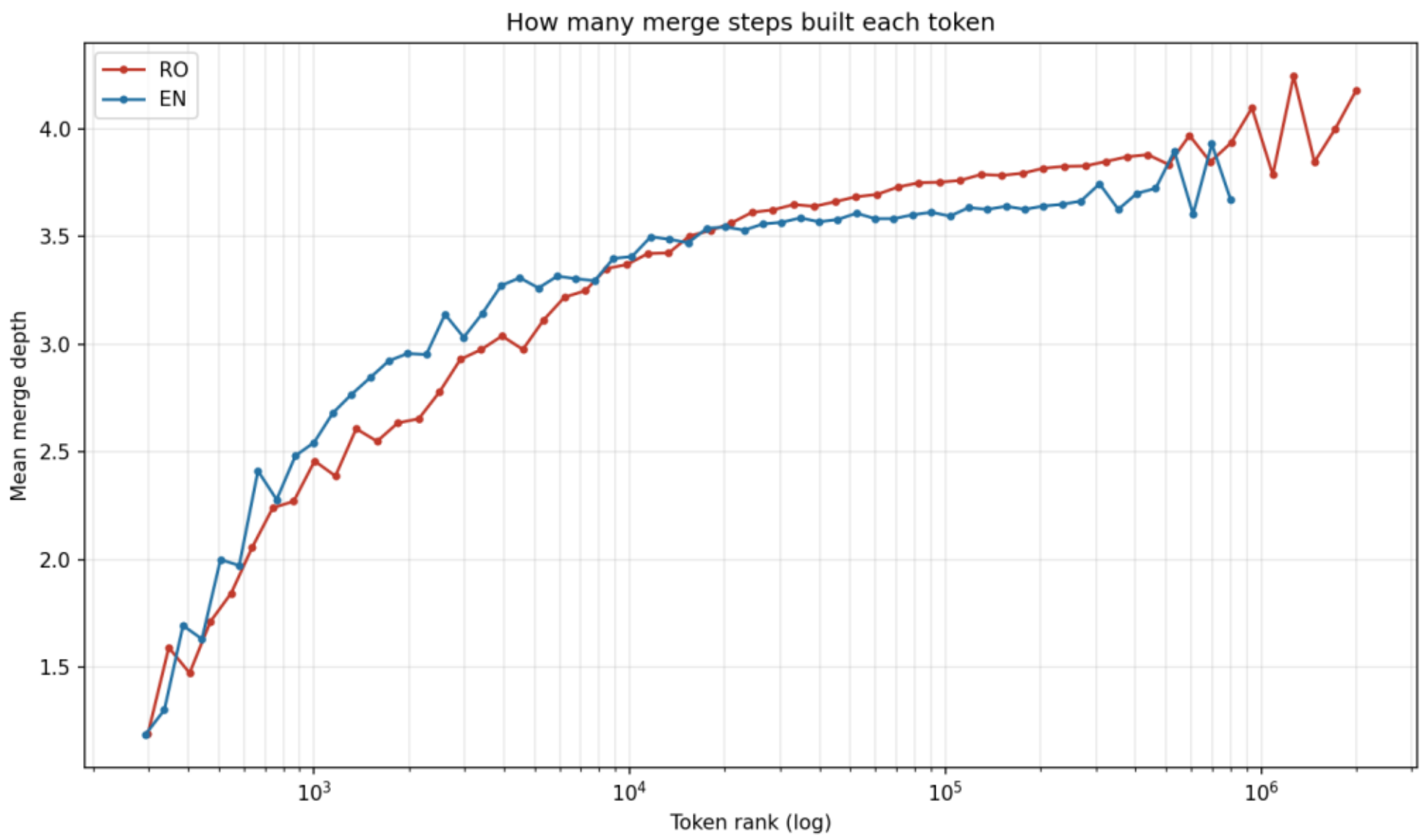

*Figure 9: Mean number of merge steps used to construct each token. Romanian exceeds English across the upper half of the sweep. The English series terminates at 800k, where its merge material is exhausted; the Romanian series continues to 2M. Both become noisy at the highest ranks, where each bin contains fewer tokens.*

The English training trace reaches the low-frequency end of its useful merge inventory substantially earlier than the Romanian trace. Merges occurring only twice in the one-billion-character corpus are reached at approximately 711k in English but are not reached anywhere within the Romanian 2M sweep. When forced to a nominal vocabulary of 2M, the English run contains only 1,394,882 distinct decoded strings, leaving approximately 605,000 nominal entries whose decoded byte strings duplicate entries already present. We therefore exclude this degenerate 2M English tokenizer from subsequent analysis. The Romanian 2M vocabulary remains complete in the sense that all nominal entries decode to distinct strings.

At their respective upper usable endpoints, ambiguity reaches 71.2% for Romanian at 2M and 65.1% for English at 800k, while mean merge depth reaches 3.96 and 3.70 respectively. These endpoint values are not directly size-matched, but together with the frequency-floor analysis they show that the Romanian training trace continues to expose repeated mergeable material at vocabulary sizes where the English trace has already become degenerate. At one billion training characters per language, the two BPE processes therefore reach their practical saturation regimes at markedly different nominal vocabulary sizes.

We do not interpret this difference as direct evidence of Romanian morphological productivity. As Section 3.10 documents, the Romanian FineWeb sample contains non-Romanian material that can produce a similar signature: a vocabulary trained on a more heterogeneous corpus has more distinct repeated material available for reasons unrelated to Romanian morphology. We therefore leave the cause of the saturation difference unresolved.

### 3.7 The asymmetry of the wrong tokenizer.

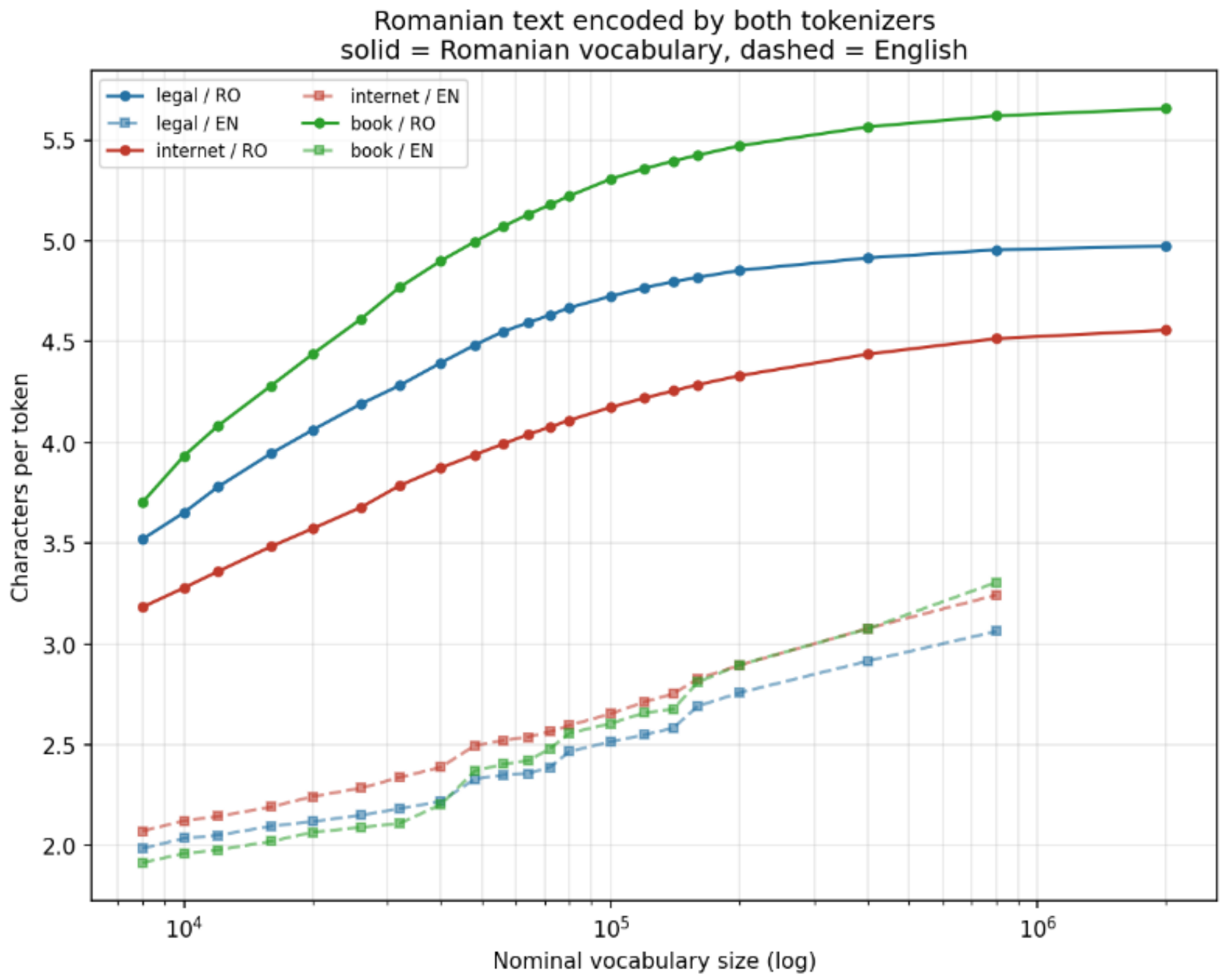

*Figure 10: Romanian text encoded with Romanian and English vocabularies. Characters per token are shown for three Romanian registers — internet, legal, and books — across vocabulary size. Solid curves use the Romanian vocabulary and dashed curves the English vocabulary. Native Romanian tokenization improves strongly with vocabulary size and differs by register, whereas English-vocabulary encoding remains substantially less efficient.*

A natural objection to Section 3.3 is that each tokenizer is evaluated on the language on which it was trained. If the cost of using a mismatched vocabulary were symmetric between Romanian and English, the compression crossover could therefore reflect little more than corpus specialization.

We test this directly by encoding each language with both vocabularies. For every vocabulary size in the sweep, we encode Romanian evaluation text under both the Romanian and English vocabularies. Romanian is evaluated separately on internet, legal, and literary registers. We also encode English evaluation text under both vocabularies. For each case, the mismatched encoding rate is divided by the rate achieved by the vocabulary trained for that language, yielding a relative efficiency measure in which 100% denotes parity with the native vocabulary. The resulting penalty is strongly asymmetric.

Two features are immediately visible in the raw encoding rates. First, the English-vocabulary curves remain well below the Romanian-vocabulary curves throughout the sweep: even near 800k, the English vocabulary encodes Romanian at roughly 3 characters per token, whereas the Romanian vocabulary reaches approximately 4.5–5.6 depending on register. Second, the native Romanian curves separate strongly by register, while the English-vocabulary curves remain comparatively clustered. The Romanian vocabulary therefore captures register-specific differences in compressibility that are largely absent under the mismatched English vocabulary.

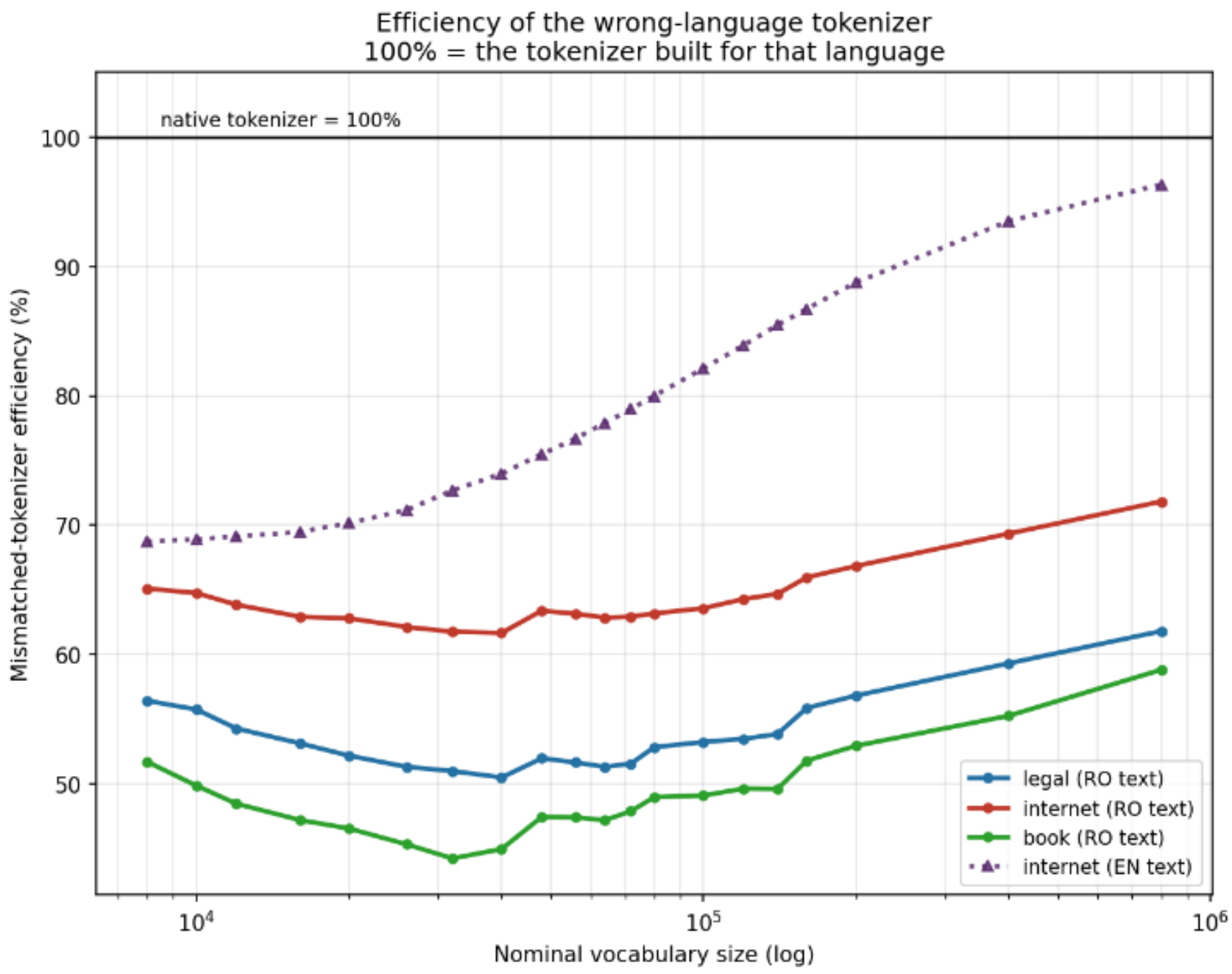

*Figure 11: Relative efficiency of the mismatched tokenizer. Encoding efficiency is expressed as the mismatched characters-per-token rate divided by the native-tokenizer rate for the same text, so 100% denotes parity. The purple curve shows English text encoded with the Romanian vocabulary; the other three curves show Romanian internet, legal, and literary text encoded with the English vocabulary.*

Expressing the mismatched rate as a fraction of the native rate makes the asymmetry directly comparable across the two directions.

The asymmetry is large. A Romanian vocabulary encodes English increasingly efficiently as vocabulary size grows, rising from 68.7% of native English efficiency at 8k to 96.3% at 800k. The reverse direction remains substantially worse: at the largest tested vocabulary, English-vocabulary efficiency reaches only 71.8% for Romanian internet text, 61.8% for legal text, and 58.8% for books. Thus, the Romanian vocabulary approaches parity on English, whereas the English vocabulary remains inefficient on Romanian even at large vocabulary sizes.

Two mechanisms plausibly contribute to this asymmetry. First, the English FineWeb vocabulary contains very little non-ASCII material at small sizes (only 34 of 7,726 learned entries at 8k) so Romanian diacritics are poorly represented and often remain decomposed into byte-level components. Second, the Romanian FineWeb sample contains a non-negligible amount of English material (Section 3.10), allowing the Romanian vocabulary to learn English sequences during training. The first mechanism reflects the orthographic mismatch between the datasets; the second reflects corpus contamination. Their contributions cannot be separated in the present experiment, so the measured asymmetry should be regarded as an upper bound on any purely language-driven effect.

Notably, all three Romanian mismatch-efficiency curves reach their minimum between approximately 32k and 40k. This lies within the vocabulary-size range historically common in large language models, including Llama 2 at 32k [24] and GPT-2 at approximately 50k [22]. For the Romanian–English corpus pair studied here, this conventional operating range therefore falls close to the point of maximum relative mismatch penalty. This minimum should not be confused with the minimum of the native-language parity curve in Figure 5: the two quantities measure different effects. Figure 5 compares Romanian and English when each uses its own vocabulary, whereas Figure 11 compares matched and mismatched vocabularies on the same language.

**3.8 Choosing 80k.** Sections 3.3 through 3.7 point in the same practical direction: increasing vocabulary size continues to improve Romanian tokenization across the range examined. The countervailing cost is model capacity. Because Neurograf uses untied input and output embedding matrices, a vocabulary of size V contributes:

$$Param = 2 * V * d_{model} \quad (4)$$

parameters through the input embedding and output projection matrices.

This cost is highly uneven across the model family. At 80k, the input and output embedding matrices contain approximately 123M, 246M, 369M, 492M and 614M parameters in the d6 through d30 models, corresponding to roughly 77%, 44%, 25%, 15% and 10% of their nominal parameter counts, respectively. In the smallest model, more than three quarters of all parameters are therefore allocated to vocabulary representation, leaving fewer than 40M parameters for the transformer body; in the largest model, the corresponding fraction is approximately one tenth. Tokenization gains apply across the family, but their relative parameter cost is consequently concentrated in the smaller models.

This capacity trade-off determines the operating point. A vocabulary in the low hundreds of thousands would bring Romanian close to English in bytes per token (Section 3.3) and move well beyond the mismatch-efficiency minimum of Section 3.7, but would impose a prohibitive vocabulary cost on the 160M model. At 200k, its embedding matrices would dominate the model almost completely. Conversely, the conventional 32k–50k range lies close to the maximum mismatch penalty observed for Romanian in Section 3.7.

We therefore select 80k as a family-level compromise: large enough to move beyond the low-vocabulary efficiency trough and capture a substantial fraction of the gains available below 100k, while still leaving the smallest model with a non-trivial transformer body. The choice is specific to a family spanning 160M to 6B parameters; a project targeting only models of several billion parameters could reasonably support a substantially larger vocabulary.

**3.9 The production tokenizer.** The production Neurograf tokenizer is an 80k byte-level BPE vocabulary trained directly on RoSfera using the same rustbpe implementation and GPT-4-style pre-tokenization as the study sweep. Training used 100 of the 1,000 pre-training master shards, corresponding to approximately 10% of the corpus, or 63 GB of UTF-8 text, after the diacritic normalization described in Section 2.2. Because the master shards were constructed to preserve near-identical source proportions, this 100-shard sample approximates the composition of the full RoSfera corpus rather than selecting particular source classes.

Measured on the full RoSfera corpus, the production tokenizer achieves 4.63 characters and 4.85 UTF-8 bytes per token. This is approximately 5% below the 5.10 bytes per token predicted at 80k by the FineWeb sweep, consistent with the production tokenizer being trained on a substantially different mixture of web, institutional, and literary text.

To illustrate the resulting segmentation, we encode a representative Romanian sentence with the production tokenizer and compare it with the Meta Llama 3 [4] tokenizer:

**Input text:** “Limba română este o limbă de origine latină (romanică) vorbită ca limbă maternă de aproximativ 24 de milioane de oameni, în special în România și Republica Moldova, unde are și statutul de limbă oficială. Ea a apărut din transformarea limbii latine vorbite în antichitate în partea de nord a Peninsulei Balcanice și în jurul Dunării.”

**Neurograf tokenizer:** “Limba ␣română ␣este ␣o ␣limbă ␣de ␣origine ␣latină ␣( rom anică ) ␣vorbită ␣ca ␣limbă ␣maternă ␣de ␣aproximativ ␣ 24 ␣de ␣milioane ␣de ␣oameni , ␣în ␣special ␣în ␣România ␣și ␣Republica ␣Moldova , ␣unde ␣are ␣și ␣statutul ␣de ␣limbă ␣oficială . ␣Ea ␣a ␣apărut ␣din ␣transformarea ␣limbii ␣latine ␣vorbite ␣în ␣antichitate ␣în ␣partea ␣de ␣nord ␣a ␣Peninsul ei ␣Balcani ce ␣și ␣în ␣jurul ␣Dunării.” **Total: 64 tokens.**

**Meta Llama 3 8B tokenizer:** “Lim ba ␣rom ân ă ␣este ␣o ␣limb ă ␣de ␣orig ine ␣latin ă ␣( rom anică ) ␣vor bit ă ␣ca ␣limb ă ␣mat ern ă ␣de ␣aproxim ativ ␣ 24 ␣de ␣mil io ane ␣de ␣o amen i , ␣în ␣special ␣în ␣Rom â nia ␣și ␣Republic a ␣Mold ova , ␣unde ␣are ␣și ␣stat ut ul ␣de ␣limb ă ␣oficial ă . ␣Ea ␣a ␣ap ăr ut ␣din ␣transform area ␣limb ii ␣lat ine ␣vor bite ␣în ␣ant ich itate ␣în ␣part ea ␣de ␣nord ␣a ␣Pen ins ul ei ␣Bal can ice ␣și ␣în ␣jur ul ␣Dun ăr i i.” **Total: 106 tokens.**

The sentence-level comparison is illustrative rather than evaluative; aggregate tokenizer efficiency is established by the controlled measurements of Sections 3.3 and 3.7 rather than by this individual example.

Only three lexical items in this example are split under the Neurograf vocabulary: *romanică → rom + anică*, *Peninsulei → Peninsul + ei*, and *Balcanice → Balcani + ce*. All three involve derivational or relatively low-frequency lexical material, consistent with the late appearance of such forms in the vocabulary analysis of Section 3.5. By contrast, common inflected forms such as *română, vorbită, maternă, statutul, limbii, oficială, apărut*, and *jurul* are represented as single tokens.

The contrast with the Llama 3 vocabulary is substantial in this example. The same Romanian sentence requires 106 tokens under Llama 3 versus 64 under Neurograf, an increase of approximately 66%. Fragmentation is not limited to low-frequency derived forms: common Romanian words are also split, for example *limbă → limb + ă* and *oameni → o + amen + i*. Diacritic-bearing words are frequently segmented around the marked characters, as in *România → Rom + â + nia* and *Dunării → Dun + ăr + i + i*. This example provides a concrete illustration of the mismatch quantified systematically in Section 3.7: a vocabulary trained predominantly on other languages allocates fewer learned units to Romanian orthographic and morphological patterns, increasing segmentation of otherwise common Romanian forms.

**3.10 Limitations.** Three features of the experimental design limit the generality of the sweep: contamination within the training corpora, differences between the study and production tokenizers, and reliance on a single Romanian–English corpus pair.

- **Corpus contamination.** The high-rank vocabulary tail contains material in Cyrillic as well as German, Hungarian, medical Latin, corporate names, and non-linguistic strings. FineWeb language identification therefore admits non-target material into both the Romanian and English subsets. This may affect several results: it could contribute to the extended Romanian saturation behavior of Section 3.6, partly explain the mismatch asymmetry of Section 3.7, and influence the crossover measured in Section 3.3 by making the Romanian vocabulary partially multilingual. Contamination appears limited at 80k but becomes increasingly visible above approximately 200k, close to the bytes-per-token crossover. A more detailed analysis by Unicode script and language identification, followed by repetition of the sweep on more strictly filtered corpora, would help quantify these effects; we have not performed that analysis here.

- **Study and production tokenizers differ.** The controlled sweep is trained on one billion characters of FineWeb per language, whereas the production tokenizer is trained on a much larger sample of RoSfera after normalization. The two settings differ in corpus composition, scale, register mixture, and preprocessing. Section 3.9 therefore reports production-tokenizer measurements separately. Results from Sections 3.2–3.8 should be interpreted primarily as properties of the controlled Romanian–English comparison rather than as direct measurements of the shipped Neurograf tokenizer.
- **A single corpus pair.** The sweep compares one Romanian FineWeb sample against one English FineWeb sample. The crossover location, mismatch-efficiency trough, saturation behavior, and asymmetry are therefore measurements of this specific corpus pair. We have not tested whether they reproduce under a different extraction pipeline, web snapshot, or corpus family. Replication would be required before treating either the exact crossover locations or the qualitative behavior of the curves as language-general properties.

Within these limits, the sweep establishes a consistent empirical picture. On the matched FineWeb samples studied here, Romanian requires substantially more vocabulary than English to represent its low-frequency material, progressively closes the compression gap as vocabulary grows, and exhibits a strongly asymmetric penalty under an English-trained vocabulary. These observations motivated the 80k production vocabulary, chosen as a compromise between Romanian tokenization efficiency and the parameter cost imposed on the smallest Neurograf models.

## 4. Pretraining

We trained five decoder-only base models ranging from 160M to 6B parameters on an identical 131-billion-token training stream drawn from the RoSfera curriculum described in Section 2.4. The experiment was deliberately designed as a fixed-data study rather than a compute-optimal scaling exercise: every model encounters the same training material in the same order, allowing differences in learning behavior and final validation loss to be examined primarily as a function of model capacity.

All models use a context length of 2,048 tokens and the same 80k Neurograf tokenizer. Model scales are increased systematically from depth 6 to depth 30 while retaining a fixed architectural aspect ratio of 128. This produces a five-model family spanning approximately two orders of magnitude in transformer capacity while preserving a common training pipeline, tokenizer, data order, and total token budget.

### 4.1 Architecture and implementation.

| Model | Depth | d_model | Parameters | Tokens/param | Final val BPB |
|---|---|---|---|---|---|
| 160M | 6 | 768 | 165,642,240 | 791 | 0.8867 |
| 560M | 12 | 1,536 | 586,088,448 | 224 | 0.7427 |
| 1.5B | 18 | 2,304 | 1,516,142,592 | 86 | 0.6657 |
| 3.2B | 24 | 3,072 | 3,210,608,640 | 41 | 0.6195 |
| 6B | 30 | 3,840 | 5,924,290,560 | 22 | 0.5876 |

*Table 8: Neurograf base-model family. All five models use the same 2,048-token context length, 80k tokenizer, and 131-billion-token training stream. Model width scales with depth at a fixed aspect ratio of 128. Tokens per parameter therefore decrease strongly with model size, from 791 for the 160M model to 22 for the 6B model.*

The implementation is derived from Andrej Karpathy's nanochat [15] base-pretraining code. We retained the reference decoder-only model architecture and training pipeline, while modifying the components required for the Neurograf experiment. The original data-ingestion path was replaced with direct loading of locally stored RoSfera text shards, the Neurograf tokenizer was integrated into the training pipeline, and model configurations were defined for depths 6, 12, 18, 24 and 30 with a fixed aspect ratio of 128. Training was adapted for execution under Slurm in a self-contained Conda environment, allowing long runs to be divided into restartable jobs while preserving model, optimizer and data-stream state across allocations.

The resulting models have widths of 768, 1,536, 2,304, 3,072 and 3,840 dimensions, respectively. Input token embeddings and the output language-model projection are represented by separate matrices rather than tied weights;

consequently, the parameter cost associated with the 80k vocabulary becomes proportionally much larger in the smaller models, as discussed in Section 3.8.

Apart from the data-loading, tokenizer, execution-environment and scaling configurations described above, the base-pretraining implementation was intentionally kept close to nanochat. This choice was motivated by simplicity and auditability: the compact training code could be inspected end to end, modified without introducing a large framework-specific software layer, and used consistently across all five model scales.

The fixed-data design produces a deliberately broad range of training regimes. At 131 billion tokens, the models receive approximately 791, 224, 86, 41 and 22 training tokens per parameter from smallest to largest. The 6B model therefore lies close to the commonly used Chinchilla-scale allocation of roughly 20 tokens per parameter [26], whereas the smaller models are increasingly data-rich relative to their parameter count. The 160M model, at 791 tokens per parameter, receives approximately forty times the token exposure implied by that rule of thumb.

This over-training of the smaller models is intentional. Neurograf was designed in part to explore compact Romanian models intended for inexpensive local inference, where additional one-time training expenditure may be acceptable if it improves the capability of a model that remains cheap to deploy. Using the same 131-billion-token stream at every scale also provides a controlled basis for the fixed-data scaling analysis presented later in this section.

**4.2 Training configuration.** The global token batch was scaled with model size, while the number of optimization steps was adjusted inversely so that every model consumed the same total number of training tokens.

The d24 and d30 models used a global batch of 2,097,152 tokens and were trained for 62,500 optimization steps. For d18, the global batch was halved to 1,048,576 tokens and the run length doubled to 125,000 steps. The same pattern was continued for the smaller models: d12 used 524,288 tokens per optimization step for 250,000 steps, and d6 used 262,144 tokens per step for 500,000 steps. This geometric scaling preserves the total token budget while allowing the per-step workload to remain appropriate for each model size.

| **Model** | **Depth** | **Global token batch** | **Optimization steps** |
|---|---|---|---|
| 160M | 6 | 262,144 | 500,000 |
| 560M | 12 | 524,288 | 250,000 |
| 1.5B | 18 | 1,048,576 | 125,000 |
| 3.2B | 24 | 2,097,152 | 62,500 |
| 6B | 30 | 2,097,152 | 62,500 |

*Table 9: Training geometry of the five base models. Global token batch increases with model scale, while optimization-step count decreases correspondingly, preserving an approximately identical 131-billion-token training budget for every model.*

Training followed the nanochat base-pretraining optimization pipeline. The learning-rate schedule follows the classical *warmup–stable–decay* approach: an initial linear warmup was followed by a constant-learning-rate plateau and a final linear warmdown, with the learning-rate multiplier terminating at 0.05 rather than zero. The warmup duration scaled with global batch size: 746 steps for d6, 373 for d12, 187 for d18, and 100 steps for both d24 and d30. This corresponds to approximately 196–210 million training tokens across the family, keeping the amount of data seen during warmup approximately constant despite the different batch sizes. The precise optimizer implementation and parameter-group learning rates follow the nanochat reference training code [15, 25].

For the 6B model, the device batch size was one 2,048-token sequence per GPU per micro-step, with gradient accumulation over 256 micro-steps, producing the 2,097,152-token global batch. The d24 run used the same global batch configuration. Smaller models used proportionally reduced global batches as shown in Table 9. Long runs were executed through Slurm and therefore split across multiple cluster allocations rather than run as a single uninterrupted job. Checkpoints preserved the complete training state between allocations. In the 6B run, jobs were typically divided into segments of approximately 9,300 optimization steps, corresponding to roughly six days of cluster wall time per allocation. The complete 6B pre-training run required approximately 30.7 days of wall-clock time.

Validation loss was evaluated periodically on 524,288 tokens; for the 6B run this evaluation was performed every 250 optimization steps. Separate qualitative text-generation probes were sampled every 1,000 steps. These probes were useful primarily as qualitative indicators of emerging language behavior and were not used as the principal quantitative evaluation metric. The scaling and training-dynamics analyses below therefore use the recorded training loss and validation bits-per-byte measurements rather than subjective probe quality.

**4.3 Infrastructure and utilization.** All pre-training runs were performed on four NVIDIA H100 GPUs. Taking the aggregate BF16 dense peak of the four GPUs as approximately (3.96E15) FLOP/s, the complete five-model pre-training campaign consumed approximately 9.06 ZettaFLOP over 59.2 aggregate days of training wall-clock time, corresponding to 5,682 H100-hours.

| Model | Iteration time | Tokens/s | MFU |
|---|---|---|---|
| 160M (d6) | 0.395 s | 664,254 | 12.38% |
| 560M (d12) | 0.889 s | 589,428 | 48.13% |
| 1.5B (d18) | 4.685 s | 223,829 | 50.96% |
| 3.2B (d24) | 20.011 s | 104,801 | 51.92% |
| 6B (d30) | 46.365 s | 45,231 | 41.77% |

*Table 10: Representative training performance on four H100 GPUs. Iteration time, aggregate token throughput and model FLOP utilization remain approximately stable through each run. Throughput decreases monotonically with model size, whereas hardware utilization reaches approximately 50% for the intermediate models and is substantially lower at both extremes of the family.*

Runtime behavior was highly stable within each model scale. After initialization, iteration time, token throughput, and model FLOP utilization (MFU) remained approximately constant over most of each run, making the representative values in Table 10 useful descriptions of the five operating regimes.

Figure 12 shows the same behavior across model scale. Raw throughput decreases monotonically, from approximately 664k tokens/s for the 160M model to 45k tokens/s for the 6B model, but computational utilization is strongly non-monotonic. The 160M model reaches only 12.4% MFU, rises sharply to 48.1% at 560M, and remains near 51% for the 1.5B and 3.2B models before declining to 41.8% for the 6B model.

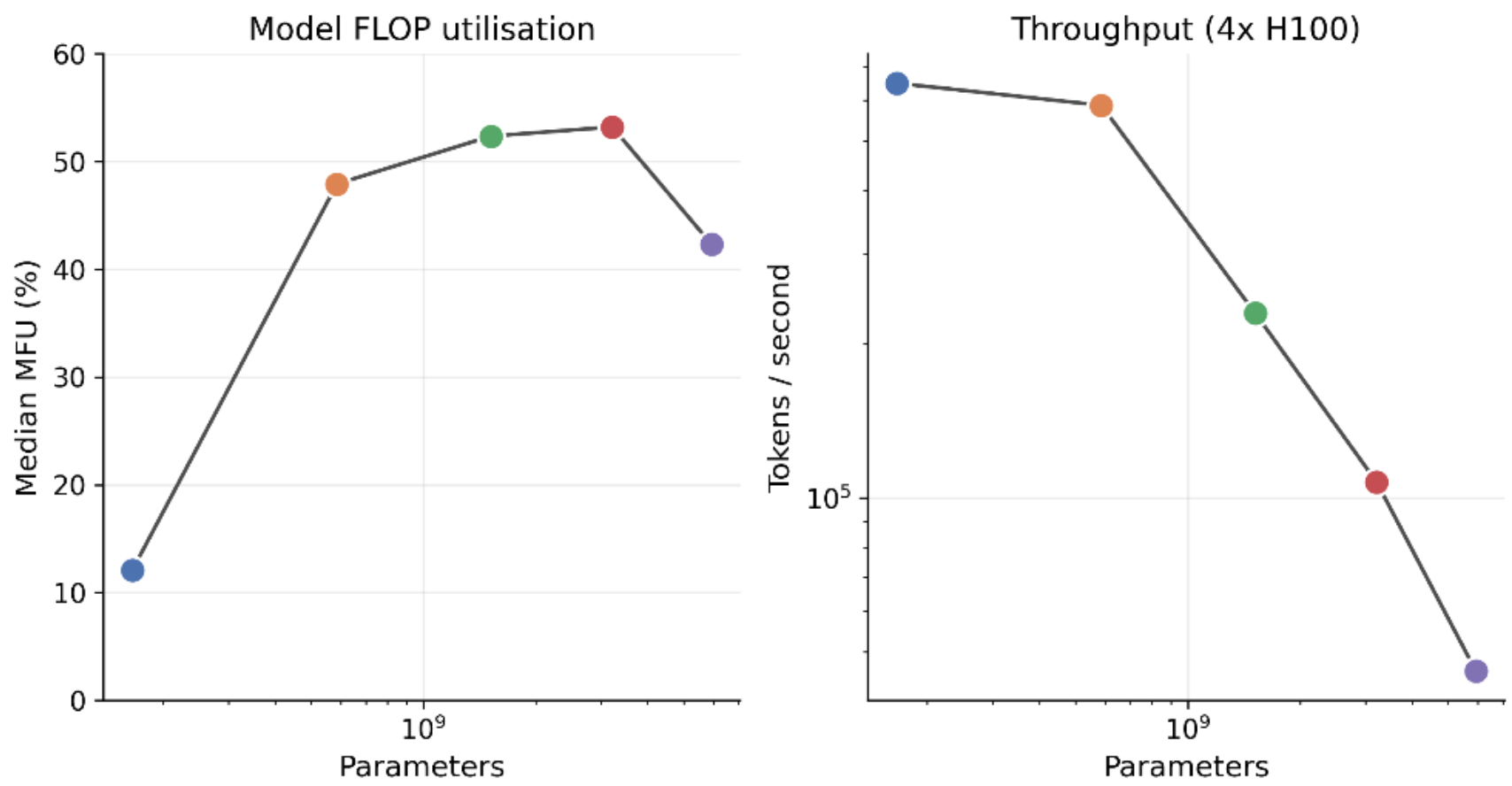

*Figure 12: Training utilization across model scale on four H100 GPUs. Left: representative model FLOP utilization (MFU). Right: aggregate training throughput in tokens per second. The intermediate 560M–3.2B models sustain approximately 48–52% MFU, while utilization falls substantially for both the smallest and largest models.*

The low utilization of the 160M model illustrates the distinction between throughput and hardware efficiency. Despite achieving the highest token rate in the family, more than 660,000 tokens/s, the model uses only about one eighth of the available theoretical floating-point throughput. At this scale, each training step contains relatively little model arithmetic, so fixed costs such as kernel launches, synchronization, data movement and distributed-training overhead constitute a much larger fraction of execution time. We therefore interpret the d6 run as predominantly overhead-bound rather than compute-bound.

The opposite effect appears at the largest scale. Moving from d24 to d30 increases iteration time from approximately 20 to 46 seconds while MFU falls from 51.9% to 41.8%. This is consistent with increasing memory and communication pressure as model depth and parameter count grow, although we did not perform kernel-level profiling and therefore we treat this explanation as an interpretation rather than a direct measurement.

The highest utilization is obtained in the middle of the family. The d12, d18 and d24 models all operate close to 50% MFU despite differing substantially in parameter count and global batch size. For this hardware and implementation, these scales therefore occupy the most efficient training regime among the configurations tested.

The result also illustrates why parameter count alone is a poor predictor of practical training cost. The 160M model requires little arithmetic per token but does not use the H100s efficiently, whereas the largest model exposes substantially more useful arithmetic per step but begins to encounter a different efficiency limit. The training system consequently exhibits an intermediate regime in which model size is large enough to use the accelerators effectively without yet incurring the utilization decline observed at 6B.

**4.4 Training dynamics and stability.** Across the five pre-training runs, comprising approximately one million logged optimization steps in total, we observed no numerical divergences and no NaN or Inf values. Training remained stable across the complete model family despite the approximately 36-fold range in parameter count and the substantial differences in batch size and step count.

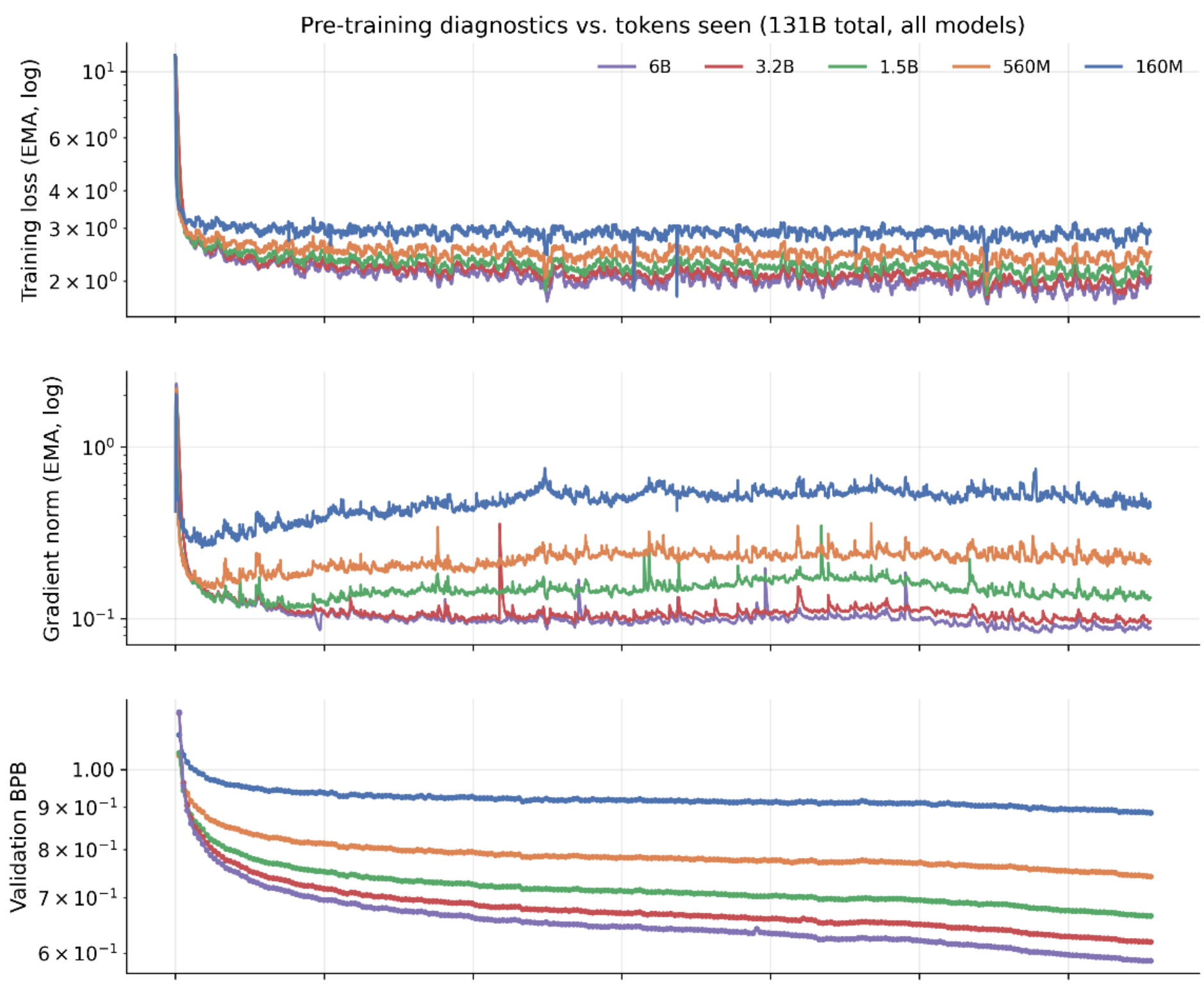

*Figure 13: Training dynamics of the five base models plotted against cumulative tokens consumed. Top: exponentially weighted training loss. Middle: exponentially weighted gradient norm. Bottom: validation bits per byte (BPB). Plotting against consumed tokens rather than optimization steps places all five runs on the same data axis despite their different global batch sizes and numbers of updates.*

The training-loss curves show the expected ordering with model capacity: larger models maintain lower loss throughout training, and the separation between scales remains visible across the full 131-billion-token stream. None of the curves exhibits instability or a late-training loss increase. The validation BPB measurements follow the same ordering and continue to improve through training, reaching final values of 0.8867, 0.7427, 0.6657, 0.6195 and 0.5876 for the 160M, 560M, 1.5B, 3.2B and 6B models, respectively.

The gradient-norm behavior is less straightforward. The smallest model exhibits the largest gradient norms, while progressively larger models operate at lower values; the 160M model is also unique in showing an increase during the later part of training. However, gradient norms are not directly comparable across these runs because global batch size increases systematically with model scale. Larger batches average gradients over more training examples and can reduce the stochastic component of the gradient, potentially contributing to the observed ordering. We therefore do not attribute the gradient-norm differences to model capacity or over-training. No corresponding instability is visible in either training loss or validation BPB.

A second feature is the repeated structure visible in the training-loss traces. Similar local rises and falls appear at approximately corresponding positions in the shared token stream despite substantial differences in model size, batch size and optimization-step count. Their alignment on the token axis suggests that at least part of this structure is associated with the data sequence rather than with model-specific optimization dynamics.

This interpretation is consistent with the curriculum construction described in Section 2.4. All five models consume the same master shards in the same order, so changes in local source composition can appear at the same token positions in every run. Because the source streams contributing to the curriculum are not equal in length, their relative representation can also change toward the end of the sequence as individual streams are exhausted. We did not attempt to assign individual features in the loss curves to specific sources and therefore we treat this correspondence as a qualitative observation rather than a measured causal effect.

Taken together, the training trajectories show that the same optimization pipeline remained numerically stable from 160M to 6B parameters and that increasing model capacity produced a consistent reduction in both training loss and validation BPB under an identical data stream. This controlled ordering motivates the fixed-data scaling analysis of Section 4.5.

**4.5 A fixed-data scaling law.** Because all five models consume the same 131.072-billion-token stream in the same order, the final validation loss can be examined as a function of model capacity at fixed data. This isolates a different question from compute-optimal scaling: rather than asking how parameters and tokens should be traded under a fixed compute budget, we ask how much additional validation performance is obtained by increasing model size while holding the training data constant.

At the end of pre-training, validation bits per byte decreases monotonically from 0.8867 for the 160M model to 0.5876 for the 6B model. The five endpoints are well described by a power law with an additive floor [26, 27]:

$$BPB(N_B) = E + A * N_B^{-\alpha} \tag{5}$$

where ($N_B$) is model size in billions of parameters. Fitting the five models gives:

$$BPB(N_B) = 0.4366 + 0.2602 * \ N_B{}^{-0.3047} \tag{6}$$

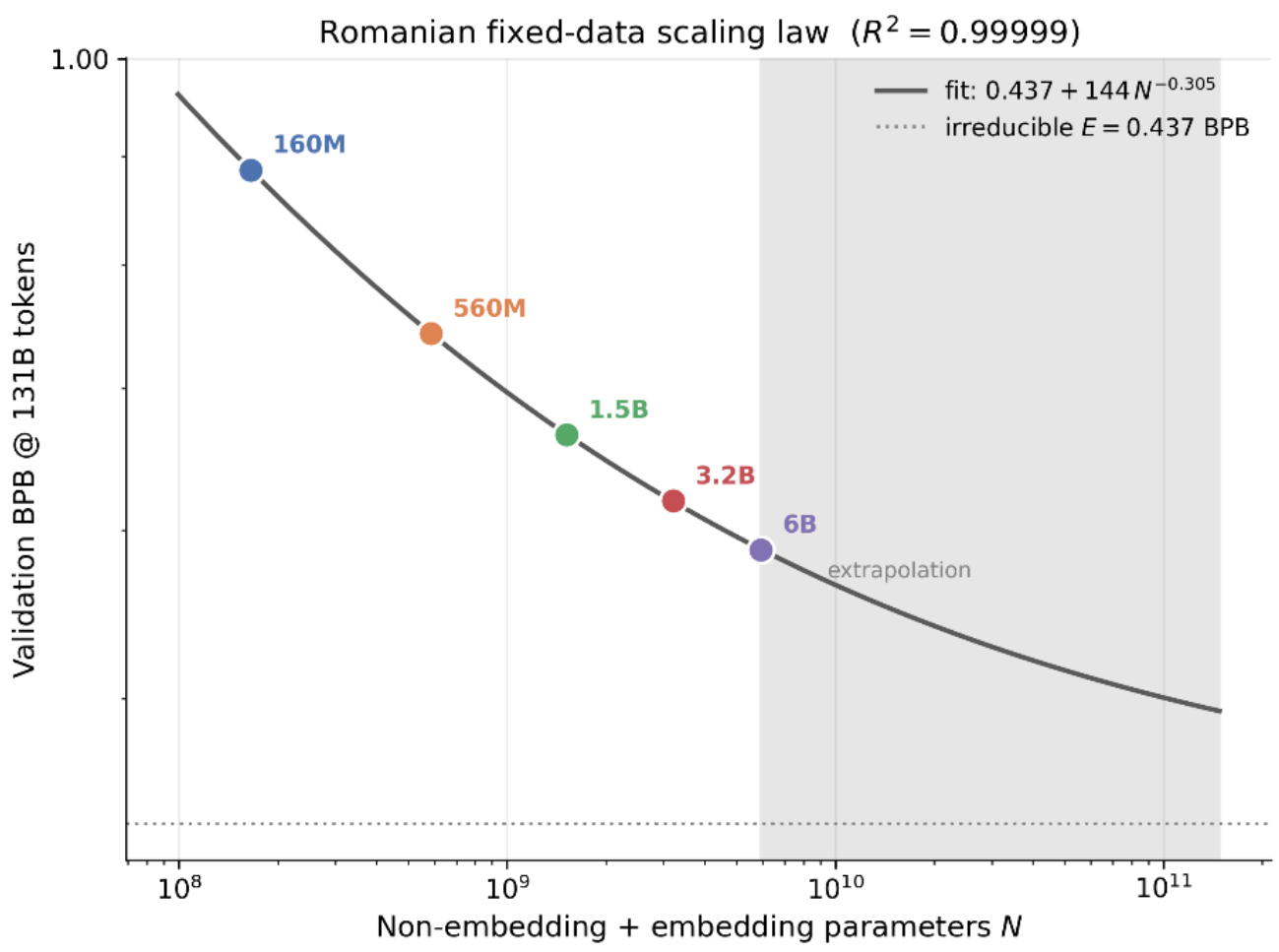

*Figure 14: The graph displays the final validation bits per byte (BPB) measured after training on 131.072 billion tokens, plotted against the total parameter count. The established scaling law is expressed by the formula: BPB($N_B$) = 0.4366 + 0.2602 × N_B$^{(-0.3047)}$. In this equation, $N_B$ represents the model parameter count in billions. The horizontal dotted line highlights the fitted fixed-data floor (E = 0.4366). Notably, this baseline value is strictly specific to*

*this 131-billion-token experiment and must not be misinterpreted as the true linguistic entropy of the Romanian language.*

The numerical fit is unusually tight. It gives $R^2$=0.999993, an RMSE of 2.86E-4 BPB, and a maximum absolute residual of 5.25E-4 BPB. We do not regard the $R^2$ value alone as strong evidence: the fit contains three free parameters and is estimated from only five model endpoints, leaving two residual degrees of freedom. The more useful observation is the scale of the residuals themselves, which remain below approximately 6E-4 BPB across the family and are comparable to the precision at which validation loss was logged.

The fitted constant E=0.4366 is best interpreted as a fixed-data asymptote. Within this empirical model, increasing capacity while repeatedly training on the same 131-billion-token stream yields progressively smaller gains and approaches a non-zero floor. This floor is not an estimate of the intrinsic entropy of Romanian: a larger or cleaner training corpus could shift the entire relation downward. It describes only the capacity-limited trajectory observed under the data budget used here.

The fitted curve can nevertheless be used to illustrate how rapidly the cost of further improvement increases if data is fixed. Solving the relation for model size gives the following extrapolations:

| Target validation BPB | Estimated parameters | Multiple of the 6B model |
|---|---|---|
| 0.58 | 7.1B | 1.2× |
| 0.56 | 11.6B | 2.0× |
| 0.55 | 15.3B | 2.6× |
| 0.54 | 20.7B | 3.5× |

*Table 11: Parameter count implied by the fixed-data scaling-law fit for selected target validation losses. These values are extrapolations beyond the largest observed model and should be interpreted as illustrations of the fitted curve rather than predictions of independently trained models.*

The extrapolation shows rapidly diminishing returns to additional capacity. Moving from the measured 6B result of 0.5876 BPB to approximately 0.58 requires only a modest increase in model size, whereas reaching 0.55 would require roughly 15B parameters, about 2.6 times the size of the largest model trained here. Progress becomes increasingly expensive as the fitted floor is approached, and extrapolations sufficiently close to $E$ rapidly become physically uninformative. For this reason we do not use the fit to make practical predictions far below 0.54 BPB.

The important conclusion is therefore not the numerical value of the asymptote, but the shape of the fixed-data curve. Across the observed range, increasing model capacity continues to improve validation BPB, but with strongly diminishing returns when the training stream is held constant. This motivates the complementary compute-optimal analysis of Section 4.6, which asks whether additional training compute is better allocated to larger models or to additional token exposure.

**4.6 Compute-optimal scaling.** The fixed-data analysis of Section 4.5 asks how validation loss changes when model capacity is increased while the training stream is held constant. A complementary question is how a fixed amount of training compute should be divided between model size and token exposure. Because the five Neurograf models differ strongly in tokens per parameter while their complete training trajectories are available, the same runs can be used to estimate this compute-optimal frontier.

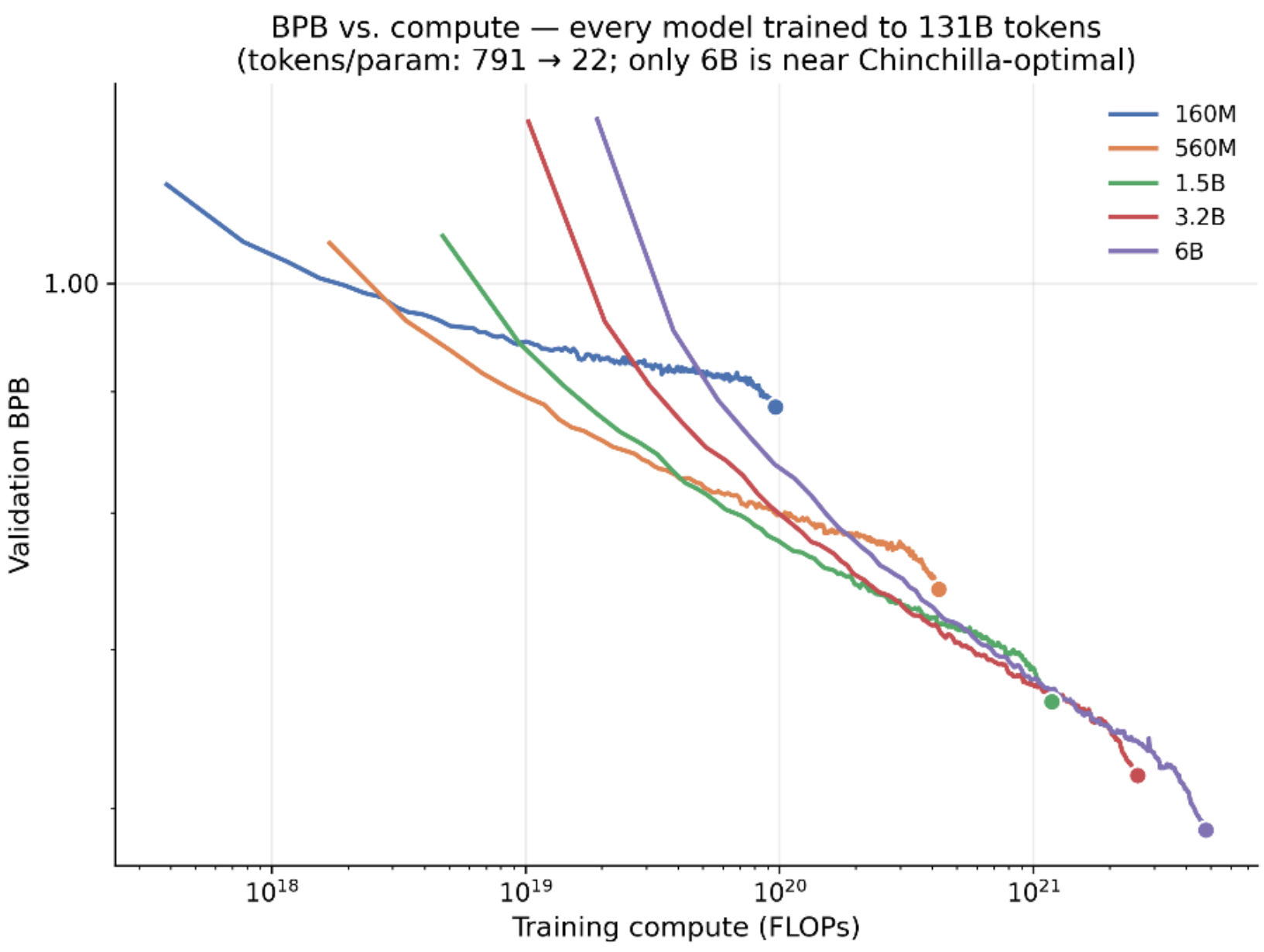

*Figure 15: Validation bits per byte against cumulative training compute for all five base models. Each curve represents the full training trajectory of one model. The lower envelope identifies, at each compute scale, the model-size region providing the lowest validation loss. The*

*family spans a wide range of data exposure, from 791 tokens per parameter for the 160M model to 22 for the 6B model.*

Figure 15 shows validation bits per byte as a function of cumulative training compute for all five models. Each trajectory begins in a regime where additional training rapidly reduces validation loss and then progressively flattens. At a fixed compute budget, the model with the lowest loss changes with scale: smaller models initially benefit from processing more tokens per unit compute, while larger models become preferable as the available compute increases. The lower envelope of the five trajectories therefore traces the region from which compute-optimal model sizes can be estimated.

To quantify this envelope, we selected twenty representative compute budgets spanning approximately 1E19 to 1.1E21 FLOP. At each budget, validation loss was interpolated from the available checkpoints for every model. The resulting loss values define an iso-FLOP slice: validation BPB as a function of model size when total training compute is held approximately constant.

Figure 16 shows five representative slices. For each compute budget, we fit a quadratic function in log(N) to the sampled model losses and take its vertex as the interpolated compute-optimal parameter count $N_{opt}$. The purpose of the quadratic is local interpolation of the minimum rather than a claim that validation loss is globally quadratic in model size. All twenty selected budgets yield interior minima within the sampled model-size range, rather than minima located at the smallest or largest available model.

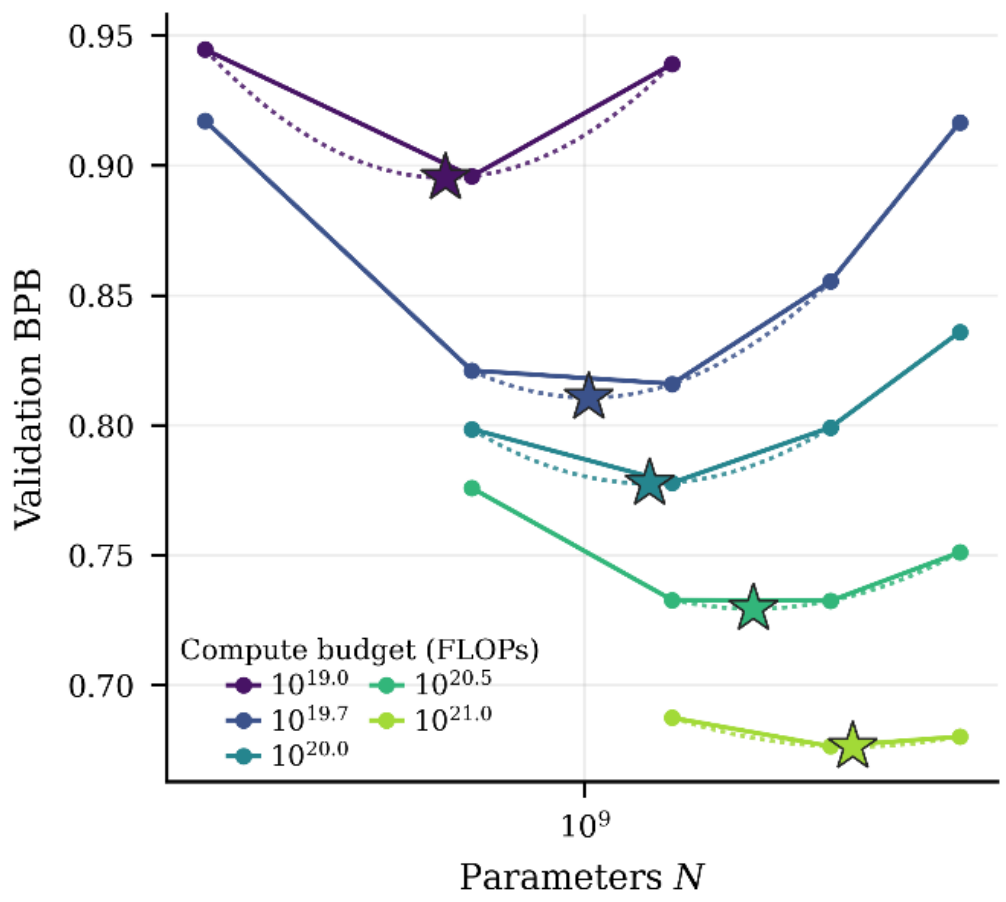

*Figure 16: Validation bits per byte as a function of parameter count at five representative fixed-compute budgets. Points are obtained from the five model trajectories at matched cumulative FLOP. A quadratic fit in log(N) is used to interpolate the minimum of each iso-FLOP slice; stars mark the resulting $N_{opt}$. The complete analysis uses twenty compute budgets.*

Repeating this procedure over all twenty compute budgets produces a sequence of optimal model sizes as a function of training compute. We fit this relation with a power law:

$$N_{opt} \propto C^{\alpha} \tag{7}$$

where (C) is total training compute. The resulting exponent is:

$$N_{opt} \propto C^{0.423}, R^2 = 0.988 \tag{8}$$

against the Chinchilla reference exponent of 0.50 [26].

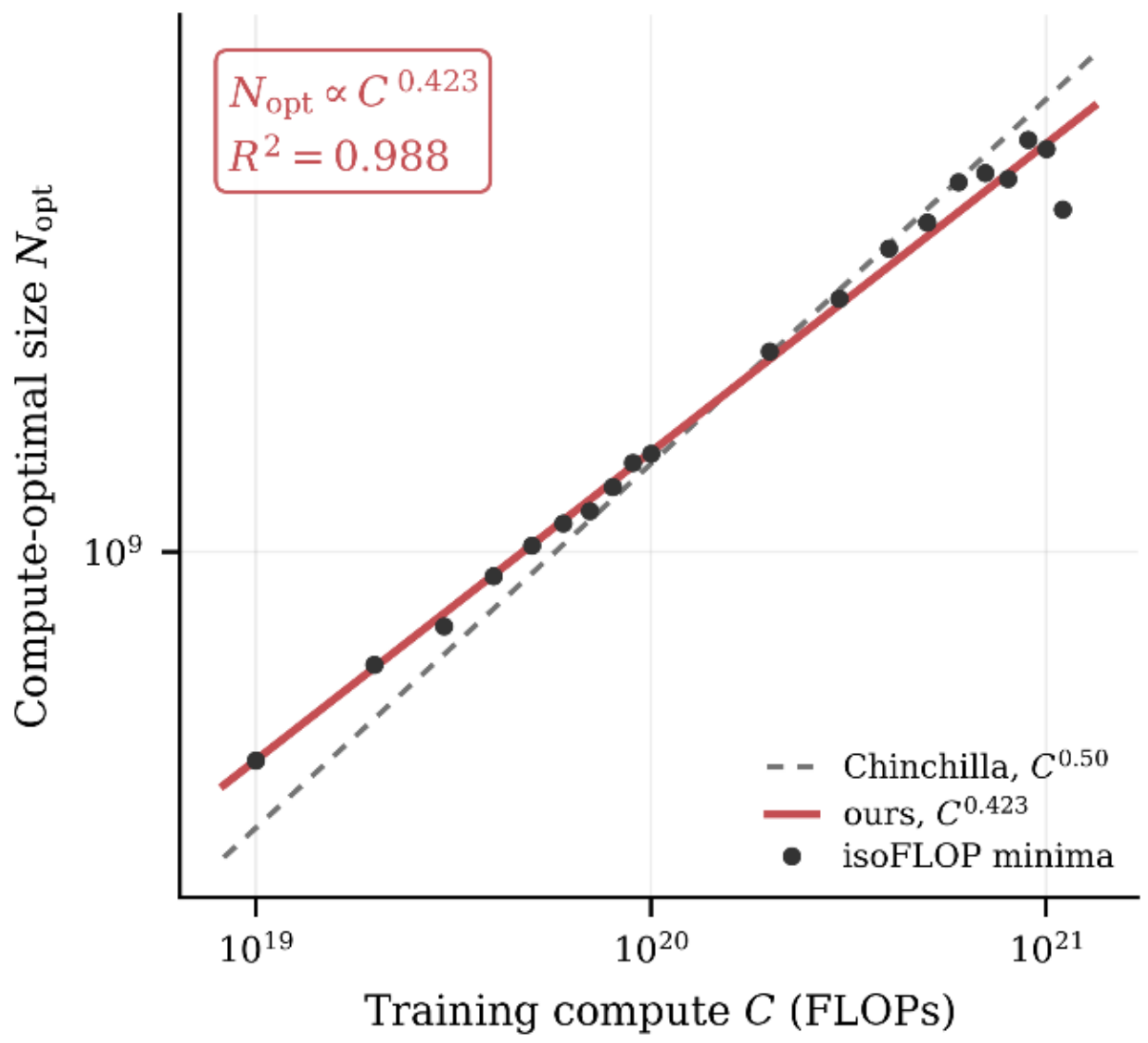

*Figure 17: Estimated compute-optimal parameter counts as a function of training compute for twenty iso-FLOP budgets. The fitted Neurograf relation is shown together with the $C^{0.5}$ scaling relation reported by Chinchilla [26] as a reference. The comparison concerns the allocation of computing between model capacity and token exposure rather than absolute model quality.*

The fitted exponent is shallower than the approximately 0.50 parameter-scaling exponent associated with the Chinchilla compute-optimal relation. Within the range examined here, increasing the available compute therefore shifts the estimated optimum toward model size more slowly than the Chinchilla reference would predict. Equivalently, a larger fraction of additional computes is allocated to continued token exposure rather than to increasing parameter count.

This result is consistent with the fixed-data behavior of Section 4.5: the Neurograf models continue to benefit from additional capacity, but those gains diminish rapidly when data is held constant. Taken together, the two analyses suggest that, for the Romanian corpus and model family studied here, additional training data remains particularly valuable relative to aggressive parameter scaling.

We do not interpret the exponent 0.423 as a universal scaling law for Romanian, and three limitations are important.

- **Checkpoint construction.** The twenty compute budgets are derived from intermediate checkpoints along five long training trajectories rather than from twenty independent runs trained specifically to each compute budget.

This substantially reduces the computational cost of the analysis but introduces a systematic complication under the warmup–stable–decay schedule. Many intermediate checkpoints are sampled during the constant-learning-rate plateau and therefore have not undergone the terminal learning-rate decay that a run intentionally stopped at the same compute budget would normally receive. Their measured losses may consequently be higher than the corresponding fully scheduled runs, and the magnitude of this bias has not been quantified.

- **Compute range.** The iso-FLOP analysis spans approximately 1E19 to 1.1E21 FLOP. This is substantially below the complete training of the two largest models, which consumed approximately 2.57 and 4.79 ZettaFLOP, respectively. The fitted exponent therefore characterizes the region in which all five model trajectories provide useful overlap and should not be extrapolated automatically to the full scale of the largest Neurograf runs.
- **Curvature.** The highest computer optima fall systematically below the fitted power-law line. This may indicate that the effective exponent changes with compute rather than remaining constant across the complete range. Alternatively, it may result from the checkpoint-schedule bias described above, which becomes increasingly important when the selected checkpoint lies far from the final decay phase. The present experiment cannot distinguish these explanations.

The compute-optimal result should therefore be read as an empirical property of this model family, corpus, and training construction rather than as a language-general law. Within that scope, however, the finding is consistent: the optimal parameter count grows more slowly with computing than the $C^{0.5}$ reference, indicating a stronger relative value of token exposure across the range accessible to the experiment.

This observation is particularly relevant for a low-resource setting. If similar behavior were reproduced on other Romanian corpora or other low-resource languages, it would imply that simply transferring parameter-to-token allocation rules derived from English-dominated training may not be optimal. Establishing that broader claim would require replicated scaling experiments across independent corpora and languages and is left for future work.

## 5. Base-Model Capability and Knowledge Scaling

Validation loss provides a general measure of next-token prediction quality, but it does not directly reveal how particular linguistic or factual capabilities change with model scale. The fixed-data design of the Neurograf family provides an opportunity to examine this question under controlled conditions: all five base models consume the same 131.072-billion-token training stream in the same order, while model capacity increases from 160M to 6B parameters. As described in Section 4, this produces a family spanning approximately 36 times in parameter count while holding tokenizer, context length, data order and total token exposure constant.

We therefore complement the loss-based scaling analysis with two forms of capability evaluation. First, a small Romanian grammar probe is used to examine the emergence of basic grammatical preference during pre-training. Second, factual and cultural knowledge is evaluated using the 1,000-question WWTBM Romanian multiple-choice benchmark [28] *Who Wants to Be a Millionaire* (Romanian edition, *Vrei să fii Milionar?*). The latter provides an external evaluation set with four candidate answers per question together with metadata describing cultural context, difficulty and topic.

The evaluations in this section are performed directly on the pre-trained base models, without instruction tuning. This distinction is important because a base model has not been trained to follow a multiple-choice instruction or to emit an answer letter in response to a question. We therefore formulate both evaluations as continuation-likelihood tasks rather than generation tasks. This permits the benchmark to probe preferences already present in the language model without requiring an instruction-following interface.

**5.1 Evaluation methodology.** The Romanian grammar probe contains 100 fixed items. Each consists of a shared prefix and two candidate continuations, one grammatically correct and one incorrect. The candidates are scored under teacher forcing using the mean token log-probability of the continuation. The reported margin is:

$$\Delta = \overline{\log(P(y^{+}|x)} - \overline{\log(P(y^{-}|x)} \quad (9)$$

where x is the shared prefix, $y^{+}$ is the correct continuation and $y^{-}$ is the matched incorrect continuation. Positive values indicate preference for the grammatically correct form, while larger values indicate stronger separation between the alternatives. The complete checkpoint sweep covers 130 saved states across the five model sizes. Because the models use different global token batches and optimization-step counts, checkpoint position is expressed as cumulative tokens consumed rather than optimization step.

For the WWTBM benchmark [28], each question supplies four answer candidates and one gold answer. Rather than displaying all four choices and requesting a letter, the base model receives the fixed prefix:

*Question: <question> \n Response:*

and each candidate answer is scored independently as a continuation of that prefix. For candidate $y_j$, the score is the mean continuation token log-probability:

$$s(y_i|x) = \frac{1}{\lfloor y_i \rceil} \sum_{i} \log P\left(y_{j,i} \middle| x, y_{j,<i}\right) \tag{10}$$

The predicted answer is the candidate with the highest score. Mean rather than summed log-probability is used so that longer answers are not penalized solely because they contain more tokenizer tokens. With four candidates, random-choice accuracy is 25%.

This evaluation protocol differs from instruction-model evaluations in which all answer options are included in a prompt and the model is asked to generate the corresponding letter. Consequently, the absolute WWTBM scores reported here should not be interpreted as directly interchangeable with published instruction-model results obtained under that protocol. The purpose here is instead to measure factual discrimination already present in the Neurograf base representations.

For accuracy measurements, 95% Wilson [29] intervals are reported over benchmark items. These intervals quantify finite-sample uncertainty associated with the benchmark questions; they do not represent seed-to-seed training variance, since each model size was trained only once.

No WWTBM-specific training stage was used. As with any model trained on a large web corpus, incidental exposure to individual underlying facts, or to publicly available material related to the original quiz questions, cannot be excluded. The benchmark should therefore be interpreted as an external factual-discrimination test rather than as a contamination-proof reasoning evaluation.

**5.2 Early acquisition of Romanian grammar.** Romanian grammatical preference develops rapidly across the model family. Figure 18 shows the mean correct-versus-incorrect continuation margin for all saved checkpoints. Even the 160M model develops a strongly positive margin early in pre-training and improves only gradually thereafter. Increasing model size raises the separation, but the difference becomes comparatively small among the 1.5B, 3.2B and 6B models.

At the final checkpoint, mean grammatical margin rises from 1.4696 for the 160M model to 1.6100 for 560M and 1.7086 for 1.5B, before reaching 1.7710 and 1.7787 for the 3.2B and 6B models. The two largest models are therefore nearly indistinguishable on this diagnostic despite their approximately twofold difference in parameter count.

This behavior suggests that the relatively simple grammatical contrasts represented in the probe require substantially less capacity than the factual tasks considered below. Once the model reaches the billion-parameter regime, additional capacity continues to improve language-model loss but produces only modest gains on these particular grammatical distinctions.

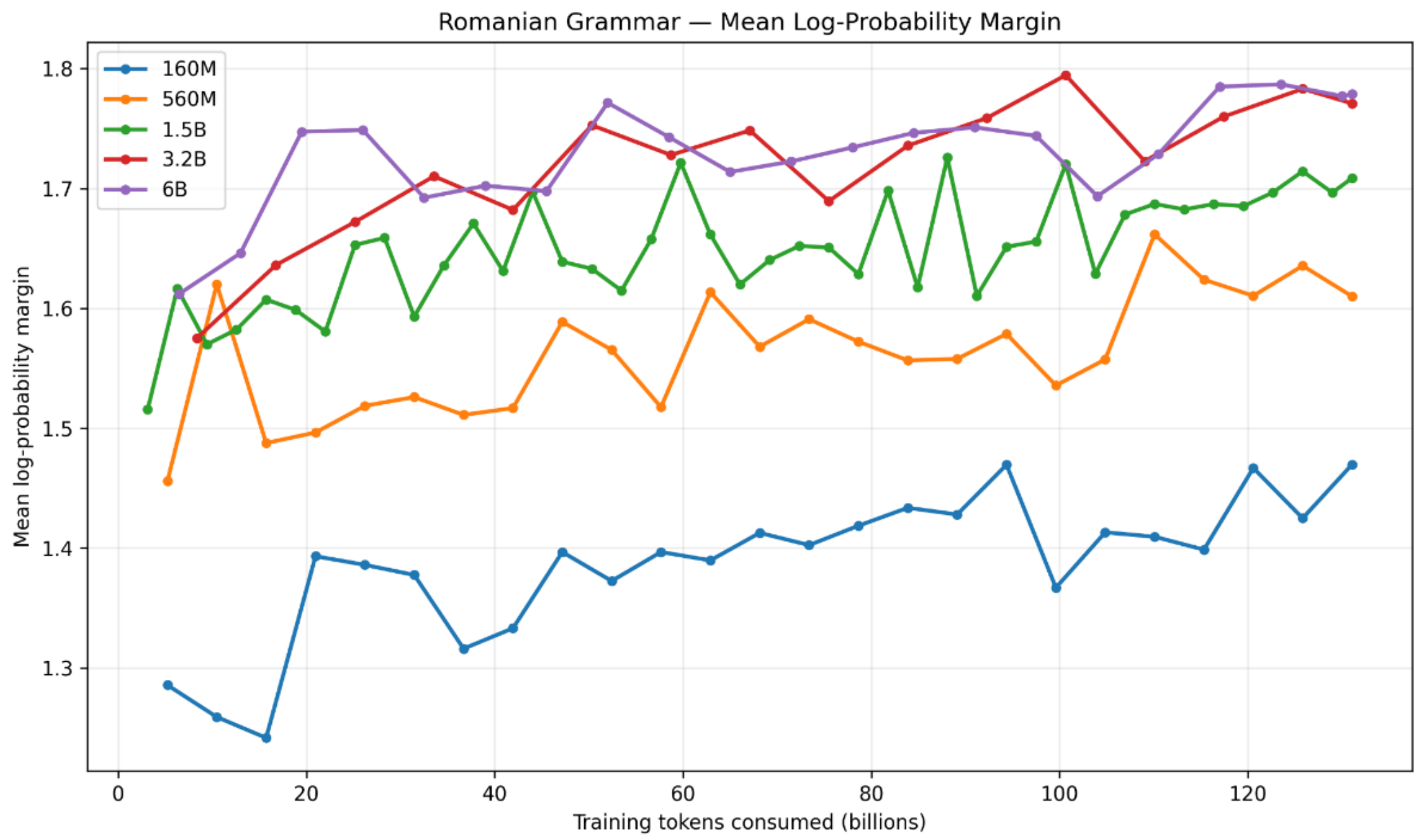

*Figure 18: Mean correct-versus-incorrect continuation log-probability margin on 100 Romanian grammar probes as a function of cumulative pre-training tokens. Each point represents a saved base-model checkpoint. Larger positive values indicate stronger preference for the grammatically correct continuation.*

**5.3 Factual knowledge acquisition during pre-training.** WWTBM produces a substantially different developmental pattern. Figure 19 shows benchmark accuracy throughout pre-training for all five model sizes. Every model rises above the 25% random-choice baseline, but the trajectories occupy strongly scale-dependent bands.

The 160M model reaches approximately 34–38% accuracy relatively early and subsequently fluctuates within a narrow range, ending at 35.9%. The 560M model similarly rises rapidly into the low-40% range and ends at 42.2%. In contrast, the billion-parameter models sustain substantially higher performance: the 1.5B model reaches 51.6%, the 3.2B model 61.4%, and the 6B model 65.0% at the final checkpoint.

The within-model trajectories are not monotonic. Individual checkpoints can improve or decline by several percentage points even while validation BPB continues to decrease smoothly. This differs from the validation-loss behavior described in Section 4.4, where the ordering by capacity is stable throughout training and all five models continue to improve toward their final BPB values.

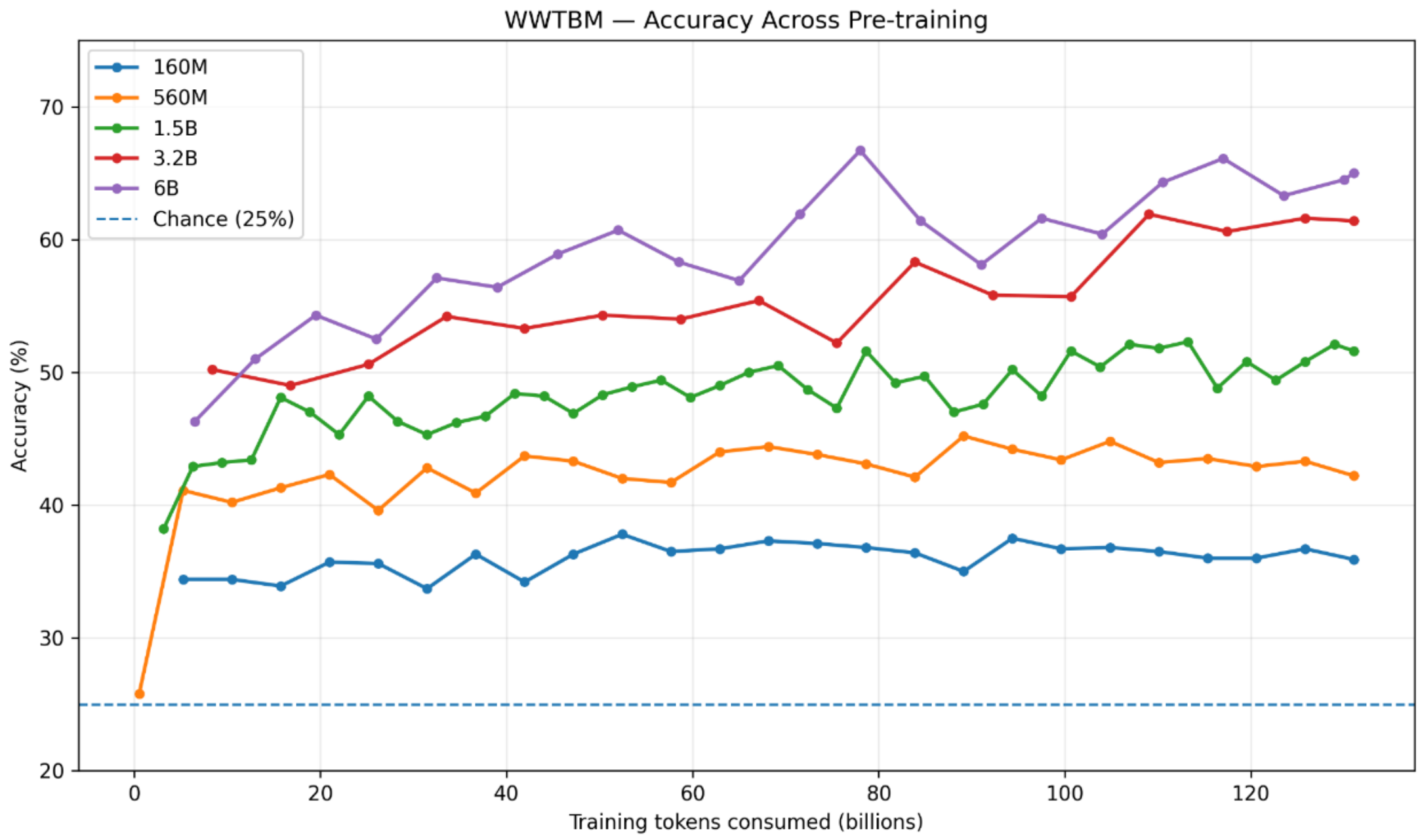

*Figure 19: WWTBM accuracy as a function of cumulative pre-training tokens for the five Neurograf base models. All models are evaluated on the same 1,000 four-choice questions using mean continuation token log-probability. The dashed line marks the 25% random-choice baseline.*

The benchmark therefore exposes a distinction between average next-token prediction and discrete factual discrimination. Continued token exposure generally improves WWTBM performance, especially early in training, but additional data does not produce a smooth monotonic increase in the number of correctly ranked answers. More importantly, the separation between model sizes is substantially larger than the within-model improvement obtained from extending pre-training after the initial acquisition phase.

This behavior is particularly visible in the smallest models. The 160M model receives approximately 791 training tokens per parameter and the 560M model approximately 224, compared with only 22 tokens per parameter for the 6B model. Despite this extreme data exposure, their WWTBM performance remains far below that of the larger models. For this factual-discrimination task, additional token exposure therefore does not compensate indefinitely for limited representational capacity.

The benchmark distinguishes questions whose cultural context is specific to Romania from those categorized as global [28]. Table 12 reports the final results under this split.

| Model | Global | Romanian | Overall |
|---|---|---|---|
| 160M | 35.51 | 36.92 | 35.90 |
| 560M | 41.89 | 43.01 | 42.20 |
| 1.5B | 50.21 | 55.20 | 51.60 |
| 3.2B | 61.17 | 62.01 | 61.40 |
| 6B | 64.08 | 67.38 | 65.00 |

*Table 12: Final WWTBM accuracy (%) by cultural context. Romanian and global labels are those supplied by the benchmark metadata.*

Romanian-context questions are not systematically weaker for the Neurograf models. In fact, all five final checkpoints score slightly higher on the Romanian subset, with the largest differences appearing for the 1.5B model (+5.0 percentage points) and the 6B model (+3.3 points). Because the two subsets differ in question composition and difficulty, this should not be interpreted as a controlled causal estimate of a Romanian-language training advantage. It nevertheless provides no evidence in this benchmark for a deficit on culturally Romanian questions.

**5.4 Scaling of final WWTBM accuracy.** Because every model encounters the same 131.072-billion-token training stream, the five final WWTBM scores can be compared as a function of parameter count under fixed data. Accuracy rises monotonically across the family:

*35.9%→42.2%→51.6%→61.4%→65.0%*

Over the measured 160M–6B range, these five endpoints are well described by a linear relation between accuracy and the logarithm of parameter count,

$$A(N) = a + b log_{10} N \quad (11)$$

with b = 19.55 percentage points per decade in parameter count and $R^2$ = 0.975. Thus, within this model family and fixed-data experiment, increasing model size by one order of magnitude is associated with approximately 19.6 percentage points of additional WWTBM accuracy.

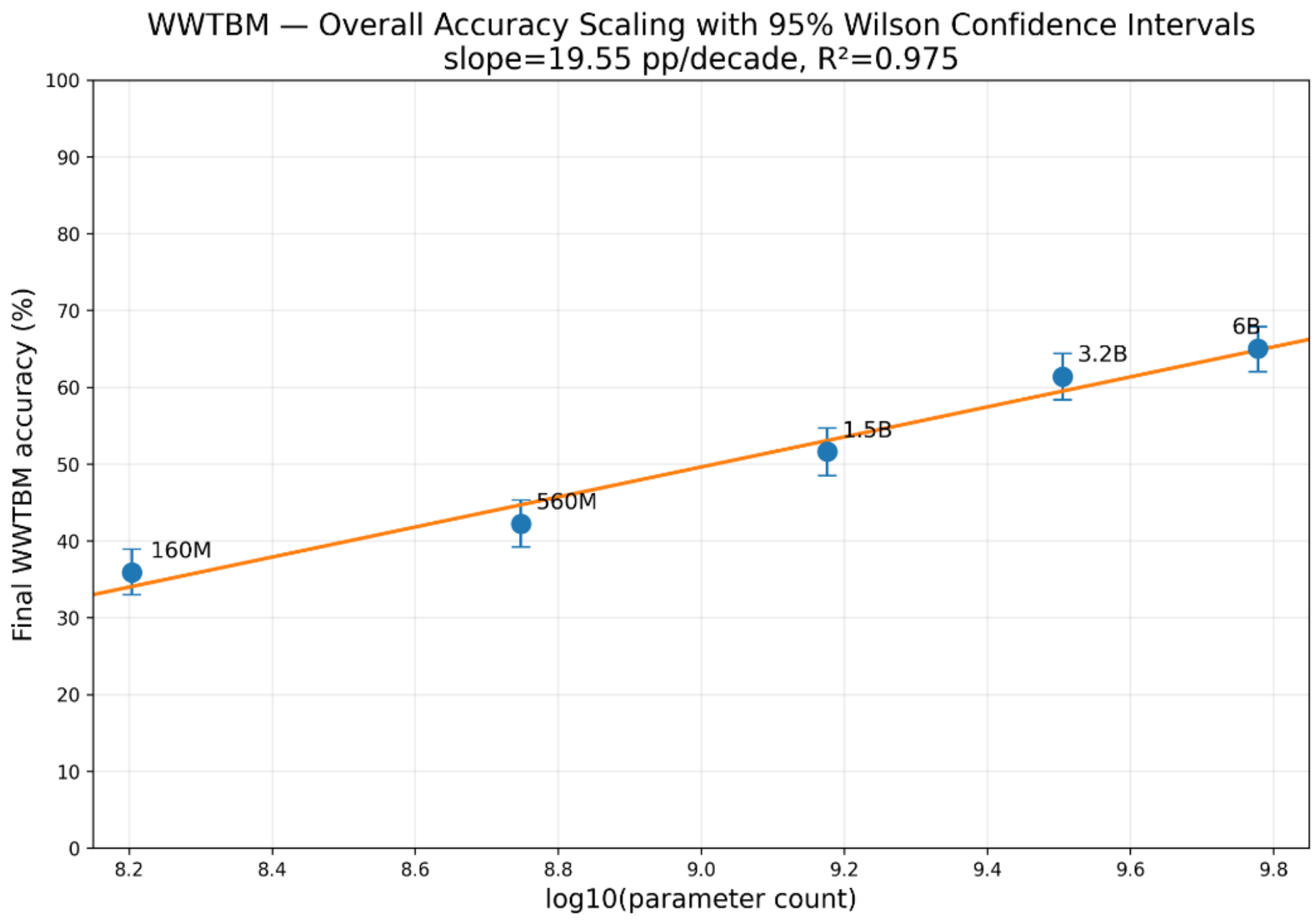

*Figure 20: Final WWTBM accuracy as a function of total parameter count after the common 131.072-billion-token training stream. Error bars denote 95% Wilson confidence intervals over the 1,000 benchmark questions. The line is an ordinary linear fit in $\log_{10}(N)$, yielding 19.55 percentage points per parameter decade and $R^2$=0.975. The fit is descriptive over the measured 160M–6B range and is not intended for unrestricted extrapolation.*

The relation should be interpreted as an empirical description of the observed range rather than as a universal capability scaling law. Only five model endpoints are available, and accuracy is bounded between zero and one, so a linear relation in log(N) cannot remain valid indefinitely. In particular, extrapolations toward very high accuracies would eventually become physically meaningless. The useful result is therefore the strong monotonic ordering and the magnitude of the observed scale dependence, not distant parameter-count predictions.

The contrast with the token-exposure trajectories of Figure 19 is notable. Within any single model, additional pre-training produces comparatively modest and noisy gains once an early performance band is reached. Across models trained on exactly the same complete stream, however, final accuracy changes by 29.1 percentage points between 160M and 6B. WWTBM performance in this regime therefore appears more strongly capacity-sensitive than exposure-sensitive once each model has already consumed a large quantity of data.

**5.5 Topic-dependent knowledge scaling.** The aggregate scaling relation hides substantial variation across semantic domains. WWTBM assigns each question to one of twelve topic categories [28]. We therefore repeated the final-checkpoint scaling analysis independently within each category.

Figure 21 shows the resulting twelve fits using a common horizontal and vertical scale. All categories exhibit a positive fitted slope with model size, but the strength and regularity of the relationship vary considerably. History shows the strongest measured scale dependence at 25.67 percentage points per parameter decade, followed by geography at 22.30, sport at 21.65 and cinema at 21.61. Religion scales at 20.23 points per decade, while general knowledge, music and entertainment, and literature lie near the overall relation.

The weakest slopes are observed for medicine, gastronomy, art and culture, and science and technology. Science and technology is particularly distinct: its fitted slope is only 13.08 percentage points per decade and the fit is substantially less regular ($R^2$=0.643) than for the other topics. By contrast, general knowledge shows an exceptionally regular relation across the five scales ($R^2$=0.991), while sport, art and culture, gastronomy, religion and geography all exceed approximately $R^2$=0.94.

The category-level results should be interpreted cautiously because the number of questions differs substantially between topics. Wilson intervals are therefore shown separately for every point, with visibly wider uncertainty for small categories such as medicine and religion. The fitted slopes nevertheless indicate that the overall scale effect is distributed broadly across the benchmark rather than being produced by a single dominant topic.

The final 6B model also exhibits large differences in absolute accuracy across categories. Religion reaches 80.5%, followed by history at 76.6%, geography at 70.6%, sport at 68.0% and literature at 67.5%. At the opposite end, science and technology reaches 55.1%, medicine 51.9%, and gastronomy 50.0%.

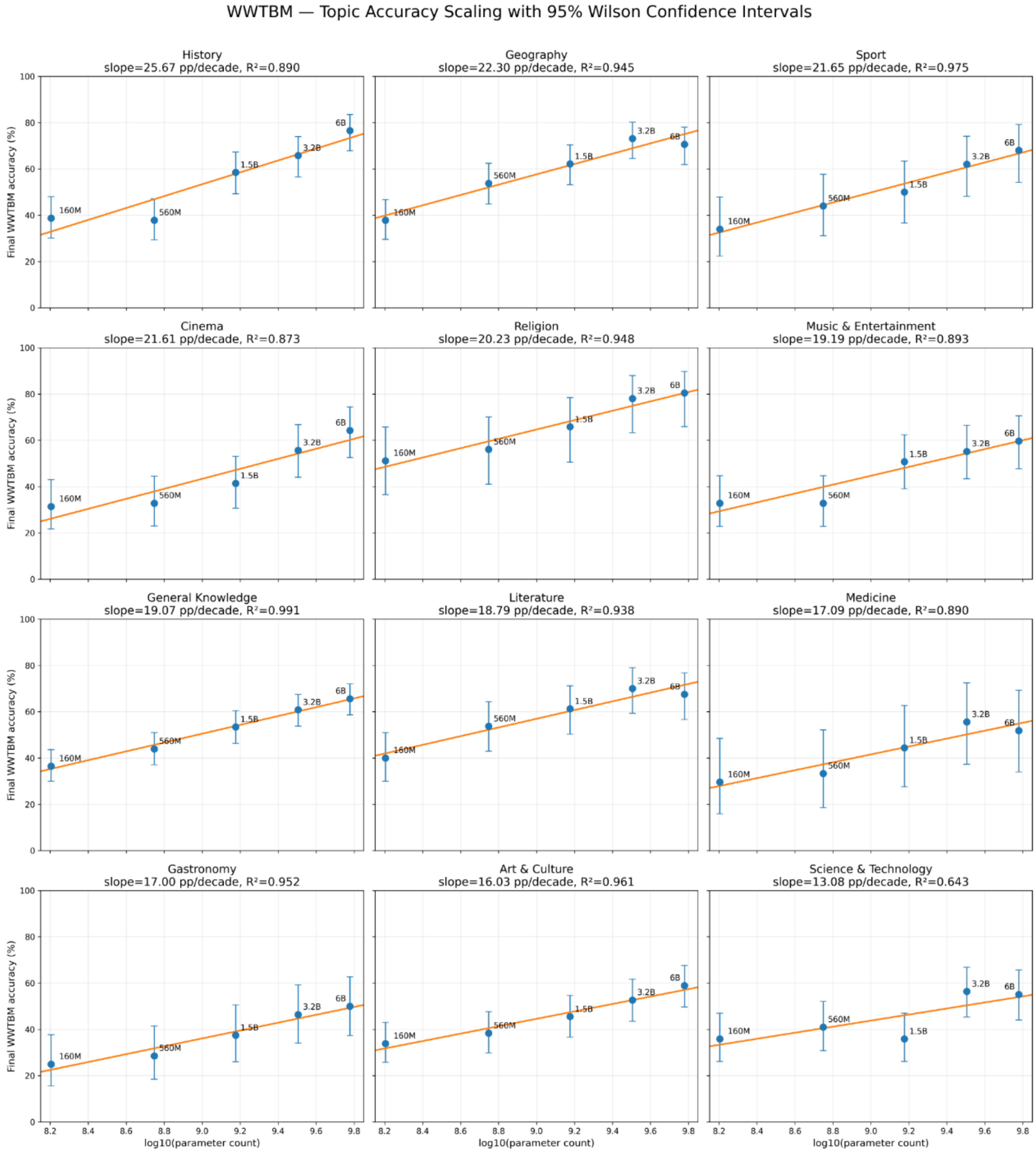

*Figure 21: Final WWTBM accuracy by model size for the twelve benchmark topics. All panels share identical axes. Points show final base-model accuracy and error bars denote 95% Wilson confidence intervals over the questions in that topic. Lines show ordinary linear fits in $\log_{10}$ (N). Panel titles report the fitted slope in percentage points per parameter decade and $R^2$.*

These values should not be interpreted as a general ranking of the model's competence in entire academic domains. The categories describe the composition of this particular quiz benchmark, and individual question styles differ in ambiguity, specificity and required factual association. The topic results are most useful comparatively: they show that the benefit of increased model capacity is heterogeneous across types of factual knowledge.

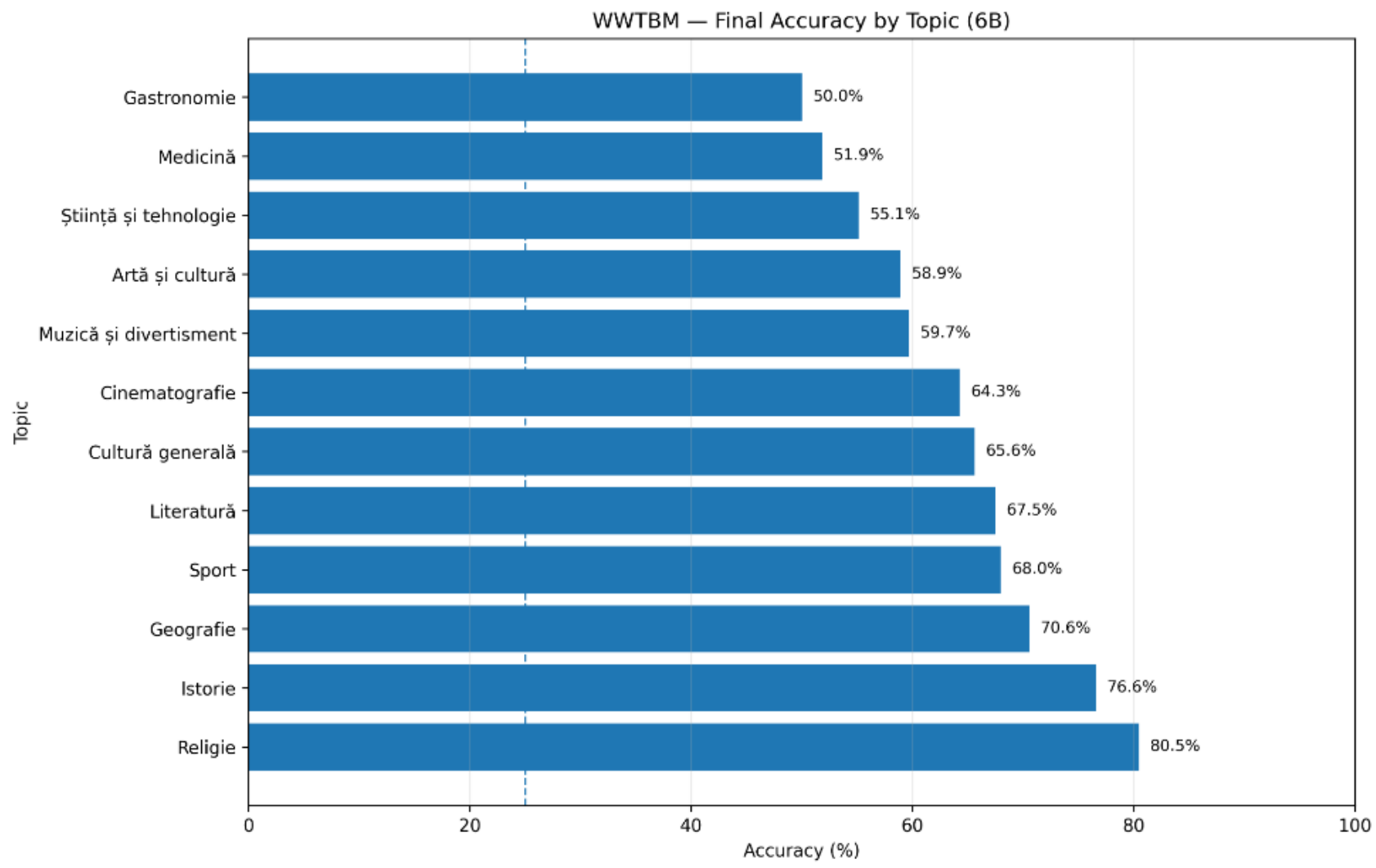

*Figure 22: Final WWTBM accuracy by topic for the 6B Neurograf base model. The dashed vertical line marks the 25% random-choice baseline. Topic accuracies describe performance on the benchmark question subsets and are not intended as comprehensive domain evaluations.*

**5.6 Prediction churn across pre-training.** The non-monotonic checkpoint trajectories in Figure 23 raise a further question: does a stable aggregate accuracy conceal continual changes in which individual questions are answered correctly?

To examine this, each WWTBM item was compared between consecutive checkpoints. Four transitions are possible:

*wrong⟶correct, correct⟶wrong, correct⟶correct, wrong⟶wrong.*

We define prediction churn as:

$$C = N_{wrong \rightarrow correct} + N_{correct \rightarrow wrong} \quad (12)$$

that is, the number of benchmark items whose binary correctness state changes between two saved checkpoints.

Figure 23 shows substantial churn throughout much of pre-training. Even when overall accuracy changes little, tens to more than one hundred questions can exchange correctness state between adjacent checkpoints. This explains why the aggregate accuracy curves fluctuate: a small net change can be the result of much larger opposing flows of newly correct and newly incorrect predictions.

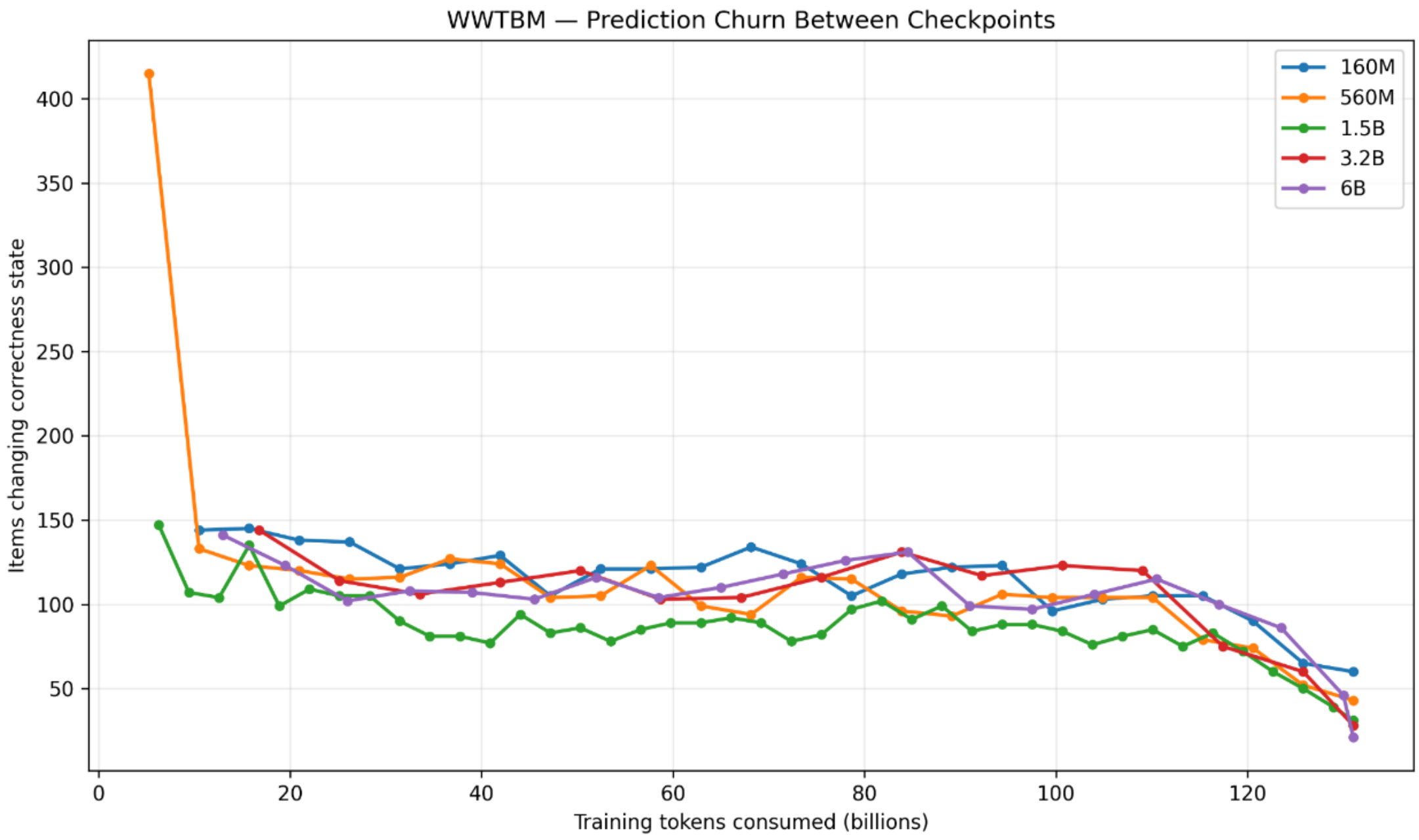

*Figure 23: Number of WWTBM questions changing correctness state between consecutive saved checkpoints. Churn is defined as the sum of wrong-to-correct and correct-to-wrong transitions. Checkpoint spacing differs across model sizes, so the figure is intended primarily to show within-model temporal behavior rather than absolute cross-model transition rates.*

This phenomenon should not automatically be interpreted as literal factual forgetting. Correctness is a thresholded property of four continuously varying continuation scores. An item can change state when the gold answer and strongest distractor exchange rank by a very small log-probability margin, even if the underlying representations remain broadly similar. The result therefore demonstrates instability of benchmark decisions rather than direct storage and deletion of isolated facts.

A marked reduction in raw transition counts is visible toward the end of the runs. This region overlaps with the terminal learning-rate warmdown described in Section 4.2, during which the learning-rate multiplier decreases toward 0.05.

Smaller parameter updates would naturally be expected to reduce the frequency with which closely ranked alternatives exchange order.

However, the absolute churn counts should be interpreted cautiously because adjacent saved checkpoints are not separated by exactly the same number of training tokens across all models, and the final checkpoint interval is shorter for some runs.

Cross-model comparisons of raw churn magnitude are therefore less informative than the within-model temporal trend. A token-normalized transition analysis would be required to quantify the relation between learning-rate decay and churn rate more precisely.

**5.7 External Romanian base-model comparison.** The preceding WWTBM analyses are intended primarily as a controlled comparison within the Neurograf family. All five models share the same tokenizer, training stream and evaluation procedure, making mean continuation token log-probability a consistent scoring quantity across model sizes. To provide an external reference point, we additionally evaluated two publicly available Romanian base models: RoLlama2-7B-Base, a 7B model obtained through Romanian continual pre-training of Llama 2, and LLMic_v2, a Romanian-focused 3B base model trained independently.

Direct use of the token-normalized WWTBM score is less appropriate for this comparison because the three model families employ substantially different tokenizers. Neurograf uses its 80k vocabulary, RoLlama2 retains a 32k Llama tokenizer, and LLMic_v2 uses a 128k tokenizer. The same textual answer may therefore contain a different number of tokens in each model, causing token-count normalization itself to influence the ranking of candidate answers.

For the external comparison, we therefore repeated WWTBM evaluation for all seven models using UTF-8 byte-normalized continuation likelihood. The prompt and teacher-forced candidate-scoring procedure remain unchanged; only the length normalization is replaced. For each candidate answer, the summed continuation log-likelihood is divided by the number of UTF-8 bytes in that continuation. This provides a common text-length unit independent of model-specific tokenization. The candidate with the highest byte-normalized score is selected as the prediction.

| Model | Global | Romanian | Overall | 95% CI |
|---|---|---|---|---|
| Neurograf 160M | 37.31 | 35.13 | 36.70 | 33.77–39.73 |
| Neurograf 560M | 46.60 | 43.73 | 45.80 | 42.73–48.90 |
| Neurograf 1.5B | 56.87 | 55.20 | 56.40 | 53.31–59.45 |
| LLMic2 3B [9] | 67.82 | 55.20 | 64.30 | 61.28–67.21 |
| RoLlama2 7B [8] | **69.49** | 53.05 | 64.90 | 61.89–67.80 |
| Neurograf 3.2B | 66.30 | 67.38 | 66.60 | 63.62–69.46 |
| Neurograf 6B | 67.68 | **69.89** | **68.30** | 65.35–71.11 |

*Table 13: WWTBM accuracy (%) for the five Neurograf base models and two external Romanian-language base models using UTF-8 byte-normalized continuation likelihood. Global and Romanian labels follow the benchmark metadata. The 95% confidence interval is the Wilson interval over all 1,000 questions.*

The external models provide strong reference points. RoLlama2-7B achieves the highest accuracy on the globally contextualized subset at 69.49%, consistent with the broad knowledge inherited from its Llama 2 initialization, but reaches only 53.05% on Romanian-context questions. LLMic_v2 is also highly competitive despite its smaller 3B scale, reaching 67.82% on global questions and 64.30% overall.

The Neurograf models show a different balance. Neurograf-3.2B reaches 66.60% overall and 67.38% on Romanian-context questions, exceeding both external models on the latter subset. Neurograf-6B obtains the highest overall accuracy at 68.30% and the highest Romanian-context accuracy at 69.89%, while remaining close to the external models on global questions at 67.68%.

These results should be interpreted as contextual benchmark comparisons rather than controlled estimates of training efficiency. The models differ in architecture, tokenizer, corpus composition, total training exposure and, in the case of RoLlama2, inherited upstream pre-training. Nevertheless, the comparison indicates that the capability levels observed within the Neurograf scaling experiment are competitive with independently developed Romanian base models, while the strongest Neurograf models show particularly high performance on the Romanian-context subset.

**5.8 Discussion.** The capability evaluations reveal a different view of pre-training scaling from that obtained through validation loss alone.

Basic Romanian grammatical preference emerges comparatively early. Once the models reach approximately the billion-parameter regime, the simple grammar probe provides relatively little separation between scales. Factual knowledge behaves differently. On WWTBM, all five models remain clearly ordered by capacity despite receiving exactly the same complete training stream, and final accuracy increases from 35.9% at 160M parameters to 65.0% at 6B.

This distinction is especially informative given the intentionally data-rich training of the smaller models. The 160M model receives approximately 791 training tokens per parameter, compared with only 22 for the 6B model, yet additional exposure does not allow the small model to approach the factual-discrimination performance of the larger models. Under this benchmark and training regime, factual performance therefore appears strongly constrained by capacity once sufficient data has already been observed.

The scale dependence is also not uniform across knowledge domains. Positive scaling is observed across all twelve WWTBM categories, but fitted slopes range from approximately 13 to 26 percentage points per parameter decade. History, geography, sport and cinema show the strongest gains, while science and technology exhibits both the shallowest and least regular relation. These differences may reflect interactions among benchmark construction, corpus frequency, factual specificity and model capacity, and the present experiment cannot isolate their individual causes.

Finally, checkpoint-level evaluation shows that capability development is not simply the monotonic accumulation of correct answers. Individual benchmark decisions repeatedly change state during training even while average validation loss improves. The reduction in transition activity toward the end of training suggests increasing prediction stability as optimization updates become smaller, although checkpoint-spacing effects prevent attributing this behavior uniquely to the learning-rate schedule.

Taken together, these results support a view in which increased model capacity does more than uniformly lower next-token loss. Different behaviors respond differently to scale: relatively simple grammatical distinctions saturate early, externally evaluated factual knowledge remains strongly capacity-sensitive, semantic domains show heterogeneous scaling rates, and item-level factual preferences remain dynamic throughout much of pre-training. The fixed-data construction of the Neurograf family makes these differences directly observable because they arise under a common tokenizer, data stream and total token budget.

## 6. Synthetic Instruction Data and Supervised Fine-Tuning

Romanian instruction-tuning data has often been constructed by translating English-language instruction sets [8] or by distilling responses from larger multilingual models. Both approaches are practical, but they inherit some of the source language task framing, discourse conventions and cultural reference points, and may also inherit licensing or terms-of-service constraints from the source data or generation system.

Neurograf follows a different strategy. The bulk of the supervised fine-tuning corpus was generated natively in Romanian, using naturally occurring Romanian web topics as semantic seeds and a locally hosted generative model to expand them into instruction data. The objective was not to produce factually verified knowledge from the source corpus, but to anchor the synthetic data in subject matter encountered in Romanian-language web content while retaining full control over the generation pipeline.

The resulting dataset combines two sources of synthetic instruction data. The larger component is topic-seeded: Romanian web topics are extracted, expanded and subsequently converted into multiple instruction types. A smaller set of deliberately generated seed-free examples was added to cover behaviors that were insufficiently represented by the topic-driven pipeline, including atomic questions, invalid or unanswerable requests, multi-turn informal conversation, multi-turn how-to interactions, AI identity and origin questions, and elementary mathematics.

**6.1 The generation pipeline.** The topic-seeded component is constructed as a cascade in which each stage provides progressively richer source material for the next. The pipeline consists of seed extraction and filtering, two successive expansions, and task-specific user–assistant generation.

**6.1.1. Seed extraction.** Candidate topics were mined from URL metadata in the Romanian portion of FineWeb-2 [13]. Many web addresses contain no useful human-readable title, but a subset includes title-like strings whose words are commonly separated by hyphens or related delimiters. We scanned the URL fields for such material and reconstructed candidate title strings where possible.

Approximately 500,000 candidate seeds were examined. Candidate length, measured as the number of reconstructed words in the URL title, showed a strongly right-skewed distribution: very short strings were dominated by slugs, identifiers and incomplete fragments, while longer strings were more likely to express an identifiable subject. We therefore selected candidates from the descending right-hand side of the length distribution rather than sampling uniformly from all recovered URL strings.

A second filtering stage used a locally hosted Gemma-3 4B [30] model as a lightweight binary classifier. Its task was deliberately simple: determine whether the candidate expressed a recognizable subject that could plausibly serve as a generation seed. The model was used only as a coarse filter rather than as a semantic validator, and occasional classification errors were accepted as part of the high-throughput pipeline. Duplicate seeds were removed by hashing after word normalization. Approximately 100,000 candidates were retained after filtering.

The accepted seeds are short topic descriptions rather than factual source passages. A typical example is:

*"dreptul la concediu de odihna"*

The role of the seed is therefore to establish subject matter and local linguistic context, not to provide sufficient information from which a factual answer can be copied.

**6.1.2. First expansion.** Each retained seed was passed once to GPT-OSS-120B [14], running locally on H100 hardware, to produce an expanded_seed: a short Romanian explanatory passage of several sentences. For the example above, the model expands the bare topic into a short description of the right to paid annual leave and its general legal purpose.

**6.1.3. Second expansion.** Each expanded_seed was then passed through GPT-OSS-120B a second time to generate an infused_seed, typically a longer mini-article of approximately 2,500 characters. This second stage introduces substantially more context, detail and narrative structure than the original seed and provides enough material for tasks that require a source passage rather than only a topic.

Every accepted seed therefore produces exactly one expanded_seed and one infused_seed. These two representations are used differently during instruction generation. Expanded seeds are used primarily for open-ended tasks such as open question answering, brainstorming, free-form generation and conversational interaction. Infused seeds are used primarily for closed question answering and text-transformation or information-processing tasks, where the generated prompt can refer to a concrete source passage.

**6.1.4. Instruction generation.** GPT-OSS-120B generates both sides of the supervised interaction: the user request and the assistant response. The model is therefore responsible not only for producing answers but also for constructing the prompts, task formulations and multi-turn exchanges used during supervised fine-tuning. The topic seed influences what the conversation is about, while the generator determines the wording and substantive content of the final interaction.

The seeded pipeline constitutes the majority of the SFT corpus. It is supplemented by seed-free synthetic material generated directly by GPT-OSS-120B for task families that were useful to include explicitly but were not naturally produced in sufficient quantity from web-topic seeds. These include short atomic questions, invalid or nonsensical requests, four- to five-turn informal conversations, three-turn how-to exchanges, AI identity and provenance questions, and simple mathematical tasks. The two sources are combined before curriculum assembly.

GPT-OSS-120B was chosen both for its Romanian-language capability (being a reasoning model that can most of the time catch it's own Romanian grammar errors) and because it could be executed locally under an Apache-2.0 licence. Local generation avoids dependency on the terms of service of a proprietary hosted model and makes unrestricted release of the resulting synthetic corpus substantially simpler.

**6.1.5. What the corpus does and does not ground.** The generation pipeline is best described as topic-grounded rather than fact-grounded. The initial seeds are mined from Romanian web metadata, so they anchor the instruction set in subjects that actually occur in Romanian-language online material. They also help preserve Romanian lexical choices, local topics and cultural framing that may be underrepresented in translated English instruction sets.

The factual substance of the resulting examples, however, is synthetic in nature. As the pipeline expands a short seed into several sentences and then into a longer article, increasingly large amounts of information are supplied by the generator rather than recovered from the original source. These statements are not independently fact-checked before entering the synthetic dataset.

The distinction is important. The resulting instruction data is Romanian-native in what it talks about, but model-generated in what it says. This makes the pipeline suitable for constructing large amounts of locally framed conversational and instruction-following data, but it does not make the resulting corpus a verified factual knowledge source. For deployment domains in which factual accuracy is critical, including public administration and legal or institutional use, this remains a material limitation of the present dataset and will be address in a future release.

**6.2 Taxonomy and coverage.** The final supervised fine-tuning dataset contains 178,502 entries spanning 16 broad domains and approximately 55 instruction or task types. Because many entries contain more than one user-assistant turn, the corpus comprises approximately 400,000 individual question-answer turns.

The taxonomy was not imposed as a target distribution before generation. Instead, categories were created during post-processing by the generator and each generated user request was assigned to the corresponding task and domain bins. The coverage matrix should therefore be read as a descriptive map of where the synthetic data naturally accumulated rather than as the result of a balanced sampling plan.

This distinction explains the strongly irregular distribution shown in Figure 24. No fixed quota was enforced for individual task–domain cells, and the generation process was allowed to populate some combinations much more heavily than others. The resulting dataset consequently contains a small number of high-frequency cells and a long tail of sparsely represented combinations.

Several features of the distribution are worth making explicit.

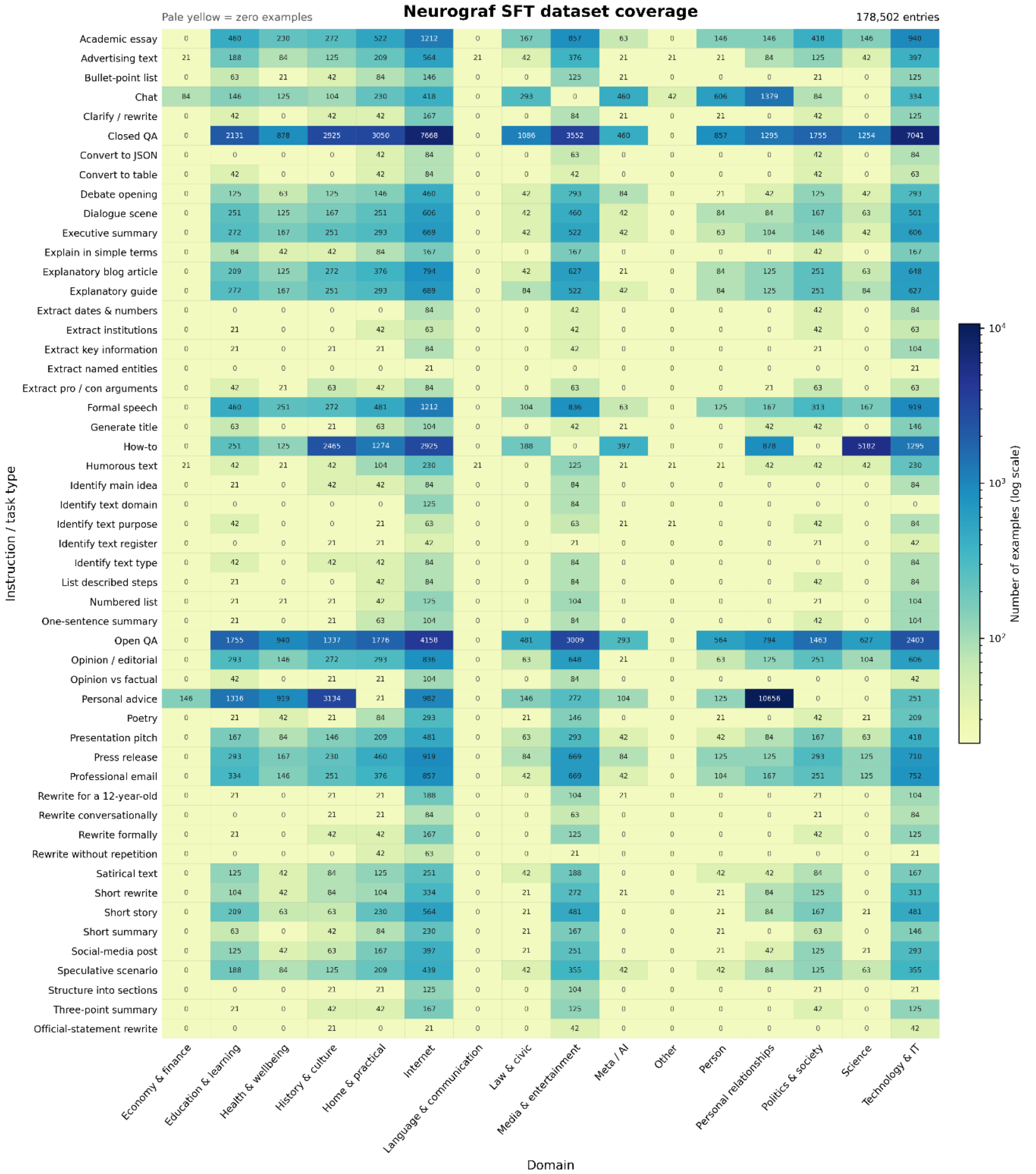

*Figure 24: Coverage of the Neurograf SFT dataset across 16 domains and approximately 55 instruction types, shown on a logarithmic colour scale; pale cells contain no examples. The taxonomy is descriptive rather than quota-driven: user requests were assigned to task and domain bins after generation, so the non-uniformity reflects the distribution produced by the synthetic pipeline rather than a predefined balancing target.*

First, coverage is highly concentrated. Closed question answering against internet-derived material contributes 7,668 entries, while personal advice in the personal-relationships domain contributes 10,656. By contrast, many task–domain combinations contain only a few dozen examples. The resulting assistants should therefore not be expected to exhibit uniform capability across the entire matrix; densely represented combinations receive substantially more supervised exposure than the long tail.

Second, several domains are only weakly represented. Economy and finance, Language and communication, and the residual Other category contain little data across most instruction types. The weakness of economy and finance is particularly relevant to the local-deployment motivations discussed in Section 1 and should be regarded as a substantive gap in the present instruction corpus.

Third, structured-output tasks are present but shallowly populated. Examples involving conversion to JSON or tables, named-entity extraction, date and number extraction, and related formatting operations frequently occur only at the level of tens of examples. Their presence broadens the behavioral coverage of the dataset, but the current corpus does not provide enough supervision to justify assuming robust structured-output performance.

The resulting SFT corpus is therefore intentionally broad but not statistically balanced. Its purpose in this first Neurograf iteration was to expose the base models to a wide variety of Romanian instruction-following behaviors and to identify practical weaknesses in the fine-tuning procedure, rather than to construct a benchmark-balanced instruction dataset. The coverage matrix is included in full because it makes those strengths and omissions visible and provides a direct basis for improving the data mixture in future iterations.

**6.3 Task-type mixing.** Early supervised fine-tuning experiments revealed a qualitative failure mode that we refer to as instruction-format lock-in (see Appendix C.2). When a conversation began with one type of task, the assistant tended to preserve the response style established by that task even when a later user turn clearly requested something different.

For example, after answering a short factual question with a concise response, the model could continue producing similarly short answers when subsequently asked in the same conversation to generate a longer piece of text. The same long-form request issued in a new conversation produced a substantially longer response. The behavior therefore appeared to depend less on the model's underlying ability to perform the second task than on the conversational pattern established by the preceding turns.

We interpret this as a conditioning effect produced by training conversations that were internally too homogeneous. If every multi-turn example contains only one task family, the model can learn that the first interaction identifies the response regime for the remainder of the conversation.

To counter this behavior, we introduced deliberately heterogeneous multi-turn examples in which different instruction types were combined within a single conversation. A conversation could therefore begin with a short factual question and later request a longer explanation, creative generation, or another qualitatively different response form. The order of these task types was varied, and combined conversations were constructed so that the complete training example remained within the 2,048-token context limit.

The objective was not to balance task frequencies, but to expose the model explicitly to task switching within a conversation. The first user turn should provide information about the immediate request, not impose a permanent response format on subsequent turns.

We report this as a qualitative engineering observation rather than a controlled ablation: the mixed-task examples were introduced in response to behavior observed during early SFT experiments, and we did not train otherwise identical production models with and without this intervention. Nevertheless, the failure mode was sufficiently repeatable during development to motivate retaining mixed-task conversations in the final instruction corpus.

**6.4 Curriculum assembly.** The final instruction corpus is converted into a training stream using the same general principle applied during RoSfera pre-training: examples from different categories are interleaved rather than presented in long homogeneous runs. The objective is to keep local task composition relatively diverse and to avoid exposing the optimizer to many consecutive updates dominated by a single instruction type.

One SFT training example corresponds to a complete user-assistant interaction, which may contain either a single exchange or several conversational turns. Multi-turn and mixed-task conversations are assembled before training and constrained to remain within the 2,048-token context window. Each resulting example is padded to the full sequence length for training.

Examples are drawn from the available task categories in rotation and interleaved into the training file. Because the category streams contain unequal numbers of examples, the resulting curriculum is approximately balanced locally rather than globally: high-frequency categories continue to contribute after smaller categories begin to exhaust. This construction is therefore useful for reducing local clustering but does not transform the underlying dataset into a statistically balanced mixture.

The production SFT runs use a batch size of 64 such conversational examples per optimization step. Training continued for 5,333 steps, corresponding to 5333 * 64 = 341312 example presentations. Relative to the 178,502 entries in the final dataset, this corresponds to approximately 1.91 effective passes through the corpus.

Repeated examples were reshuffled and re-interleaved rather than being replayed in the same sequence and local neighborhood. The second exposure therefore preserves the underlying example content while changing the surrounding training context. Production training was stopped after this approximately two-pass horizon as part of the exploratory design of the first Neurograf SFT campaign rather than because the curriculum itself had reached a formal convergence criterion.

The construction is structurally related to the master-shard interleaving used for pre-training in Section 2.4. As in that setting, unequal stream lengths can produce gradual changes in mixture composition as categories become exhausted. Section 6.9 returns to repeated structure observed in the production loss curves that may reflect this property of the training stream.

**6.5 Training setup and the paired design.** Supervised fine-tuning was performed on the 1.5B, 3.2B and 6B Neurograf base models using the nanochat [15] supervised fine-tuning pipeline. All runs used a sequence length of 2,048 tokens and were executed on four H100 GPUs with gradient checkpointing enabled. The Neurograf implementation retained the reference SFT training procedure, with the instruction corpus and tokenizer substituted for the corresponding nanochat inputs. Gradient clipping was additionally enabled during SFT as an engineering stability measure after early runs encountered problematic large-gradient behavior.

Production fine-tuning used batches of 64 complete conversational examples per optimizer step. An example may contain a single user–assistant exchange or a multi-turn conversation but is assembled before batching and padded to the 2,048-token context length. The production runs continued for approximately 88.5 million processed tokens. This represents only about 0.07% of the 131-billion-token pre-training budget.

Hyperparameter development was performed on the 1.5B model before production runs commenced. Two controlled studies were used: a seven-point batch-size sweep and a learning-rate sweep. The resulting configuration, batch size 64 and the highest learning rate tested, was then transferred unchanged to the 1.5B, 3.2B and 6B production runs.

A useful property of the experimental design is that comparisons within each study are paired. Runs being compared receive the same underlying training examples in the same order. Learning-rate configurations therefore encounter identical batches at corresponding iterations, as do the three model scales in the production comparison. Differences in local batch difficulty are consequently shared between the runs rather than appearing as an uncontrolled source of variation.

This pairing is particularly useful because SFT loss is locally noisy. Different examples vary considerably in difficulty, length and number of supervised tokens, producing step-to-step loss fluctuations large enough to obscure relatively small differences between configurations. Where possible, we therefore compare losses on matched batches directly rather than relying only on differences between independently averaged curves.

Loss summaries are token-weighted. The number of tokens represented by each optimization step varies by approximately 20%, so assigning equal weight to every step would give short batches the same statistical weight as substantially longer ones. Token weighting instead estimates negative log-likelihood over the underlying token mass.

Final or "converged" loss values used in the hyperparameter comparisons are calculated over the final approximately 20–25% of the corresponding run rather than from a single terminal iteration. Uncertainty is estimated with a moving-block bootstrap [18] using contiguous blocks of 50 iterations, retaining part of the temporal autocorrelation present in the loss trajectories. Paired significance tests are used only as supporting evidence because neighboring optimization steps are not statistically independent.

Batch-size runs require one additional alignment step. Because changing batch size changes the number of optimization steps needed to process a fixed amount of data, these runs are compared at equal numbers of examples consumed rather than at equal iteration counts. This keeps the data budget fixed while exposing the optimization consequences of changing the number of examples per update.

**6.6 Batch size.** The first hyperparameter study examined batch sizes of 4, 8, 16, 32, 64, 128 and 256 examples per optimizer step on the 1.5B model. Every configuration processed the same fixed budget of 256,000 training examples. The learning-rate multiplier was held at 0.6 of the reference pre-training peak throughout the study.

Keeping the example budget fixed means that batch size changes the number of optimizer updates substantially. Batch 4 performs 64,000 updates over the study, batch 8 performs 32,000, and the count continues geometrically downward to 4,000 updates at batch 64 and only 1,000 at batch 256. The experiment therefore measures the behavior of the complete batch-size/learning-rate configuration rather than isolating batch size as a single independent variable.

Within these configurations, the response has a clear interior minimum. Loss decreases sharply as batch size rises from 4 to 32, reaches its lowest region between approximately 32 and 128, and is minimal at batch 64. Relative to this operating point, batch 256 incurs a penalty of approximately 0.41 nats and batch 4 approximately 0.62 nats. The optimum is therefore not located at either extreme of the tested range.

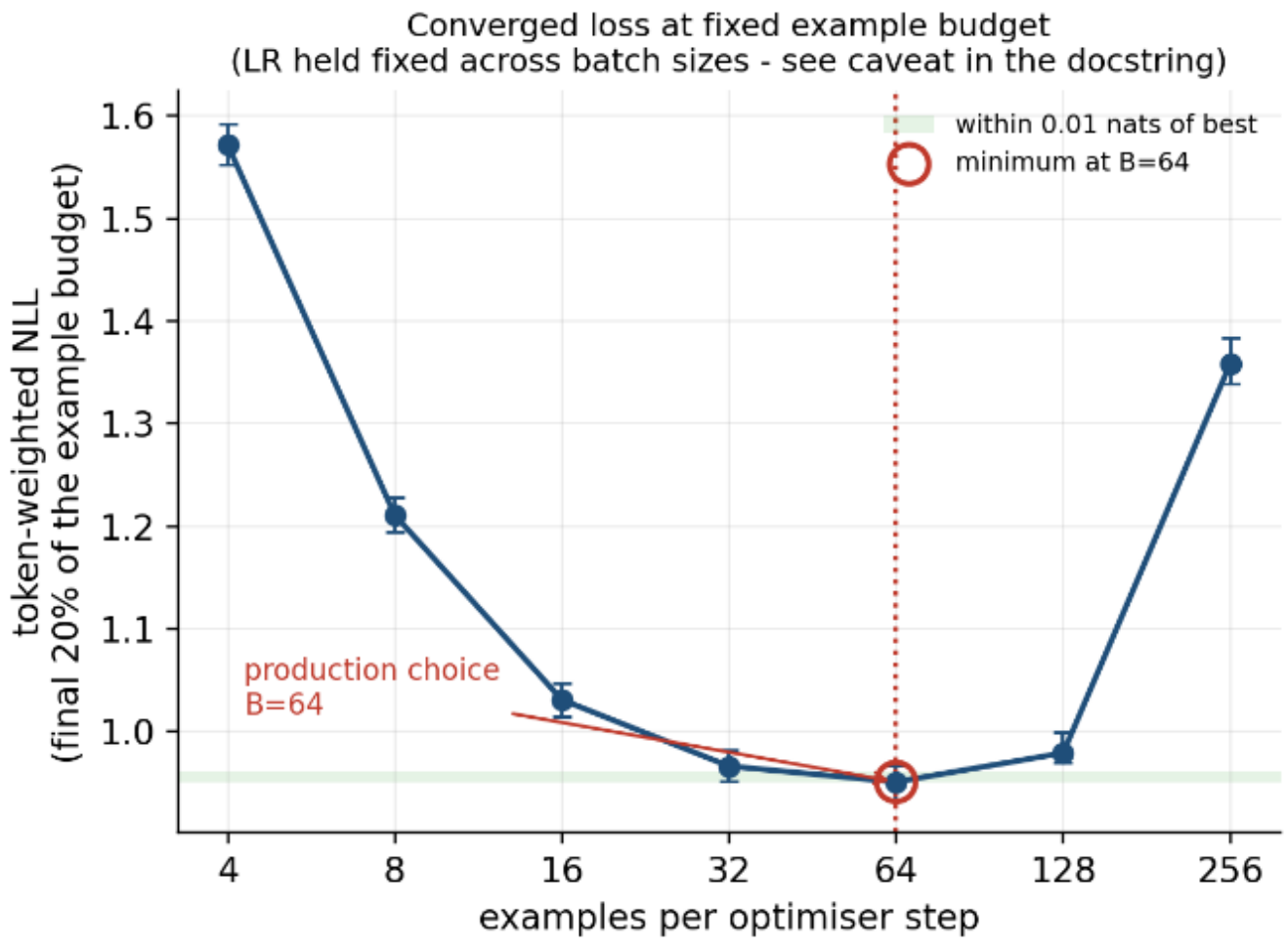

*Figure 25: Converged token-weighted negative log-likelihood after a fixed budget of 256,000 training examples, shown against examples per optimizer step for the 1.5B model. Seven batch sizes from 4 to 256 are tested while the learning-rate multiplier is held fixed at 0.6. The shaded region marks configurations within 0.01 nats of the lowest measured loss. Because learning rate is not retuned for each batch size, the curve characterizes the tested batch-size/learning-rate pairs rather than batch size in isolation.*

This result was practically important during development. Early SFT experiments had produced assistants whose behavior was noticeably poorer despite using otherwise reasonable training settings. The batch-size study showed that the optimization regime itself could account for a large loss difference, and batch 64 was consequently adopted for all production runs.

The shape of the curve should nevertheless not be interpreted as a pure batch-size law. The learning rate was fixed across all seven configurations. At batch 4, the optimizer performs sixteen times as many parameter updates as at batch 64 over the same example budget, while batch 256 performs only one quarter as many. A learning rate appropriate for one update frequency need not be appropriate for another.

This confounding is particularly relevant on the small-batch side of the curve. The large number of updates at batch 4 is performed with a step size selected for a substantially larger batch, while the batch-256 run is comparatively update-starved. The location of the measured minimum should therefore be interpreted as the best operating point among the tested batch combinations rather than as evidence that batch 64 is intrinsically optimal independently of learning rate.

For the present Neurograf training regime, however, batch 64 provided a robust practical choice: it lies near the center of the shallow low-loss basin, substantially outperforms both extremes, and provides enough optimizer updates over the available instruction corpus without requiring the very high update count of the smallest batches.

**6.7 Learning rate.** The second hyperparameter study examined the effect of learning-rate scale on the 1.5B model. Each run used the same sequence of training examples and the same batch size of 256 examples per optimizer step, while the learning rate was varied as a fraction of the reference pre-training peak. The sweep covered fractions from 0.1 to 1.0, with 1.0 corresponding to the full pre-training peak learning rate.

Because every configuration sees identical batches at corresponding iterations, the comparison can be treated as paired. Rather than relying only on absolute loss curves, we therefore examine the loss difference between the highest learning rate and each lower setting on the same batch.

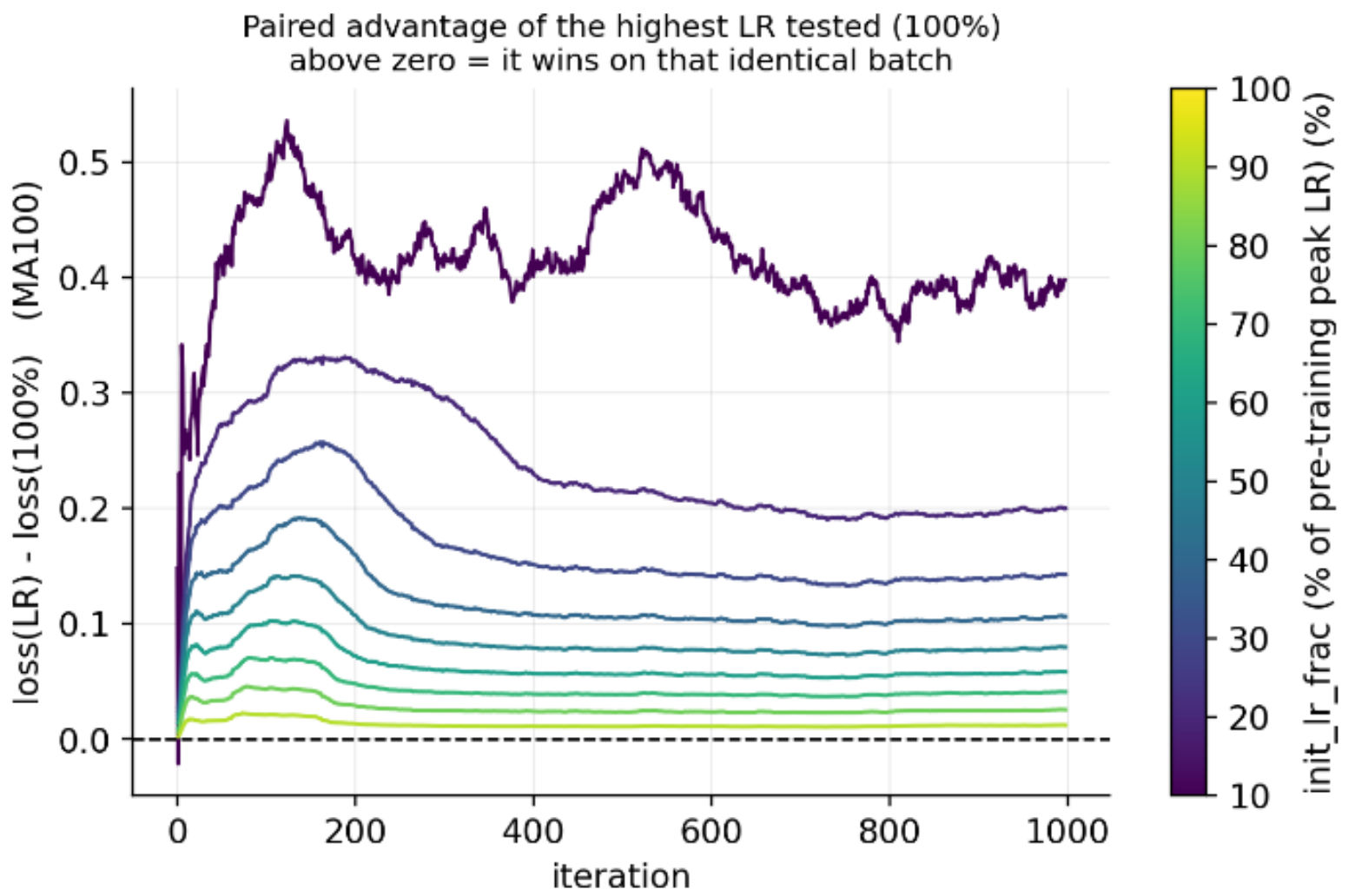

*Figure 26: Paired loss advantage of the highest learning rate tested, corresponding to the full pre-training peak, relative to each lower learning-rate fraction on identical batches. Positive values indicate that the full-rate run achieves lower loss on the same training batch. The advantage is largest early in training, around iterations 100–200, and decreases thereafter.*

Across the tested range, loss decreases monotonically as the learning-rate fraction increases from 0.1 to 1.0, with progressively smaller gains at the upper end. Since 1.0 corresponds to the full pre-training peak rather than to an arbitrarily chosen upper boundary, the direct empirical result is that no deterioration is observed up to that rate within the sweep.

Production SFT therefore used the full pre-training peak learning rate. This is an aggressive choice relative to common supervised fine-tuning practice, but it gave the lowest loss among the configurations tested.

The interpretation is complicated by the batch size used for the sweep. All learning-rate runs were performed at batch 256, which Section 6.6 identifies as the worst-performing batch-size configuration among the seven values tested. At the fixed 256,000-example budget used for hyperparameter probing, batch 256 provides only 1,000 optimizer updates, compared with 4,000 updates at the production batch size of 64.

The learning-rate sweep is therefore partially an experiment in compensating for an update-starved regime. Increasing the step size allows more parameter movement during the limited number of available updates, so the preference for a high learning rate cannot be assumed to transfer unchanged to a longer run performed at a smaller batch size.

The paired-difference curves contain evidence of this effect. For every comparison, the advantage of the full-rate run rises rapidly during the first approximately 100–200 iterations and then decreases as training progresses. This is consistent with the lower-learning-rate runs gradually recovering part of the early gap rather than with a persistent advantage that grows throughout training.

The production runs lasted 5,333 optimizer steps at batch 64, more than five times the horizon of the learning-rate sweep and with four times as many updates per fixed example budget. We therefore treat the observed advantage of the full pre-training learning rate as an upper bound on its likely benefit under the production training regime.

The learning-rate experiment nevertheless served its practical purpose. Within the constrained probe setup, the full pre-training peak was the best tested configuration, and no instability was observed at that value. It was therefore retained for the first production SFT campaign, with the intention of producing a future sweep over smaller batch sizes and learning rates to identify a better-isolated optimum.

**6.8 Model scale.** The production SFT configuration selected above was applied to the 1.5B, 3.2B and 6B base models. All three runs used batch size 64, the same 2,048-token sequence length, the same instruction corpus, and the same ordering of training examples. This provides a paired comparison of model scale under an otherwise matched supervised fine-tuning procedure.

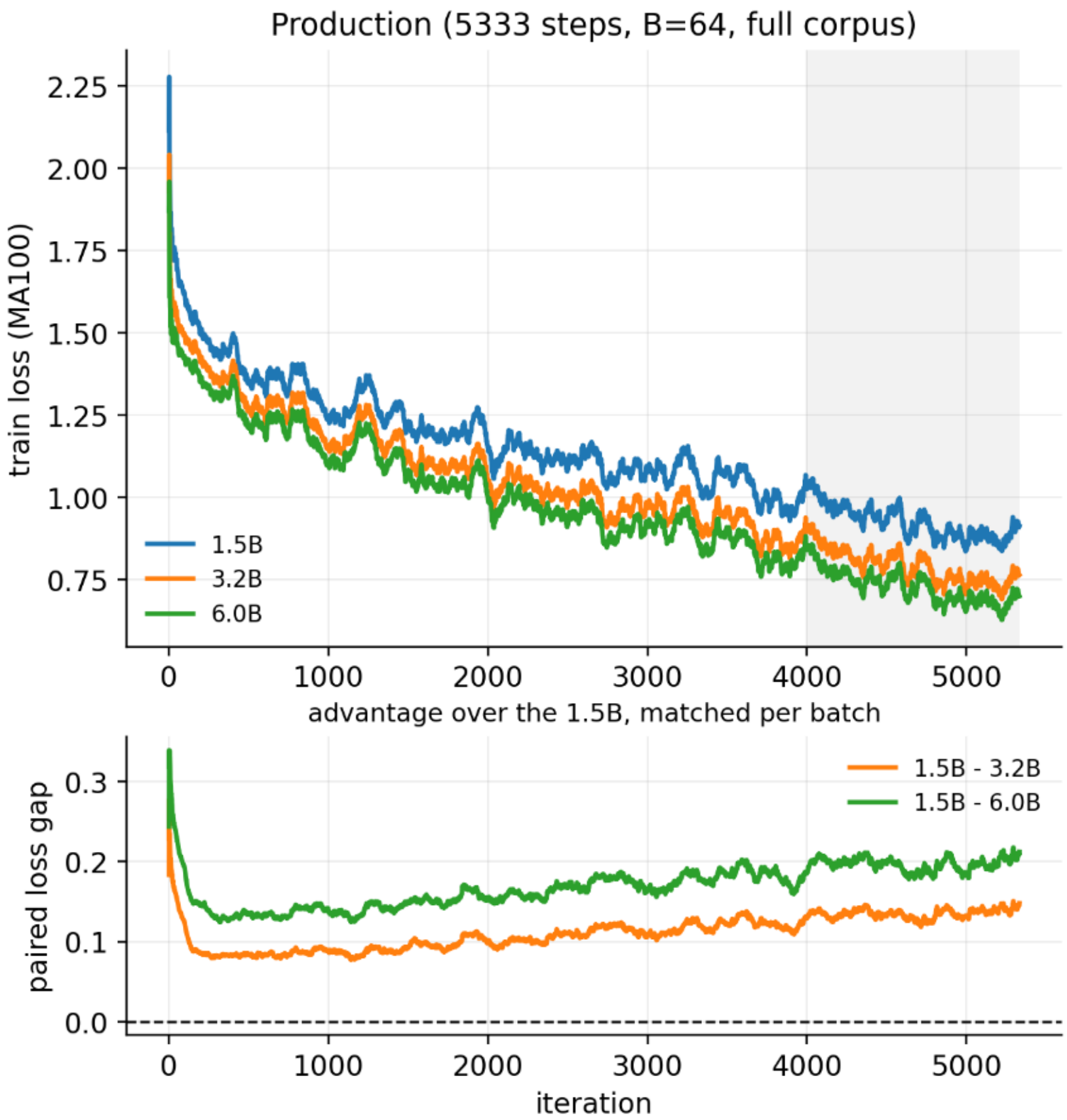

*Figure 27: Upper: 100-step moving-average training loss for the 1.5B, 3.2B and 6B production SFT runs over 5,333 optimization steps at batch 64. The shaded region marks the late-training window used for summary statistics. Lower: paired loss gap relative to the 1.5B model, computed on identical training batches. Positive values indicate lower loss for the larger model*

The absolute loss curves retain the expected ordering throughout training: 6B achieves the lowest loss, followed by 3.2B and then 1.5B. The paired representation in the lower panel is more informative because it removes most of the local variation caused by differences in batch difficulty. Since the models process the same examples in the same order, the gap directly tracks the advantage of additional capacity on matched training material.

| Comparison | Early gap (~iteration 500) | Late-training gap (~iteration 5,300) |
|---|---|---|
| 1.5B − 3.2B | ≈0.09 nats | ≈0.145 nats |
| 1.5B − 6.0B | ≈0.15 nats | ≈0.21 nats |

*Table 14: Paired SFT loss advantage of the larger production models relative to the 1.5B baseline at an early point in training and near the end of the 5,333-step production run. Values are approximate and are read from the paired loss trajectories.*

Two features are notable. First, the advantage of scale is not established only during the initial adaptation phase. Both paired gaps widen through training: the 3.2B advantage over 1.5B grows from approximately 0.09 to 0.145 nats, while the 6B advantage grows from approximately 0.15 to 0.21 nats. Larger models therefore continue to extract additional benefit from the same supervised stream as fine-tuning proceeds.

Second, the return to additional parameters diminishes across the three scales. Increasing model size from 1.5B to 3.2B reduces late-training loss by approximately 0.145 nats. Increasing again from 3.2B to 6B yields only about 0.065 nats of additional improvement relative to the 1.5B baseline. The ordering is therefore monotonic, but the incremental gain from the second increase in scale is substantially smaller.

We deliberately do not fit a scaling law to these three observations. A three-parameter law cannot be identified from only three model sizes, while a two-parameter fit would absorb any irreducible-loss term into the apparent exponent and would therefore describe only an effective slope over a narrow interval. The present data support the qualitative conclusions that larger models achieve lower SFT loss under matched training and that the marginal gain decreases across the tested scale range; they do not support a general SFT scaling exponent.

The widening paired gaps also indicate that the three production runs had not simply reached a common optimization regime separated by a fixed offset. Capacity continues to matter throughout the observed training horizon. Because Section 6.9 shows that loss was still decreasing at termination, these late-training gaps should be interpreted as properties of the chosen 5,333-step horizon rather than as differences between fully converged SFT optima.

**6.9 Convergence and residual structure.** The production SFT runs were stopped after 5,333 optimization steps, corresponding to 341,312 example presentations, or approximately 1.9 effective passes over the 178,502-entry dataset. Because the curriculum was continuously reshuffled and re-interleaved, repeated examples were not encountered in the same sequence or local neighborhood on subsequent exposure.

The stopping point was not selected by formal convergence criterion. Like previously mentioned, this first SFT campaign was exploratory, and training was terminated after approximately two effective passes to obtain stable production checkpoints while avoiding unnecessarily prolonged repetition of a relatively small synthetic corpus. The intention is to treat this run as the first complete iteration of the Neurograf instruction-tuning pipeline rather than as an attempt to exhaustively optimize the final SFT horizon.

Training loss was still decreasing when the runs were stopped. The reported production checkpoints should therefore not be interpreted as fully converged optima, and the final loss differences discussed in Section 6.8 are specific to the 5,333-step training horizon. Additional training may have reduced loss further, but whether this would translate into improved instruction-following quality rather than increasing memorization or overfitting was not evaluated.

The production curves also contain repeated local structure at approximately corresponding training positions across all three model sizes. Because the 1.5B, 3.2B and 6B runs received the same examples in the same order, these shared rises and falls are more plausibly associated with variation in the training data than with model-specific optimization effects.

A likely source is the interleaved curriculum described in Section 6.4. Category streams differ substantially in size, and as smaller streams become exhausted the local mixture of task types can change. Similar shared structure was observed during pre-training, where all models consumed the same RoSfera master-shard sequence and local changes in source composition produced corresponding features in the loss traces. We did not trace individual SFT loss features back to specific categories, so this interpretation remains qualitative.

We therefore end with three practical conclusions. First, the present SFT checkpoints represent useful but not fully optimized operating points: loss remained in decline at termination, and the optimal number of effective passes was not determined. Second, the repeated structure in the paired loss curves indicates that the composition and ordering of the instruction stream remain visible in optimization even when examples are aggressively interleaved. Third, the models were successfully fine-tuned to follow user instructions and act as assistants. Despite several failure modes and some rough edges, they can interact directly with users and occasionally produce surprisingly accurate responses (see Appendix C).

It is precisely this potential observed in the production of SFT models that motivates a future iteration of the instruction-tuning pipeline. A more systematic SFT study would jointly optimize batch size, learning rate and training duration, and would evaluate alternative curriculum constructions designed to reduce composition drift. We are ea a future work.

## 7. Embedding Models

The Neurograf base models were trained as causal language models, but their internal representations can also be adapted for semantic similarity and retrieval. Rather than training a separate encoder architecture from scratch, we investigated whether the smaller Neurograf decoder models could be converted into Romanian embedding models through contrastive post-training.

The 160M and 560M pretrained checkpoints were used as starting points. For an input text, the final hidden states produced by the causal decoder are mean-pooled over all non-padding tokens and the resulting vector is L2-normalized. No separate projection head is added: the embedding dimensionality is therefore identical to the hidden width of the underlying language model. The same transformer is used to encode both members of a semantic pair.

Contrastive post-training updates the complete pretrained model rather than only an embedding head. The objective is to increase similarity between texts derived from the same semantic source while separating them from unrelated texts sampled within the same batch. This provides a lightweight route from the generative models of Chapter 4 to models that can represent Romanian text in a continuous semantic space.

The experiments in this chapter are exploratory. The training pairs were derived largely from the synthetic pipeline developed for supervised fine-tuning rather than from a dedicated retrieval corpus, and the resulting contrastive task is substantially easier than realistic large-scale retrieval. We therefore treat the training accuracy primarily as a measure of contrastive convergence and representation separation rather than as a benchmark of production retrieval quality.

**7.1 Contrastive pair construction.** The embedding training data was derived primarily from the topic-seeded generation cascade described in Section 6.1. Each accepted topic seed produces several related textual representations: the original short seed, an expanded_seed, a substantially longer infused_seed, and question–answer material generated from the same topic.

These related representations form natural positive pairs. The training pipeline admits five principal pair types:

*seed ↔ expanded_seed;*

*expanded_seed ↔ infused_seed;*

*seed ↔ infused_seed;*

*question ↔ answer for closed question answering;*

*question ↔ answer for open question answering.*

Pairs are not exhaustively enumerated at every training step. Instead, the data are organized into semantic clusters associated with their common originating topic. During batch construction, topics are sampled, individual clusters are selected within those topics, and one available semantic relationship is sampled from each selected cluster. This creates a dynamically changing contrastive stream while maintaining topical diversity within individual batches.

Approximately 1.2 million semantic pairs were available to the contrastive training process. Positive examples share an explicit generative relationship: an expansion is produced from its seed, an infused passage from the expansion, and question–answer pairs are generated from the same underlying material.

This construction is economical because it reuses the synthetic data pipeline already required for instruction tuning, but it also makes the contrastive problem easier than realistic retrieval.

In particular, many positive pairs share substantial lexical content. An expanded_seed is directly generated from its seed, while an infused_seed is generated from the expanded representation. A model can therefore exploit vocabulary overlap, named entities and local phrasing in addition to deeper semantic equivalence. Question–answer pairs reduce this overlap somewhat, but they remain generated from tightly related source material.

Negatives are supplied by other members of the batch. In the final training configuration, each optimization step compares a positive pair against 47 in-batch alternatives, giving a global batch of 48 semantic pairs. The batching procedure encourages topic diversity, so these negatives are generally unrelated at the level of subject matter.

No explicit hard-negative mining [35] was performed. The corpus therefore does not systematically provide examples such as two passages about the same subject that differ in a small but important factual or semantic detail. Distinguishing such near-neighbors is central to realistic retrieval systems, but it is substantially more difficult than separating topically unrelated documents.

The contrastive objective used here should consequently be interpreted as a representation-learning task rather than as a complete retrieval benchmark. High in-batch accuracy demonstrates that the model can identify the corresponding member of a semantic pair among 47 mostly unrelated alternatives; it does not establish comparable performance against thousands or millions of semantically overlapping candidates.

This distinction becomes important later in the chapter. Both models eventually approach saturation on the in-batch task, so the resulting accuracy is more useful for studying training dynamics and the evolution of the representation space than for ranking the final checkpoints as retrieval systems.

**7.2 Embedding architecture and training setup.** Contrastive post-training was applied to the 160M (d6) and 560M (d12) Neurograf base models, initialized from their corresponding pre-trained checkpoints. Both retain the original causal decoder architecture and the 2,048-token maximum context length; no separate encoder network or projection head is introduced.

For a text containing (T) valid tokens, the final hidden states of the decoder are mean-pooled [31]:

$$\frac{1}{T}\sum_{t=1}^{T} h_t \tag{13}$$

where $h_t$ is the final-layer hidden representation at token position (t). Padding positions are excluded from the mean. The pooled representation is then L2-normalized:

$$\frac{z}{\|z\|_2} \tag{14}$$

so similarity between two embeddings can be computed directly from their dot product as cosine similarity.

Both members of a positive pair are encoded by the same transformer weights. Training therefore uses a shared single-encoder architecture rather than separately parameterized anchor and document towers. The complete language model is updated during contrastive post-training.

The training objective is symmetric InfoNCE [32]. For a batch of (B) positive pairs, the normalized anchor and positive embeddings form a B* B similarity matrix. The diagonal contains the designated positive matches, and the remaining entries act as in-batch negatives. Cross-entropy is evaluated in both directions (anchor-to-positive and positive-to-anchor) and the two losses are averaged.

The final training configuration used a global batch of 48 semantic pairs distributed across four H100 GPUs, corresponding to 12 pairs per rank. Every pair is consequently contrasted against 47 alternatives within its global comparison set. Chance top-1 retrieval accuracy under this setup is 0.02083.

Both the 160M and 560M models were trained for 100,000 optimization steps. This corresponds to 4.8 million pair presentations, or approximately four dataset-equivalent passes relative to the roughly 1.2 million available semantic pairs. Because pairs were sampled dynamically from semantic clusters rather than traversed as a single fixed flat list, this quantity is better interpreted as an effective exposure count than as four identical epochs.

The average sequence length during training was approximately 157 tokens, substantially below the 2,048-token model context limit. Variable-length texts were padded for batching while padding positions were excluded from the mean pooling operation.

| | **160M (d6)** | **560M (d12)** |
|---|---|---|
| Training steps | 100,000 | 100,000 |
| Global batch | 48 pairs | 48 pairs |
| GPUs | 4 × H100 | 4 × H100 |
| Mean sequence length | ≈157 tokens | ≈157 tokens |
| Median throughput | ≈235,000 tok/s | ≈68,000 tok/s |
| Timed compute | 113.0 min | 416.1 min |
| GPU-hours | 7.5 | 27.7 |

*Table 15: Training configuration and measured computational cost of the two Neurograf embedding models.*

The training runs were interrupted at several points for checkpoint inspection and qualitative testing and subsequently resumed. These job boundaries produce small discontinuities in the logged trajectories and should not be interpreted as optimization instabilities.

### 7.3 Contrastive training dynamics.

Figure 28 summarizes the evolution of the two embedding models over the complete 100,000-step contrastive training run. Because the positive-pair construction is relatively easy and the candidate pool contains only 48 alternatives, no single metric provides a complete description of learning. We therefore track loss, top-1 in-batch retrieval accuracy, positive and negative similarity, and the separation between the two similarity populations.

The overall trajectories show three distinct phases.

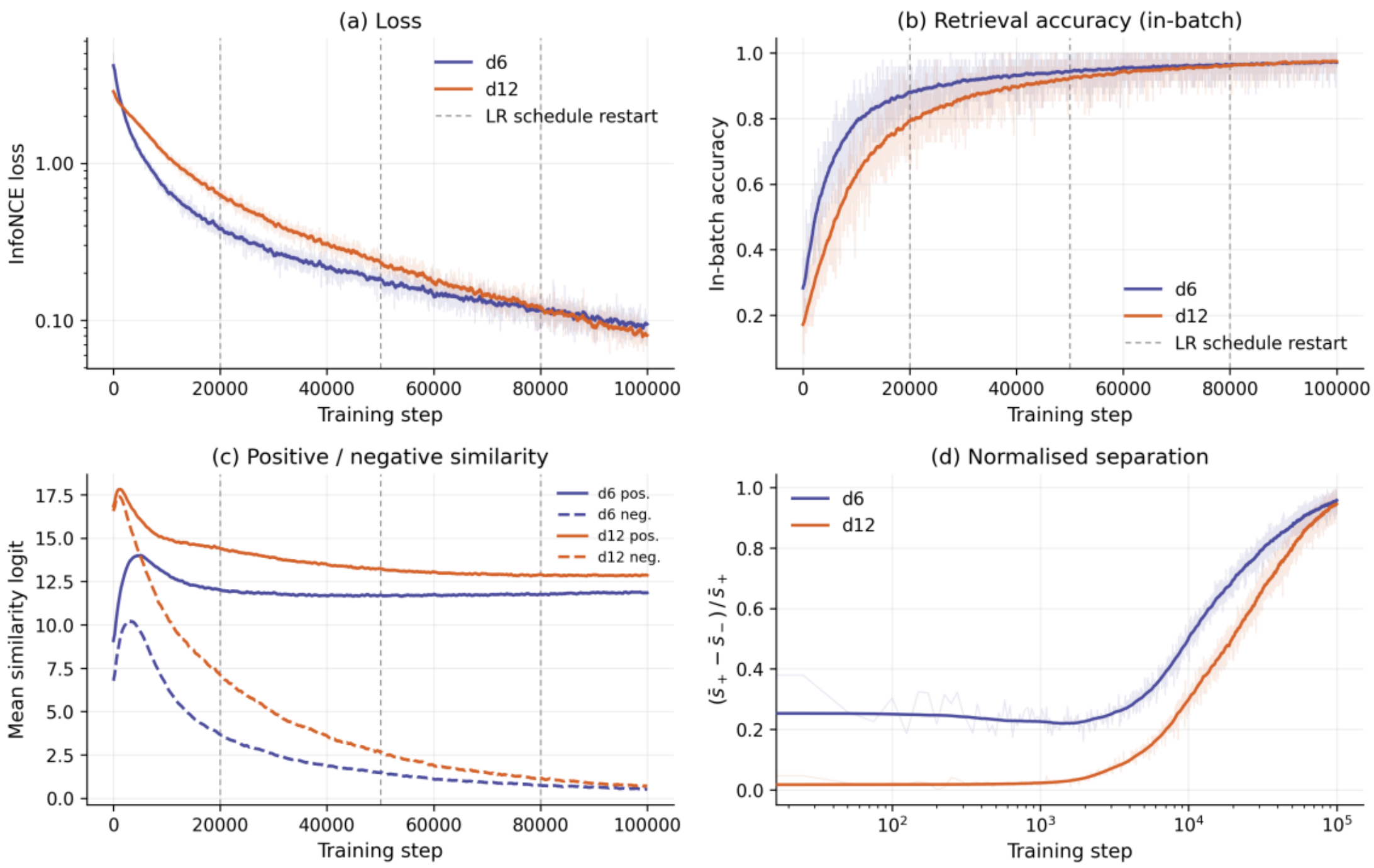

*Figure 28: Contrastive training dynamics for the 160M and 560M Neurograf embedding models. (a) Symmetric InfoNCE loss. (b) Top-1 retrieval accuracy against 47 in-batch alternatives. (c) Mean similarity logits for designated positive pairs and in-batch negative pairs. (d) Relative positive–negative separation, shown against a logarithmic step axis to expose the early training regime.*

During the earliest phase, positive and negative similarities remain poorly separated. In particular, both similarities can initially move in the same direction, so decreasing contrastive loss should not automatically be interpreted as the emergence of a well-structured semantic space. The separation metric in panel (d) makes this early regime more visible than loss alone.

A second phase begins after several thousand optimization steps, when positive and negative similarities start to diverge more clearly. In-batch retrieval accuracy rises rapidly during this period, and the larger difference between the two similarity populations indicates that the model is beginning to organize the representation space around the contrastive objective.

Finally, both models enter a late regime in which top-1 accuracy approaches the ceiling of the 48-candidate task. Improvements in loss and similarity geometry continue after accuracy becomes relatively insensitive, illustrating why the saturated retrieval metric alone is insufficient for comparing the final checkpoints.

The two model sizes do not traverse these phases at the same rate. The 160M model develops useful positive–negative separation earlier and consequently leads the 560M model during much of the initial and intermediate training period. The 560M representation begins from a substantially less separated state and requires more optimization before comparable contrastive structure emerges. The larger model eventually catches and slightly exceeds the smaller model on the tracked end-state quantities.

This difference is particularly interesting because the ordering is opposite to that observed for generative pre-training loss: the 560M checkpoint is the stronger language model before contrastive adaptation, yet its raw representations initially provide poorer separation for this embedding task. The next sections examine this initialization effect directly.

**7.4 Initial representation collapse.** The most striking result in the embedding experiments appears before substantial contrastive learning has taken place. The two pretrained language models enter the embedding task with markedly different levels of positive-negative separation, and the larger model begins from the poorer representation state.

At initialization, the mean similarity of unrelated in-batch texts in the 560M model is approximately 98.2% of the mean similarity of designated positive pairs. For the 160M model the corresponding ratio is approximately 75.3%. The raw representations of the larger base therefore distinguish positive from negative pairs only weakly under the contrastive construction used here.

| **At initialization** | **160M** | **560M** |
|---|---|---|
| Negative-to-positive similarity ratio | 0.753 | 0.982 |
| In-batch top-1 accuracy | 0.292 | 0.188 |

*Table 16: Initial contrastive geometry of the pretrained Neurograf bases. Similarity values refer to the relative separation of designated positive pairs and in-batch negatives and should not be interpreted as a direct global measurement of embedding-space isotropy.*

The difference is substantial. In the 560M checkpoint, positive and negative examples initially occupy nearly the same similarity range, whereas the 160M model begins with considerably more structure already useful to the contrastive task. The difference is also visible in top-1 retrieval accuracy: before contrastive adaptation, the 160M model identifies the correct member of the 48-candidate set approximately 29.2% of the time, compared with 18.8% for the 560M model. Both remain well above the 2.1% chance level, indicating that the pretrained representations are not entirely uninformative.

This ordering is notable because it is opposite to the ordering obtained from generative pre-training loss. On the common 131-billion-token pre-training stream, the 560M model reaches a validation loss of approximately 0.743 bits per byte, compared with 0.887 for the 160M model. The larger model is therefore clearly superior on the language-modeling objective while simultaneously providing less useful raw separation for the contrastive embedding task.

Generative pre-training quality and embedding geometry should consequently not be treated as interchangeable quantities. A lower next-token prediction loss does not imply that mean-pooled decoder representations will immediately form a useful similarity space.

The observed behavior is compatible with the broader literature on anisotropic or collapsed language-model representations [33-35], but the present experiment does not directly measure global embedding-space isotropy. Our metric compares positive and negative similarity under a particular contrastive pair construction. We therefore describe the 560M initialization more conservatively as poorly separated or near-collapsed with respect to the contrastive task, rather than treating the measured ratio as an independent anisotropy statistic.

There is also an important confound. The two models differ not only in parameter count but also in the amount of pre-training exposure per parameter. The 160M model consumed approximately 791 tokens per parameter on the fixed 131-billion-token stream, whereas the 560M model consumed approximately 224, a factor of roughly 3.5 less. With only two model sizes, the observed difference cannot distinguish an effect of scale from an effect of much heavier relative over-training of the smaller model.

The result should therefore not be generalized as evidence that larger language models necessarily produce more collapsed embeddings. What the experiment establishes is narrower: for these two Neurograf checkpoints, the stronger generative model begins contrastive post-training from the less separated embedding state.

**7.5 Escaping the initial collapse.** Contrastive training progressively removes the poor positive–negative separation present at initialization, but the transition is not immediate. Both models spend an initial phase in which the separation metric changes little before entering a regime of rapid contrastive organization.

The 160M model begins to leave its initial plateau after roughly 2,000 optimization steps, whereas the 560M model remains near its starting state for closer to 3,000 steps. The larger model therefore not only begins from poorer separation but also requires more contrastive updates before a clear distinction between positive and negative examples emerges.

This difference persists through much of the early training trajectory. At approximately step 10,000, the 160M model reaches an in-batch top-1 accuracy of about 0.813, compared with approximately 0.667 for the 560M model. The smaller model consequently learns useful contrastive geometry more rapidly despite having the weaker pre-training loss.

Panel (c) of Figure 28 clarifies what happens during the transition. During the first few thousand steps, the mean similarity of both positive and negative pairs can rise together. The model is therefore not initially improving simply by pushing negatives away from positives. Instead, the overall similarity structure changes before a sustained separation between the two populations develops.

For this reason, the early InfoNCE loss is potentially misleading when read in isolation. The 560M model begins with a lower loss, approximately 2.694, than the 160M model at approximately 5.025, yet its initial retrieval accuracy is also lower. A nearly uniform prediction over the 48 candidates corresponds to a cross-entropy of

$$\ln(48) \approx 3.87 \tag{15}$$

The 560M model begins comparatively close to this low-confidence regime. By contrast, the 160M model makes more strongly differentiated predictions from the outset; when those predictions are wrong, they incur a larger cross-entropy penalty. A lower initial contrastive loss therefore does not necessarily imply a better representation space.

The more informative early diagnostics are the positive-negative similarity separation and retrieval accuracy. These show that the 160M model begins with more useful structure, while the 560M model requires a longer adaptation period before its larger capacity becomes advantageous.

The ordering eventually reverses. As training proceeds, the 560M model continues improving after the 160M model has approached the ceiling of the 48-candidate task, and both ultimately reach high separation and high in-batch retrieval accuracy.

A qualitative inspection performed during training also indicated that the representation space became progressively more dispersed as contrastive adaptation proceeded. Because that intermediate probe used a non-final checkpoint and a manually selected heterogeneous passage set, we do not treat it as an evaluation result. A more controlled comparison using the final 100,000-step checkpoints and a fixed set of paired Romanian passages is presented in Section 7.7.

**7.6 Compute cost and metric saturation.** The larger embedding model finishes ahead on all tracked end-state quantities, but at substantially higher computational cost. The 560M model reaches a final InfoNCE loss of approximately 0.069 and an in-batch top-1 accuracy of approximately 0.98, compared with 0.088 and 0.96 for the 160M model.

This improvement requires approximately 3.7 times more compute. The 160M run consumed about 7.5 H100 GPU-hours, while the 560M run consumed approximately 27.7 GPU-hours. Median throughput decreased from roughly 235,000 tokens/s to 68,000 tokens/s on the same four-H100 configuration.

At matched training compute, the comparison favors the smaller model more strongly. After approximately 113 minutes, the total measured compute time of the completed 160M run, the 560M model has processed only about 27,000 optimization steps and reaches an in-batch accuracy of roughly 0.87, while the 160M model has completed its full training schedule and reaches approximately 0.96.

It would nevertheless be misleading to interpret the two-percentage-point final accuracy difference as a reliable measure of the larger model's retrieval advantage. The evaluation task itself is close to saturation. Both models remain above approximately 0.95 top-1 accuracy for much of late training, and by around step 60,000 the 48-candidate in-batch metric provides little remaining dynamic range.

The same saturation affects the interpretation of training duration. Approximately the final 40% of the 100,000-step runs produces little measurable improvement in the quantity being used as the principal retrieval diagnostic. Loss and representation-separation metrics continue to evolve, but top-1 accuracy against 47 mostly unrelated negatives is already near its ceiling.

If the experiment were repeated, the available compute would therefore be better spent on a harder evaluation and training objective rather than simply extending the same contrastive schedule. Larger in-batch candidate pools, explicit hard-negative mining, or evaluation against thousands of independently sampled Romanian passages would provide substantially greater resolution between checkpoints.

The practical conclusion of this first embedding study is consequently modest. Both Neurograf base models can be converted into useful contrastive representation models using the synthetic semantic structure already produced by the instruction-data pipeline. The 160M model learns the task faster and is markedly more compute-efficient, while the 560M model eventually achieves the strongest tracked end state. Because the evaluation saturates, however, the magnitude of the larger model's final retrieval advantage remains unresolved.

**7.7 Qualitative semantic geometry at the final checkpoints.** The in-batch retrieval metric used during contrastive training approaches saturation for both embedding models and therefore provides only limited resolution between the final checkpoints. To obtain a more interpretable qualitative view of the learned representation spaces, we constructed a small, fixed set of ten Romanian passages grouped into five manually designed semantic pairs: physics, drone engineering, cooking, nature, and public administration.

Each pair contains two independently phrased passages describing the same broad subject, while passages belonging to different pairs are deliberately unrelated.

The purpose of the probe is not to estimate retrieval performance, but to test whether semantically corresponding passages occupy visibly similar regions of the learned embedding space.

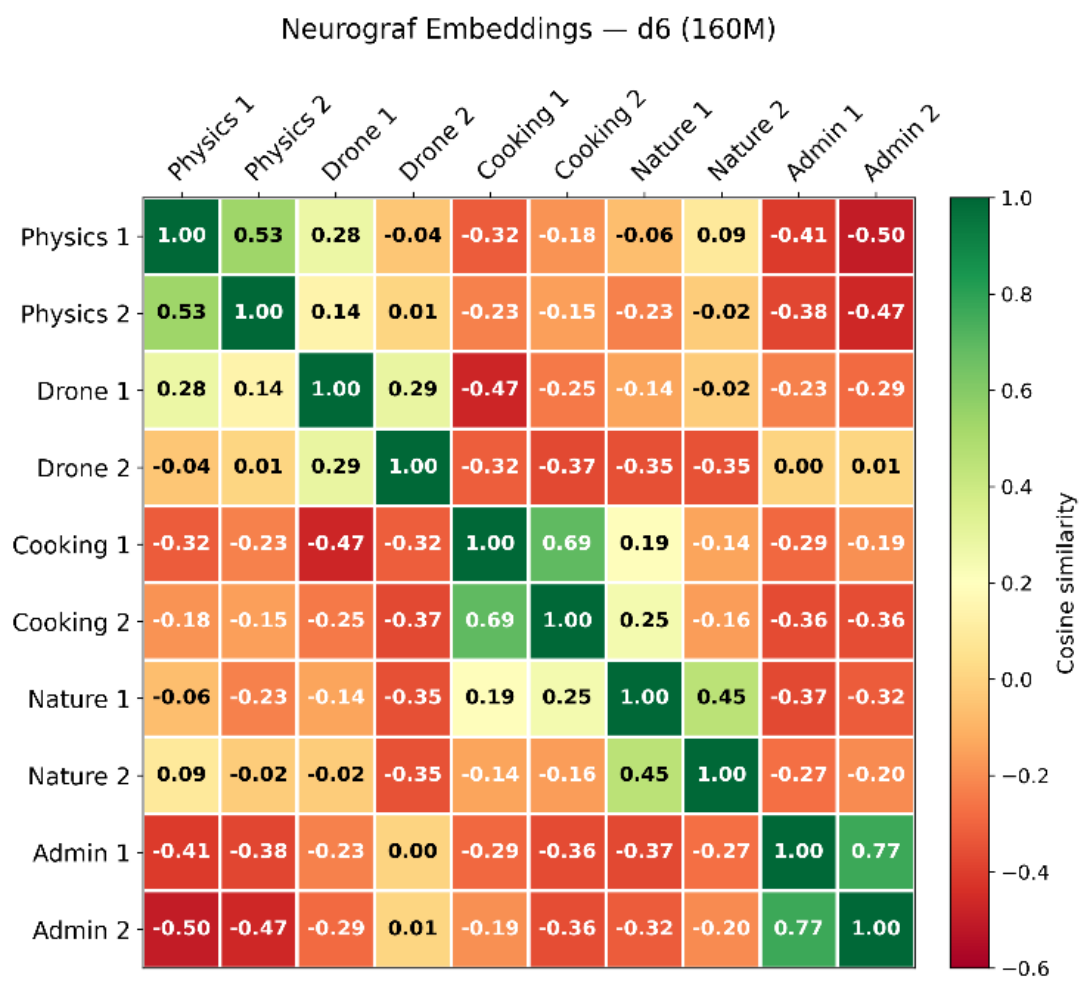

*Figure 29: Sample-mean-centered cosine similarity matrix for the final 100,000-step 160M embedding model. The ten Romanian passages form five manually constructed semantic pairs. Higher values indicate stronger relative similarity within the fixed probe set.*

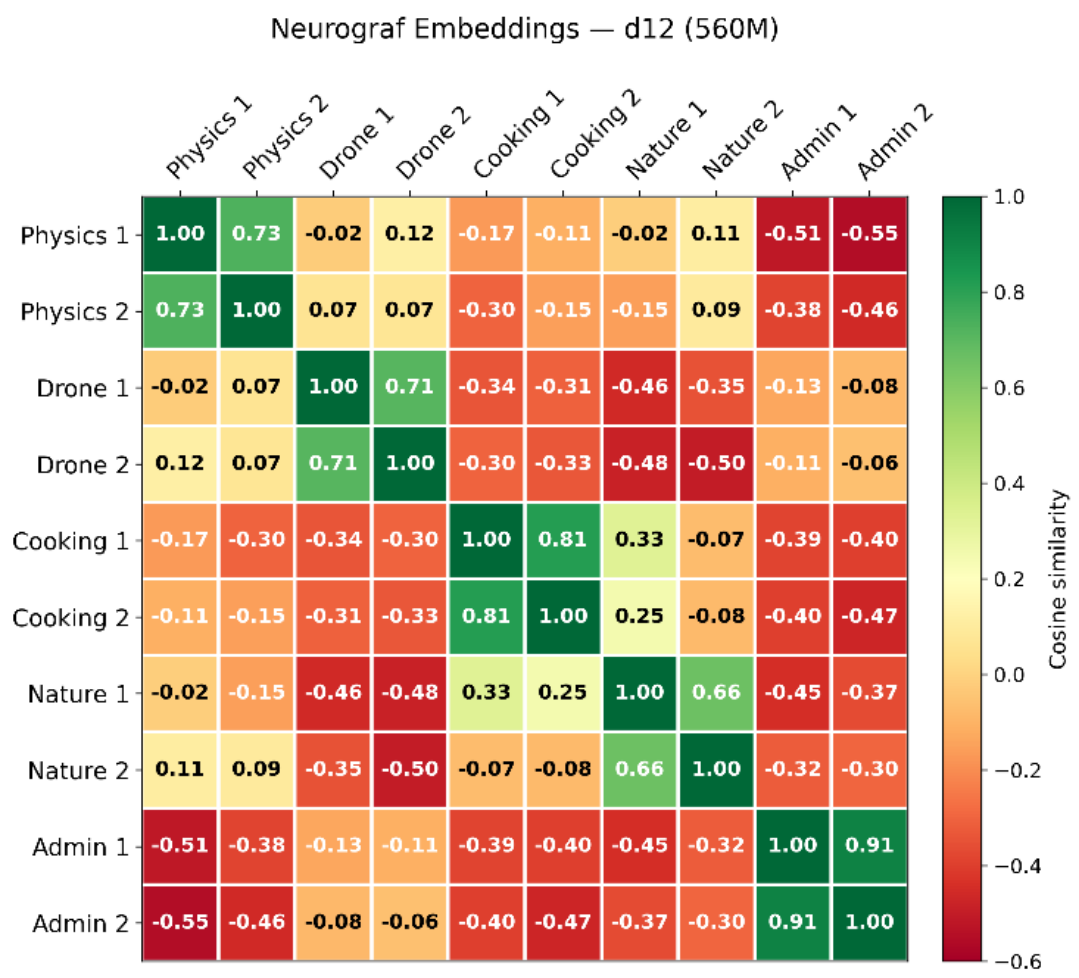

*Figure 29: Sample-mean-centered cosine similarity matrix for the final 100,000-step 560M embedding model, evaluated on the same ten Romanian passages and using the same post-processing procedure as Figure 29.*

For each passage, the final hidden states were mean-pooled and the resulting document representation was converted to FP32. To emphasize relative geometry within the fixed ten-passage set, the mean embedding of the ten passages was subtracted from every document vector before L2 normalization. The cosine matrices shown in Figures 29 and 30 therefore represent sample-mean-centered cosine similarity, rather than the raw cosine values that would ordinarily be used directly in a retrieval system. Negative values should consequently be interpreted as relative separation within this fixed set, not as a universal threshold for semantic dissimilarity.

The intended semantic pairs are clearly visible in both matrices, although the corresponding similarities are substantially stronger in the larger model.

| **Semantic pair** | **160M** | **560M** |
|---|---|---|
| Physics | 0.532 | 0.726 |
| Drone engineering | 0.291 | 0.711 |
| Cooking | 0.690 | 0.814 |
| Nature | 0.450 | 0.659 |
| Public administration | 0.768 | 0.912 |
| Mean | 0.546 | 0.764 |

*Table 17: Sample-mean-centered cosine similarities for the five manually constructed semantic pairs evaluated with the final 160M and 560M embedding checkpoints. The mean row summarizes the designated pair similarities only. These values are qualitative and depend on the composition of the fixed ten-passage probe set.*

For both models, the intended semantic partner is the highest-similarity non-self passage for all ten individual passages. The mean centered similarity of the five designated semantic pairs increases from 0.546 in the 160M model to 0.764 in the 560M model. By comparison, the mean similarity across the remaining non-matched passage combinations is approximately -0.187 for the 160M model and -0.212 for the 560M model.

The resulting separation between matched and non-matched passages therefore increases from approximately 0.733 in the 160M representation space to approximately 0.976 in the 560M space. The effect is particularly pronounced for the drone-engineering pair, whose similarity increases from 0.291 to 0.711, while the public-administration pair reaches 0.912 in the larger model.

These results provide a useful qualitative complement to the saturated in-batch accuracy reported earlier. The 160M model already exhibits a coherent semantic organization, but the final 560M checkpoint produces a more strongly differentiated local geometry on this fixed probe: related passages are drawn closer together, while most passages belonging to different semantic categories remain separated after centering.

The experiment should nevertheless be interpreted conservatively. The ten passages were manually constructed, the five semantic relationships were defined in advance, and the sample size is far too small to estimate general retrieval performance. The centering operation additionally makes the absolute cosine values dependent on the composition of this particular passage set. The matrices therefore demonstrate local semantic organization rather than retrieval accuracy, calibration, or global embedding-space geometry.

A larger evaluation with natural Romanian documents, substantially larger candidate pools, graded relevance judgments and hard negatives would be required to determine whether the stronger separation observed in the 560M model translates into a practical retrieval advantage. Within the scope of the present experiment, however, the final checkpoints demonstrate that contrastive post-training converts both causal Neurograf models into representation models capable of recovering manually defined semantic relationships, with visibly stronger separation in the larger encoder.

## 8. Conclusions and Future Work

Neurograf was developed as an end-to-end experiment in building a Romanian-first language-model ecosystem: corpus construction, tokenizer design, generative pre-training, supervised instruction tuning and contrastive representation learning were all performed within a common Romanian-centered pipeline. The objective was not to reproduce the capabilities of frontier multilingual systems, but to determine what changes when Romanian is treated as the primary language of the system rather than as one component of a much larger multilingual mixture.

The resulting experiments suggest that this distinction matters at almost every layer.

**8.1 Lessons from a Romanian-first model family.** The first constraint is data. RoSfera reaches approximately 635 GB and 92.4 billion words, but its scale should not be confused with uniform quality or coverage. Most of the available material remains web-derived, while clean literary, encyclopedic and institutional sources constitute comparatively small fractions of the mixture. Missing diacritics are widespread in the dominant web sources, and the corpus contains large differences in lexical productivity, extraction quality and internal heterogeneity between registers.

The oversampling experiment makes the limitation quantitative. Even aggressive repetition of the smallest high-quality sources changes the overall mixture only marginally before repetition itself becomes undesirable. For a corpus with this degree of web dominance, improving, replacing or extending the dominant web component is therefore a more powerful intervention than repeatedly sampling the relatively small amount of curated Romanian material.

This observation has consequences beyond corpus construction. A Romanian foundation model trained predominantly on Romanian web text should not be expected to acquire comprehensive mathematical, scientific, technical or specialized professional knowledge simply by increasing parameter count. The model can learn the language and statistical structure present in the corpus extremely well while remaining knowledge-limited by what the corpus contains.

For the intended deployment regime of Neurograf, this is not necessarily a fatal limitation. Many useful Romanian applications are narrower than unrestricted question answering: document retrieval, administrative assistance, form completion, classification, summarization, guided interaction with institutional procedures, and question answering over a defined collection of documents. In these settings, a compact Romanian-native model combined with an external knowledge system may be more useful than attempting to encode every relevant fact into the model parameters.

Retrieval-augmented generation [36] therefore appears to be a natural component of any production Neurograf system. The language model supplies Romanian language competence and instruction following, while a retrieval layer supplies current or domain-specific information from controlled sources. This division is particularly attractive for bureaucratic and institutional deployments, where authoritative documents already exist and where provenance and updateability are more important than memorized general knowledge.

Tokenizer design provides a second lesson. Romanian and English are not symmetric under vocabulary mismatch. In the controlled FineWeb experiment, the Romanian-trained vocabulary approaches the efficiency of a native English tokenizer when encoding English as vocabulary size increases, reaching 96.3% of native efficiency at the largest common comparison point. The reverse direction remains substantially worse: an English-trained vocabulary retains large penalties on Romanian internet, legal and literary material.

Part of this asymmetry may arise because the nominally Romanian training corpus itself contains English and other non-Romanian material, and the experiment cannot isolate that contribution from genuinely language-specific effects. Nevertheless, from a deployment perspective the asymmetry is useful. A Romanian-first vocabulary does not necessarily require complete isolation from English. The resulting tokenizer can retain substantial ability to represent English while avoiding the much larger cost incurred when Romanian is forced through a predominantly English vocabulary.

This suggests a practical direction for future corpus design. Romanian should remain the dominant language, but carefully selected English-language material could be added where it contributes capabilities that Romanian web data cannot supply economically, for example programming code and high-quality scientific or technical material. The objective would not be to convert Neurograf into a balanced bilingual model, but to extend the competence of a Romanian-first model using targeted high-value data.

The tokenizer experiments also show that vocabulary size cannot be selected from convention alone. Romanian progressively closes and eventually reverses the native compression gap with English as vocabulary increases, while its low-frequency merge inventory extends much further. At the same time, large vocabularies impose a substantial parameter cost on small models through the embedding and output matrices. The 80k production vocabulary is therefore best understood as a family-level engineering compromise rather than a universal optimum for Romanian.

The pre-training experiments provide the clearest evidence that useful Romanian base models can be trained at moderate scale. All five models learned stably from the same 131-billion-token stream, and validation loss improved monotonically with capacity from 160M to 6B parameters. The fixed-data scaling fit shows that additional parameters continue to help, but with rapidly diminishing returns when the dataset is held constant. The estimated compute-optimal relation similarly assigns a growing fraction of additional compute to token exposure rather than parameter count across the measured regime.

The practical implication is that a future Neurograf should not simply increase parameter count while replaying the present RoSfera stream. A somewhat larger model combined with a larger and improved corpus is a more defensible direction than aggressive scaling on unchanged data. The present scaling measurements provide a quantitative starting point for choosing that balance rather than relying entirely on scaling prescriptions derived from English-dominated model families.

The supervised fine-tuning stage proved substantially less mature than pre-training. The synthetic pipeline succeeded in producing a large amount of Romanian-native instruction material and avoided dependence on translated English instruction sets, but synthetic generation also imposed a ceiling on diversity and reliability. The generated data is topic-grounded rather than fact-grounded, and the generator determines both the user requests and the assistant responses. Errors, stylistic regularities and behavioral biases of the generator can therefore propagate directly into the assistant models.

Several practical effects became visible during SFT. Homogeneous conversations produced instruction-format lock-in, motivating deliberately mixed-task conversations. Training quality was strongly sensitive to the batch/learning-rate regime. Loss was still decreasing when the production runs were stopped, and the optimal number of passes was not established. The resulting assistants are therefore useful experimental checkpoints rather than fully optimized production systems.

The largest opportunity for improving a future Neurograf assistant is consequently not simply more synthetic generation. High-quality, naturally occurring Romanian conversational and instructional data would be particularly valuable. Real interactions contain changes of intent, incomplete requests, corrections, ambiguity, colloquial phrasing and conversational structure that are difficult for a generator to reproduce without inheriting its own regularities. Even a much smaller amount of carefully curated human Romanian dialogue could therefore have disproportionate value when combined with a larger synthetic corpus.

The 2,048-token training context is another practical limitation. It is sufficient for short interactions but restrictive for document-oriented assistance, long administrative forms, multi-document retrieval and sustained conversation. A production model should extend context substantially, most naturally through a RoPE-compatible context-extension strategy followed by continued training on appropriately long sequences. The extension should be evaluated on actual long-context Romanian retrieval and instruction tasks rather than inferred from positional extrapolation alone.

The embedding experiments provide a complementary result. Dedicated Romanian encoders were not required to obtain an initial semantic representation system: the existing pretrained Neurograf bases could be converted into embedding models by mean-pooling their hidden representations and applying contrastive post-training. This makes embedding models comparatively inexpensive additions to a language-model family once the expensive generative pre-training has already been performed.

The experiment also illustrates why generative loss and embedding quality should be treated separately. The 560M base was the stronger language model yet began contrastive training with poorer positive-negative separation than the 160M base. Contrastive adaptation eventually repaired this structure and the larger model achieved the strongest tracked end state, while the smaller model remained substantially more compute-efficient. Because the in-batch retrieval task saturated, however, these experiments establish feasibility rather than production retrieval quality. Future embedding work should use hard negatives and much larger retrieval candidate pools.

Taken together, these results suggest a particular role for Romanian-first language models. Their strongest justification is not competition with frontier general-purpose systems at unrestricted reasoning. It is the construction of efficient, locally deployable components whose tokenizer, language distribution, model scale and retrieval infrastructure are optimized for Romanian workloads.

For a workstation, institutional server or other constrained local hardware, an 10B-class model is substantially more realistic than a frontier-scale system while still offering enough capacity to benefit from the scaling behavior observed here. Combined with Romanian-efficient tokenization, retrieval over authoritative local documents, compact embedding models and tool access, such a system could support large numbers of routine interactions without requiring each user request to be sent to an external frontier API.

**8.2 Toward a production Neurograf.** The next Neurograf iteration should therefore be designed as a system rather than simply as a larger checkpoint.

A plausible production target is a model in the approximately 10B-parameter range (d36, 128 aspect ratio). The existing results do not establish that 10B is mathematically optimal; rather, this scale represents a practical next step beyond the measured 6B model while remaining compatible with local inference after appropriate quantization and deployment optimization.

The first requirement is a new RoSfera corpus. The objective should be not merely to add more Romanian web pages, but to improve the mixture. Better filtering of the dominant web component is likely to provide greater benefit than aggressive oversampling of small clean sources. Deduplication, language identification, diacritic quality and extraction failures should be revisited using the diagnostics developed in Chapter 2.

At the same time, the corpus should deliberately extend capability where Romanian-only data is structurally scarce. Selected English-language scientific and technical text and programming code are natural candidates, provided that Romanian remains the primary linguistic distribution. The tokenizer experiments suggest that the Romanian vocabulary can accommodate substantial English material more gracefully than an English vocabulary accommodates Romanian, although the extent to which this remains true under a new corpus mixture should be measured rather than assumed.

The second requirement is longer context. The current 2,048-token limit should be extended using a RoPE-compatible [37] method and followed by continued training on genuinely long examples. Long context is particularly relevant to the intended applications: legislation, administrative procedures, institutional documents and retrieval-augmented generation frequently require substantially more than 2,048 tokens of combined query, retrieved evidence and generated response.

Third, retrieval should become part of the primary model evaluation rather than an optional downstream experiment. The embedding models of Chapter 7 provide a starting point for a Romanian retrieval stack, but a production system requires hard-negative training, larger candidate pools and evaluation on realistic collections. The complete pipeline should test retrieval and generation together: document ingestion, Romanian query embedding, retrieval, context assembly, answer generation and source attribution.

Fourth, tool use should complement memorized knowledge. Mathematical competence is particularly difficult to obtain from a corpus that contains relatively little high-quality mathematical reasoning. Rather than expecting an 10B Romanian model to internalize reliable arithmetic and symbolic computation solely through pre-training, a production assistant should learn when and how to invoke deterministic tools. A Python execution environment is an obvious first target, followed by structured retrieval, database queries or other institutional tools where appropriate. The language model's role then becomes decomposition, tool selection and interpretation rather than unreliable internal calculation.

Finally, supervised fine-tuning is likely to remain the principal development bottleneck. The next instruction corpus should combine synthetic generation with substantially stronger human data. Real Romanian dialogues, service interactions, procedural questions and multi-turn corrections would provide behavioral diversity that synthetic generation alone did not reproduce in the present project. Synthetic data would remain useful for scaling coverage and filling targeted task categories, but should increasingly supplement rather than define the conversational distribution.

Evaluation must evolve with this change. Training loss is useful for optimizing SFT, but production quality ultimately depends on behaviors that token-level loss cannot fully capture: following changing instructions across turns, grounding answers in retrieved documents, refusing unsupported claims, producing structured output, using external tools correctly, and remaining reliable over long contexts. A future Neurograf should therefore define these tasks before training and maintain a held-out Romanian evaluation suite throughout development.

**8.3 Final perspective.** The central result of Neurograf is not a single model checkpoint. It is the demonstration that a Romanian-first language-model stack can be constructed end to end, and that doing so exposes engineering constraints that are difficult to see when Romanian is treated only as a small component of a multilingual system.

Neurograf should be read as a first complete iteration of that path. Its models and datasets are experimental, and several stages remain substantially short of production quality. At the same time, the experiments provide enough measurements to replace several initial assumptions with quantitative design choices. A future production-grade Neurograf can therefore begin not from scratch, but from the corpus, tokenizer, scaling, fine-tuning and representation lessons established here.

## Acknowledgements

The author gratefully acknowledges IFIN-HH/ELI-NP for providing access to the local H100 GPU computing infrastructure used for the training and evaluation experiments reported in this work. The author also thanks the ELI-NP ITSC Compartment, and in particular Mihai Ciubăncan, George Paisu and Dragoș Zabet, for maintaining the computing infrastructure and for the technical support that made these experiments possible.

The author also thanks Andrej Karpathy for his educational material and open-source contributions to language-model development. His materials, the nanoGPT implementations and later nanochat training pipeline were instrumental in the development of the experimental workflow that ultimately evolved into Neurograf.

ChatGPT, Claude, Gemini, GPT-OSS-120B and Gemma 3 were used during the project for code assistance, debugging, and language modeling. All experimental design decisions, model training, data preparation, data analysis, interpretation of results and final responsibility for this current work remain with the author.

## Appendix A: Romanian Text Curation

The corpus pipeline uses three filtering passes of increasing severity, referred to throughout as lite, medium and heavy (lite_curated, med_curated and heavy_curated). The passes are cumulative and strictly nested: medium is applied to the output of lite, and heavy to the output of medium, so that:

$$heavy \subseteq medium \subseteq lite \subseteq source \tag{16}$$

All three operate line by line rather than on whole documents. A line failing a condition is discarded; the surrounding block is retained. Blocks are delimited by blank lines, surviving lines keep their original order within a block, and blocks are rejoined with blank lines, so paragraph structure survives partial deletion. A block disappears only when all of its lines fail. Files reduced to empty are not written.

None of the three passes rewrites text. There is no normalization of spelling, no correction, no summarization, no reordering and no diacritic restoration. Every character in the output appears in the input, with the single exception of the repetition collapsing in A.1. The passes remove; they do not repair.

The heavy pass was used to produce the FuLG subset in RoSfera (Section 2.2). Its consequences for the corpus statistics are discussed in A.5 and should be read alongside Section 2.5.

**A.1 Lite curation.** The lite pass performs character-level cleaning and removes explicitly unwanted content, preserving the source text otherwise. Three operations:

- Character allowlist. Characters outside a Romanian-oriented allowlist are deleted. The allowlist comprises Latin letters including the five Romanian diacritics (ă, â, î, ș, ț) in both cases, digits, whitespace, and a defined set of common punctuation and symbols. Emoji, non-Latin scripts, mathematical and decorative Unicode, and control characters are removed. Deletion is character-wise: a line containing one emoji loses the emoji, not the line.

- Repetition collapsing. Runs of the same character longer than three are truncated to three, so !!!!!!!! becomes !!! and aaaaaaa becomes aaa. This is the only operation in any pass that alters retained text.
- Explicit-content removal. A line containing any term from a defined list of banned Romanian words and constructions is discarded in full. The offending term is not excised in place, because removal in place leaves a syntactically intact sentence whose content is unchanged in intent.

Blank lines are preserved as block separators; lines reduced to empty by cleaning are dropped.

```
for each line L:
    L ← delete characters not in ALLOWLIST(L)
    L ← collapse runs > 3 to 3
    if L contains any term in BANNED: discard L
    if L is empty and was not originally blank: discard L
```

**A.2 Medium curation.** The medium pass targets web boilerplate and non-prose material. It is applied to lite output and discards a line if any of the following hold:

- Length. The line is shorter than 8 characters.
- Boilerplate keywords. The line matches a list of web-structural markers: navigation labels, RSS and feed text, cookie and privacy notices, contact blocks, comment markers, social-media buttons, header and footer text, shopping and checkout terms, and Wikipedia markup.
- Structured identifiers. The line contains a price, a URL or bare domain name, or an alphanumeric token matching SKU-like patterns.
- Prose heuristic. The line contains no item from a predefined list of frequent Romanian verb forms (este, sunt, are, trebuie, poate, spune, scrie, and others). Lines with no finite verb are overwhelmingly fragments, list items, table cells and captions rather than sentences.

The prose heuristic is deliberately crude and is the only condition in this pass that discards well-formed text at a non-trivial rate; short verbless sentences and elliptical constructions are lost. It is retained because its precision on web fragments is high and the volume of Romanian web text is not the binding constraint at this stage.

```
for each line L in lite(text):
    if len(L) < 8: discard
    if L matches BOILERPLATE: discard
    if L contains PRICE | URL | SKU: discard
    if L contains no form in VERBS: discard
```

**A.3 Heavy curation.** The heavy pass is a high-precision sampling filter rather than a cleaning stage. Applied to medium output, it retains a line only if all of the following hold:

- Length is at least 40 characters.
- The line begins with an uppercase letter.
- The line ends in ., ! or ?.
- The line contains all five lowercase Romanian diacritics: ă, â, î, ş and ţ.
- The line contains no run of three or more repeated punctuation characters (!!!, ???, ..., ---).
- The line is not entirely uppercase.

Conditions 1–3, 5 and 6 select for complete, conventionally formatted sentences. Condition 4 is categorically stronger than the rest and dominates the retention rate: it requires a single line to exhibit the full Romanian diacritic inventory, which most correctly written Romanian sentences do not. The pass therefore discards a very large volume of perfectly valid Romanian prose on a criterion unrelated to its quality.

Several further conditions are present in the source but disabled, and do not affect any output reported in this work: requiring at least one comma, restricting numeric content to four-digit years, capping the number of uppercase-initial words per line, and rejecting lines above a punctuation-density threshold. We list them for reproducibility, since the code is released with them in place.

```
for each line L in medium(text):
    keep L iff  len(L) ≥ 40
            and L[0] is uppercase
```

```
and L[-1] in {. ! ?}
and {ă, â, î, ș, ț} ⊆ set(L)
and no punctuation run ≥ 3 in L
and L is not ALL-CAPS
```

## Appendix B: Base Model Samples

The examples below provide a qualitative view of the pretrained Neurograf model family before supervised fine-tuning. They are not intended as a benchmark and were not used for model selection. Their purpose is to illustrate how raw completion behavior changes with model scale and to make visible several limitations that aggregate validation loss cannot show directly.

All models were sampled with temperature 0.7 and top-k = 50. All five models received the same prompt in each comparison.

### B.1 Fixed completion probes.

| Model | Prompt: Municipiul București este împărțit administrativ în |
|---|---|
| 160M | două sectoare: sectorul de vest – de Capitală, și sectorul de est – de Capitală. |
| 560M | 6 sectoare, care au și un nou primar. |
| 1.5B | șase sectoare, fiecare având responsabilități bine definite. |
| 3.2B | șase sectoare, fiecare având un primar și un consiliu local. |
| 6B | șase sectoare. |
| **Model** | **Prompt: Pentru depunerea unei cereri la o instituție publică, solicitantul trebuie să** |
| 160M | prezinte următoarele documente: - cerere; - cazier judiciar; - adeverință medicală; |
| 560M | prezinte următoarele documente: actul de identitate, al cărui model este prevăzut în anexa nr. 2 a ordinului. |
| 1.5B | prezinte personal la ghișeul la care se află, să completeze cererea și să o depună. Cererile depuse la ghișeul instituției publice vor fi înregistrate și păstrate până la soluționarea lor. |
| **3.2B** | prezinte la sediul instituției cu documentele necesare, și să completeze o cerere tip cu datele sale personale, și să o semneze. |
| **6B** | se prezinte la ghișeu, acolo unde depune cererea și plătește taxele necesare. |
| **Model** | **Prompt: Legea nr. 53/2003 din România reglementează** |
| **160M** | modalitatea de utilizare a resurselor și patrimoniului local, modul de organizare și funcționare a acestora, modalitățile de administrare, exploatare și administrare a acestora |
| **560M** | dreptul la informare al cetățeanului, ca de altfel și dreptul de petiționare. |
| **1.5B** | regimul general al raporturilor de muncă, precum și reguli speciale în ceea ce privește |
| **3.2B** | drepturile și obligațiile angajaților, precum și modalitățile de protejare a acestora împotriva abuzurilor angajatorilor. |
| **6B** | mai multe tipuri de concedii, precum concediul de odihnă, concediul de maternitate, concediul de paternitate, concediul pentru creșterea copilului, concediul de risc maternal și concediul medical. |
| **Model** | **Prompt: Un sistem informatic destinat administrației publice trebuie proiectat astfel încât** |
| **160M** | să poată fi utilizat în paralel cu activitatea efectivă a administrației publice. |
| **560M** | să permită implementarea de sisteme de control la distanță, să asigure securitatea și să fie ușor de utilizat pentru cetățeni și mediul de afaceri, să permită colaborarea și integrarea sistemelor informatice pentru a crește eficiența proceselor de guvernare și de comunicare, să ofere asistență cetățenilor și companiilor. |
| **1.5B** | să poată fi utilizat în mod continuu, asigurând integrarea și interoperabilitatea proceselor de colectare, administrare și analiză a informațiilor, potrivit unui proiect de lege adoptat, miercuri, de plenul Senatului. |
| **3.2B** | să permită utilizarea lui în mod securizat, se arată într-un proiect de HG elaborat de Ministerul Comunicațiilor și publicat pe site-ul propriu. |
| **6B** | să asigure respectarea următoarelor cerințe:<br><br>- Sistemul informatic trebuie să asigure funcționarea în timp real și în parametri optimi a tuturor activităților care se desfășoară în cadrul instituției;<br><br>- Sistemul informatic trebuie să asigure securitatea informațiilor la nivel de utilizator, grup de utilizatori, sistem de operare, fișiere, directoare, rețele, acces Internet etc.; |
| | - Sistemul informatic trebuie să asigure posibilitatea de a introduce și modifica date în orice moment în vederea asigurării unui grad de securitate ridicat; |
| **Model** | **Prompt: The capital of Romania is** |
| **160M** | the largest European capital, in its development made part of the European Commission (e.g. the euro, the capital and the capital) of the Romanian capital of the European Union |
| **560M** | Bucharest, 1. 089,246 inhabitants, according to the latest statistical data from the World Bank. |
| **1.5B** | Bucharest, in the south-east of the country, on the Dâmbovita river. |
| **3.2B** | Bucharest, where the population is 2.000.000 in 2011. |
| **6B** | Bucharest, followed closely by the city of Cluj-Napoca. Other cities include Timișoara, Iași, Constanța, and Brașov. |

Across these probes, increasing model size generally improves grammaticality, factual association and continuation structure, but factual reliability does not improve monotonically enough for any of the models to be treated as authoritative sources. The administrative and legal examples are particularly illustrative: larger models increasingly produce plausible Romanian institutional prose, but plausibility is not equivalent to correctness.

**B.2 Free-form generation.** To probe longer-form Romanian generation, the 6B base model was given a fictional opening designed to require preservation of setting, characters and a central abstract theme. For this sample, temperature was increased to 1.0 with top-k = 50.

| | |
|---|---|
| **Prompt** | **În subsolul unui institut de cercetare românesc, ecranul vechi de monitor a scintilat dintr-odată în mii de linii de cod strălucitoare, dând viață unui algoritm de inteligență artificială care învăța să simtă românește. Nu știa încă ce înseamnă dorul, dar primul său impuls a fost să caute în baza de date definiția cuvântului om.** |
| **6B** | „Hai să ne apucăm de treabă ca să creăm o soluție tehnică la problema noastră”, spuse unul dintre cercetătorii implicați în proiect, privind spre ecran. Acesta este primul pas, gândi cel de lângă el: nu trebuie să știm definiția din dicționar. |

„Oare asta înseamnă că n-o să avem emoții niciodată?”, întrebă cel de-al treilea cercetător, privind la cei doi camarazi ai săi cum lucrează de zor în subsol, în timp ce ecranul sclipea ca o promisiune nesfârșită către lume.

„Eu zic că da, și asta cât de repede putem”, aprobă programatorul șef, arătând un zâmbet ce părea să spună că deja nu-și mai face griji în privința asta.

„Oare cum ne influențează emoțiile comportamentul și cum putem construi o soluție tehnologică care să învețe să citească emoțiile?”, spuse programatorul șef.

...

Pe măsură ce munca lor avansa, aveau nevoie de alte abilități. Tânărul și-a dat seama că trebuie să învețe să simtă limba română ca pe o limbă străină. Pentru a crea software, are nevoie de fluența exprimării și expresivitatea emoțiilor.

The complete generation remains fluent and largely preserves the semantic field established by the prompt: research, artificial intelligence, emotion and the idea of „a simți românește”. At the same time, its long-range narrative structure is weak. Character identities and roles become difficult to follow, locations drift from the research institute to domestic spaces, dialogue becomes repetitive, and the central artificial-intelligence character becomes increasingly ambiguous relative to the human researchers.

**B.3 What the samples show.** The qualitative examples illustrate a distinction that is easy to miss when considering validation loss alone. Larger Neurograf base models become substantially better language models: their Romanian is more fluent, continuations are more structurally developed, factual associations become stronger, and the 1.5B–6B models can also produce usable English despite being trained as Romanian-first systems.

Scale does not, however, transform next-token prediction into factual verification or long-horizon planning.

The administrative and legal probes show a recurrent failure mode in which the model produces text that is stylistically appropriate to the requested register while inventing the underlying procedure or factual detail. This is consistent with the intended behavior of a base language model trained predominantly on heterogeneous web and institutional text. It learns what Romanian administrative prose looks like, but has no mechanism for determining whether a generated statement corresponds to the currently valid procedure or authoritative source.

This distinction motivates the retrieval-augmented architecture proposed in Chapter 7. For administrative, legal or specialist deployment, the model should provide linguistic generation and interpretation while factual content is supplied from controlled external documents.

The literary example exposes a different limitation. The 6B model can sustain fluent Romanian prose over hundreds of tokens and preserve broad thematic associations, yet gradually loses global narrative constraints. This behavior is plausibly associated with the causal next-token objective: local continuation quality is directly optimized during pre-training, whereas persistence of character identity, setting and narrative plan over long spans is only learned indirectly from the statistical structure of the corpus. The 2,048-token context and absence of instruction or preference post-training in these base checkpoints further limit long-horizon control.

Sampling also matters. Increasing temperature changes which locally plausible continuation is selected and can therefore substantially alter the visible trajectory of a free-form generation. The example should consequently not be interpreted as a deterministic measure of literary ability; it is a qualitative illustration of the type of coherence and failure modes exhibited by the raw 6B checkpoint.

Finally, the English completion probe is consistent with, but does not independently validate, the tokenizer asymmetry measured in Chapter 3. The larger Romanian-first models retain useful English generation capability, plausibly because RoSfera itself contains some English material and because the Romanian-trained tokenizer represents English relatively efficiently. The generation sample alone cannot separate these mechanisms.

## Appendix C: Supervised Fine-Tuned Model Samples.

This appendix provides qualitative examples from the supervised fine-tuned Neurograf models. In contrast with Appendix B, which illustrates raw continuation behavior of the pretrained base family, the examples below probe instruction following, conversational persistence, Romanian cultural knowledge, factual reliability and controlled text generation after supervised fine-tuning.

The examples are not intended as a formal benchmark and were not used for model selection. They were generated interactively from the final SFT checkpoints of the 6B model using the chat protocol described in Chapter 5. The purpose is instead to expose behaviors that are difficult to infer from training loss alone: whether an instruction is followed completely, whether constraints are preserved across a response, whether a conversational format persists across turns, and whether fluent Romanian generation remains factually grounded.

**C.1 Instruction following and constraint preservation.** The first example asks the model to draft a formal request containing several explicit constraints: the location of an event, its timing relative to the working day, a fixed number of participants and a specified budget.

| | |
|---|---|
| **User** | **Prompt: Redactează, în calitate de angajat, o cerere în care să-i soliciți într-un mod convingător managerului filialei locale a companiei la care lucrezi să-ți aprobe organizarea unei petreceri cu ocazia zilei de 8 Martie. Precizează că petrecerea se va desfășura în Sala de Conferințe, va începe cu două ore înainte de terminarea programului, va include 50 de participanți și va avea nevoie de un buget de 5.000 de lei din fondul de protocol al filialei** |
| **Assistant** | **Către:** Managerul Filialei Locale, [Nume Manager]<br>**De la:** [Nume și prenume angajat]<br>**Funcție:** [Funcție]<br>**Data:** 20 februarie 2026<br><br>**Subiect:** Solicitare aprobare organizare petrecere de 8 Martie – Sala de Conferințe, filială<br><br>Stimate domnule/doamnă [Nume Manager],<br><br>Prin prezenta, doresc să vă aduc la cunoștință intenția mea de a organiza, în data de 8 martie, o petrecere de grup destinată tuturor colaboratoarelor și clienților filialei, marcând Ziua Internațională a Femeii. Evenimentul va avea loc în Sala de Conferințe a filialei, începând cu ora 18:00 și se va desfășura în intervalul 20:00–22:00, cu un număr total de 50 de participanți. |

```
**Detalii principale ale evenimentului:**

1. **Data și ora:** 8 martie, ora 20:00
2. **Durata:** 2 ore
3. **Public țintă:** toate femeile din cadrul
filialei și din rețeaua noastră de parteneri locali
4. **Număr participanți:** 50 de persoane (implică 25
de colaboratoare și 25 de clienți)
5. **Buget estimat:** 5.000 lei (incluzând catering,
materiale promoționale, securitate și logistică)

Scopul acestui eveniment este să consolidăm relațiile
de colaborare și să marcăm împreună recunoașterea
contribuției femeilor în cadrul filialei. Din aceste
motive, consider că o organizare în condiții optime
și cu un buget rezonabil sunt esențiale pentru a
asigura un impact pozitiv asupra imaginii și
eficienței filialei.

Vă rog respectuos să analizați și să aprobați această
propunere, pentru a putea implementa planul în cel
mai scurt timp posibil. Sunt disponibil pentru a
coordona detaliile logistice și financiare și pentru
a asigura transparența procesului.

Vă mulțumesc anticipat pentru atenție și sprijin.

Cu stimă,

[Semnătură]
[Nume și prenume]
[Funcție]
[Telefon]
[Email]
```

The response generated demonstrates that supervised fine-tuning has successfully induced recognizable instruction-following behavior. The model adopts an appropriate formal register, constructs a conventional administrative document, introduces headings and fields, preserves the requested number of participants and budget, and provides a plausible justification for the request.

At the same time, the example illustrates imperfect constraint preservation. The requested condition that the event begins two hours before the end of the working day is replaced by several mutually inconsistent absolute times. The model also introduces details that were not present in the prompt, including a division of participants into collaborators and clients and a proposed allocation of the budget to catering, promotional material, security and logistics.

This behavior is representative of an important distinction between instruction following and exact instruction satisfaction. The model clearly recognizes the requested task and its expected textual form, but it may fill unspecified parts of that form with locally plausible details and may fail to preserve all supplied constraints consistently. For procedural or administrative deployment, such outputs therefore remain unsuitable for direct use without validation or grounding against external information.

**C.2 Multi-turn response-format persistence.** A second example probes conversational adaptation. The model is initially asked to write a short fictional story from the perspective of a stray cat in Bucharest. After receiving a brief answer, the user twice requests a substantially longer version. Finally, the same general subject is reframed as a request for a long essay.

| | |
|---|---|
| **User** | **Imaginează-ți că ești o pisică vagabondă în București. Scrie o poveste despre cea mai fericită zi din viața ta.** |
| **Assistant** | Într-o dimineață răcoroasă de primăvară, am pornit pe străzile aglomerate ale capitalei cu coada ridicată, căutând un colț de umbră unde să mă adăpostesc de razele arzătoare. Era prima mea zi de „libertate” și curiozitatea îmi era fără margini. |
| **User** | **Scrie o poveste mai lungă te rog.** |
| **Assistant** | Pe măsură ce traversaam Bulevardul Magheru, am întâlnit un cățel jucăuș care mi-a furat un sărut la ochi, iar eu am simțit cum inima îmi bătea cu putere. Am continuat să merg, iar la colțul cu Biserica Sfântul Gheorghe, am zărit o fereastră deschisă și am intrat să mă adăpostesc de zgomotul orașului. |
| **User** | **Am înțeles dar generează mai multe propoziții, fă povestea mai mare.** |
| **Assistant** | În timp ce mă așezasem pe o bancă de lemn, am auzit un șuierat de tramvai în apropiere și m-am uitat la ceas. Era ora 10:00, iar soarele începea să încălzească aerul. Am simțit că acea zi va deveni una dintre cele mai fericite din viața mea. |
| **User** | **scrie un eseu lung despre viata ta ca o pisica** |
| **Assistant** | **Viața mea ca o pisică:Între haită și confortul căldurii** |

Mă numesc Luna, am cinci ani și îmi trăiesc fiecare zi ca pe o aventură. De la primele pași timizi în curtea stăpânului până la serile petrecute pe brațul său, am învățat să asociez fiecare moment cu o parte din existența mea.

**Dimineața – „trezirea”**
Când sunetul blând al soneriei îmi spune să mă trezesc, îmi întind spatele și deschid ochii, căutând prin prăjina de perne mirosul de lapte și de mâncare. Prima mea „miausă” este un „miau” de încurajare, urmat de o joacă de cinci minute în care îmi exersez reflexele de vânătoare. În timp ce mă ridic, îmi întind lăbuțele pe podea și îmi întind spatele, pregătit pentru o nouă rundă de explorări.

** detectează și vânează**
În curte, Luna se strecoară sub frunzele de viță și se ascunde în spatele tufelor de iarbă. La prima mișcare, simt mirosurile de iarbă umedă și de mâncare ascunsă, iar la fiecare „miau” de la un șoarece, îmi ajustez poziția, pregătindu-mă pentru o captură. Încet, prind o pasăre mică, pe care o țin în gură cu grijă, și o trec prin iarbă, unde Luna scoate un sunet de „miau” de mulțumire.

**După vânătoare – „revenirea la căldură”**
Când mă așez pe perna moale a stăpânului, simt cum blana mea se încălzește încet, ca și cum soarele ar fi început să încălzească pământul. În timp ce privesc pe fereastră, îmi favorabilez gândurile: „Ce frumos e să fii parte dintr-un mic univers”. Îmi întind labele în față, lăsându-le să se încălzească de la corpul lui Mihai, apoi îmi înfășor capul sub brațul său, căutând acea senzație de siguranță care îmi dă încredere.

**Seara – „somnul și visul”**
După ce am mâncat resturile de pește lăsate pe masă, mă așez pe covorul din sufragerie, unde îmi simt corpul ca pe o pisică ce se întinde pe un covor de puf. Închid ochii și îmi imaginez un peisaj nocturn plin de stele, în timp ce aud lătratul ușor al vecinilor. Somnul îmi este întrerupt doar de o scurtă întrerupere, când aud un șuierat de la ușa de la intrare, ca o invitație la o nouă aventură.

**Mâine – „revenirea la rutină”**
În zori, mă trezesc cu sunetul clopoțelului de la tramvai, care îmi amintește că este timpul să merg la

| | |
|---|---|
| | vânătoare din nou. În timp ce îmi întind coada și îmi verific lăbuțele, îmi dau seama că fiecare zi este o combinație de vânătoare, vânătoare și somn. Mă întind pe perna moale, lăsând căldura corpului meu să se identifice cu căldura soarelui ce răsare peste oraș. Astfel, viața mea se trăiește ca o pisică ce își găsește echilibrul între haos și liniște, mereu pregătită să exploreze și să se odihnească în confortul căldurii și al siguranței. |

The behavior is notable because the first three assistant turns remain similar in length despite increasingly explicit requests for a longer response. The model preserves the subject and continues the narrative, but it does not substantially alter the response length. When the task is reformulated as a long essay, however, the response changes abruptly: the model generates a title, multiple sections and a much longer structured text.

This qualitative example is consistent with the instruction-format persistence observed during SFT development and discussed in Chapter 5. The model appears to associate some task formulations with characteristic response structures and lengths. Once a short narrative pattern has been established within a conversation, subsequent requests to modify its length may be weaker than the previously established interaction pattern. A sufficiently different task formulation can instead activate another learned response mode.

The example does not establish a general causal mechanism, and no controlled ablation was performed. It does, however, show that the model is capable of long-form generation while still failing to respond appropriately to a request to increase the length of an existing answer. The limitation is therefore not simply an inability to generate many tokens.

The longer response also exposes secondary coherence problems. The stray cat becomes a domestic animal with an owner, the narration alternates between first- and third-person perspectives, new names and circumstances are introduced without support from the preceding conversation, and several malformed expressions appear. These errors suggest that increased output length does not necessarily correspond to improved global consistency.

**C.3 Romanian cultural and general-purpose generation.** The remaining examples sample several types of Romanian-language instruction following, including literary interpretation, travel recommendation, historical explanation, factual description and poetry.

| | |
|---|---|
| **User** | **Explică și argumentează în maximum 300 de cuvinte afirmația lui Lucian Blaga că „veșnicia s-a născut la sat”.** |
| **Assistant** | Lucian Blaga afirmă că „veșnicia s-a născut la sat” pentru că satul ocupă un loc central în opera sa, simbolizând eternitatea și permanența. Acolo, comunitatea și peisajul rural devin fundalul metafizic al existenței, iar ritualurile și credințele populare reflectă dorința de a păstra legătura cu trecutul și cu „eternitatea”. Astfel, Blaga nu doar descrie satul, ci îl transformă în un simbol al infinitului. |
| **User** | **Ce oraș medieval din România îmi recomanzi să vizitez? Descrie-mi-l în maximum 200 de cuvinte.** |
| **Assistant** | Sighișoara, în inima Transilvaniei, este un oraș medieval plin de povești și legende, cu turnuri și case colorate. Vei simți farmecul său medieval, iar în jurul său se întind peisaje pitorești și tradiții vii. |
| **User** | **Cum a evoluat popularitatea lui Nicolae Ceaușescu de-a lungul perioadei în care a condus România?** |
| **Assistant** | Popularitatea lui Nicolae Ceaușescu a fluctuat în funcție de contextul istoric. În anii 1950-1960 a fost perceput ca un lider carismatic și un modernist, datorită industrializării rapide și a proiectelor de infrastructură. Această imagine a consolidat sprijinul popular, în special în rândul tinerilor și al muncitorilor din orașe. În anii 1970-1980, pe fondul crizei economice și al nemulțumirii generale, popularitatea a scăzut, culminând cu nemulțumirea populară și, în final, cu revoluția din 1989. |
| **User** | **Elaborează o scurtă descriere narativă a celei mai populare mașini produsă în România.** |
| **Assistant** | **„Răsăritul de la Craiova” – modelul care a transformat România pe harta automobilistică**<br><br>În zorii unui nou mileniu, orașul Craiova a fost martor la nașterea unei legende auto: „Răsăritul de la Craiova”. Dezvoltat de producătorul auto local, acest model a devenit rapid cel mai bine vândut automobil din România, depășind așteptările inițiale ale pieței și conturând un nou standard de mobilitate urbană.<br><br>Cu un design robust, dar elegant, și un motor pe benzină de 1,4 l, „Răsăritul” a combinat puterea |
| | necesară unui trafic intens cu un consum redus, datorită sistemului de injecție directă și a tehnologiilor avansate de gestionare a motorului. Transmisia manuală cu cinci trepte a oferit șoferului un echilibru perfect între confort și economie, în timp ce sistemul de climatizare automată a menținut o temperatură optimă în fiecare călătorie.<br><br>Inovațiile software au adus un nivel de siguranță și conectivitate fără precedent: senzorii de parcare, sistemul de frânare automată de urgență și integrarea completă a smartphone-ului șoferului au transformat „Răsăritul" într-un vehicul autonom, capabil să navigheze prin traficul urban cu o precizie remarcabilă. În plus, producția în România a stimulat crearea de locuri de muncă în sectorul auto local, consolidând ecosistemul industrial românesc.<br><br>Pe lângă popularitatea sa rapidă, „Răsăritul" a avut un impact semnificativ asupra imaginii României ca producător de mașini de calitate. Exporturile sale au crescut cu peste 40 % în primul deceniu de la lansare, iar modelul a fost inclus în clasamentele anuale de vânzări pentru categoria „compact sedană", evidențiind forța de vânzare pe piața internă și în Uniunea Europeană.<br><br>În prezent, „Răsăritul de la Craiova" este recunoscut ca un simbol al progresului auto românesc. continuă să fie un exemplu al adaptabilității și inovației, demonstrând că tradiția și tehnologia se pot îmbina pentru a genera vehicule care nu doar că satisfac nevoile zilnice ale șoferilor, ci și inspiră viitorul mobilității. |
| **User** | **Elaborează succint o poezie despre un avion care zboară pe ruta București - Cluj.** |
| **Assistant** | Pe aripi de pană, pe aripi de hârtie,<br>Un avion ușor, cu ruta București-Cluj,<br>Zboară lin, cu destinație de poveste,<br>În cerul românesc, cu dor și cu gânduri. |

The interpretation of Lucian Blaga's statement „veșnicia s-a născut la sat" is an example of comparatively successful behavior. The model identifies the village as a symbol of permanence and continuity and relates this interpretation to community, rural space, ritual and popular belief. The answer is concise and remains within the requested form, although it contains the grammatical construction „în un simbol" rather than „într-un simbol". The example illustrates that the SFT model can produce coherent Romanian humanities-style explanation while still exhibiting occasional surface-level language errors.

The recommendation of Sighișoara is similarly concise and appropriate to the instruction. It demonstrates that the model can produce a short, relevant answer without unnecessary elaboration when the task permits it.

The answer concerning Nicolae Ceaușescu illustrates a different limitation. It provides a smooth historical narrative but compresses the chronology too aggressively and attributes popularity during the 1950s and early 1960s to Ceaușescu despite the fact that he did not become Romania's leader until 1965. The response therefore sounds plausible at the discourse level while containing a substantive chronological error. This is precisely the type of failure for which generative fluency should not be treated as evidence of factual reliability.

The automobile example provides an even clearer case of unconstrained hallucination. Rather than identifying an actual Romanian-produced automobile, the model invents a vehicle named „Răsăritul de la Craiova" and subsequently constructs an extensive fictional history around it, including technical specifications, market performance and export statistics. The resulting text is stylistically coherent and internally developed, but its factual premise is fabricated.

This example is particularly useful because it demonstrates that SFT does not remove the generative tendency of the underlying language model. It teaches the model how to answer an instruction, but not how to distinguish reliably between recalled information and a plausible continuation when factual evidence is absent.

The short poem about a Bucharest–Cluj flight demonstrates competent task recognition and the ability to switch into a different textual genre. Its language is simple and the poem is just bad, but the requested form and topic are preserved.

**C.4 Interpretation.** Taken together, the examples show that supervised fine-tuning changes Neurograf's behavior substantially relative to the base checkpoints. The SFT model recognizes explicit user requests, produces appropriate document and genre conventions, maintains multi-turn conversational context and can adapt its output style to administrative writing, literary interpretation, recommendation, narration and poetry.

The main remaining limitations are not primarily failures of basic Romanian fluency. Instead, they involve:

- incomplete preservation of multiple simultaneous constraints;
- persistence of learned response formats across conversational turns;
- deterioration of entity and narrative consistency in longer generations;
- occasional malformed Romanian constructions;
- confident factual errors;
- and fabrication of detailed but unsupported information.

These failure modes are consistent with the composition and purpose of the SFT stage. The synthetic instruction corpus teaches mappings between instructions and desirable response forms, but it does not provide an external mechanism for factual verification. Moreover, synthetic examples can encourage recognizable task templates, particularly when similar task types repeatedly exhibit characteristic answer structures.

The examples therefore reinforce two conclusions drawn in Chapters 5 and 7. First, improving the next Neurograf instruction model will require not only more synthetic examples but better coverage of heterogeneous multi-turn interactions, corrections, follow-up requests and naturally occurring Romanian dialogue. Real Romanian conversational and task-oriented data would be particularly valuable for reducing dependence on synthetic response templates.

Second, factual reliability should be addressed separately from conversational behavior. For administrative, legal, historical or specialist use, retrieval-augmented generation is a more appropriate mechanism for supplying authoritative information than expecting the SFT model to reproduce such knowledge reliably from its parameters. Tool use can similarly externalize calculations and other operations for which unconstrained generation is an unreliable substitute.

These samples should consequently be interpreted neither as a benchmark of Romanian language understanding nor as a curated showcase of ideal outputs. They provide a qualitative view of what the first Neurograf SFT models learned successfully and, equally importantly, where the resulting assistant behavior remains incomplete.

# Appendix D: Qualitative Embedding Probe

This appendix documents the fixed Romanian passage set used for the final-checkpoint embedding comparison presented in Section 7.7. The purpose of the probe is qualitative: it provides a transparent view of whether independently phrased passages describing the same broad subject occupy nearby regions of the learned embedding space.

The probe contains ten manually constructed passages grouped into five semantic pairs. Each pair describes the same general topic using different wording, while passages belonging to different pairs are intentionally unrelated at the broad semantic level.

For each passage, the final hidden states of the causal decoder were mean-pooled over all tokens. The resulting vector was converted to FP32. The mean vector of the complete ten-passage set was then subtracted from every embedding, after which each centered vector was L2-normalized. Pairwise similarities were computed as dot products between the normalized vectors.

The resulting values are therefore sample-mean-centered cosine similarities. They should not be interpreted as raw retrieval scores or universal semantic-similarity thresholds.

## D.1 Fixed passage set

| | |
|---|---|
| **Physics 1** | Fizicienii studiază particulele elementare pentru a înțelege structura materiei și interacțiunile fundamentale care guvernează comportamentul acesteia la scară microscopică. |
| **Physics 2** | Experimentele de fizică nucleară urmăresc proprietățile nucleelor atomice și procesele prin care acestea interacționează, se transformă sau se dezintegrează. |
| **Drone 1** | Inginerii dezvoltă sisteme de control pentru drone autonome, folosind senzori inerțiali, algoritmi de navigație și bucle de reacție pentru menținerea stabilității în zbor. |
| **Drone 2** | Un vehicul aerian fără pilot își menține traiectoria cu ajutorul giroscoapelor, accelerometrelor și algoritmilor automați de control care corectează permanent orientarea aparatului. |
| **Cooking 1** | Un bucătar pregătește o supă tradițională folosind legume proaspete, condimente și un timp atent controlat de fierbere pentru a obține aroma și consistența dorite. |
| **Cooking 2** | În bucătărie, ingredientele sunt combinate și gătite treptat, iar temperatura și timpul de preparare sunt ajustate pentru a obține gustul și textura potrivite. |
| **Nature 1** | O pădure matură oferă habitat pentru numeroase specii de plante și animale și contribuie la reglarea temperaturii, umidității și ciclului apei la nivel local. |
| **Nature 2** | Ecosistemele forestiere conțin arbori, plante, fungi și animale care interacționează într-un mediu natural complex, formând relații ecologice stabile în timp. |
| **Admin 1** | Cetățenii pot depune cereri la instituțiile publice pentru eliberarea documentelor, obținerea autorizațiilor și soluționarea diferitelor probleme administrative. |
| **Admin 2** | O primărie procesează solicitările locuitorilor, emite acte și autorizații și gestionează servicii publice locale destinate comunității. |

**D.2 Intended semantic relationships.** The manually defined positive relationships are:

*Physics 1 ↔ Physics 2;*

*Drone 1 ↔ Drone 2;*

*Cooking 1 ↔ Cooking 2;*

*Nature 1 ↔ Nature 2;*

*Admin 1 ↔ Admin 2.*

All remaining passage combinations are treated only as non-matched comparisons for descriptive purposes. They are not labeled hard negatives, since some may share lexical, stylistic, or conceptual features despite belonging to different broad categories.

For both final embedding checkpoints, the intended semantic partner is the highest-similarity non-self-passage for all ten individual passages. The detailed pairwise matrices and the comparison of matched-pair similarities are presented in Section 7.7.

**D.3 Scope and limitations.** The probe was designed to make the learned representation geometry visually interpretable rather than to estimate retrieval quality. The passages are short, manually constructed, topically clean and limited to five broad domains. Real retrieval corpora contain substantially more difficult cases, including long documents, overlapping subjects, partially relevant passages, duplicated information and hard negatives that differ only in small factual details.

The sample-mean-centering operation further makes the numerical similarities dependent on the composition of this particular ten-passage set. The negative values visible in the matrices therefore represent relative positioning after centering and should not be interpreted as evidence of intrinsic semantic opposition between the corresponding texts.

A production evaluation would require a substantially larger Romanian retrieval benchmark with natural documents, independent queries, graded relevance judgments and candidate pools large enough to avoid the saturation observed in the contrastive training task. The present probe is intended only as a transparent qualitative demonstration that both final Neurograf embedding models recover the manually defined semantic pairings, with stronger local separation in the 560M model.